%% file: neurips_2026.tex
\documentclass{article}

\PassOptionsToPackage{numbers,sort&compress}{natbib}
\usepackage[main, preprint]{neurips_2026}

\usepackage[utf8]{inputenc} 
\usepackage[T1]{fontenc}    
\usepackage{url}            
\usepackage{booktabs}       
\usepackage{amsfonts}       
\usepackage{amsmath, amssymb, amsthm} % Unified math block
\usepackage{mathtools}      % Extends amsmath
\usepackage{bm}             % Best for bold math symbols
\usepackage{nicefrac}       
\usepackage{microtype}      
\usepackage[dvipsnames]{xcolor} % Unified color block (dvipsnames gives more options)
\usepackage{enumitem}
\usepackage{tcolorbox}
\usepackage{graphicx}
\usepackage{subcaption}
\usepackage{float}
\usepackage{todonotes}
\usepackage{multirow}
\usepackage{natbib}

\usepackage{hyperref} 

\newtheorem{proposition}{Proposition}

\definecolor{accentblue}{RGB}{24,95,165}
\definecolor{findinggreen}{RGB}{15,110,86}

\newtcolorbox{keytakeaway}[1][]{
  colframe=blue!30!black,
  colback=blue!3!white,
  boxrule=0.4mm,
  arc=1.5mm,
  title=\textbf{Key Takeaway},
  fonttitle=\bfseries\small, % Reduces the height of the colored bar
  toptitle=0.5mm,            % Minimal padding above title text
  bottomtitle=0.5mm,         % Minimal padding below title text
  fontupper=\small,          % Smaller body text
  boxsep=0.5mm,              % Reduces base padding
  left=1mm, right=1mm,       % Text pushes to edges
  top=0.5mm, bottom=0.5mm,   % Text pushes to header/bottom
  before skip=4pt,           
  after skip=4pt,
  #1
}

\newtcolorbox[auto counter]{takeaway}[1][]{
  colframe=blue!30!black, 
  colback=blue!3!white,   
  boxrule=0.4mm,             % Slightly thinner rule looks more elegant
  arc=1.5mm,                 % Slightly sharper corners look more modern
  title=Takeaway~\thetcbcounter,
  fonttitle=\bfseries\small, 
  toptitle=0.5mm,            
  bottomtitle=0.5mm,         
  fontupper=\small,          
  boxsep=0.5mm,              
  left=1mm,                  
  right=1mm,                 
  top=0.5mm,                 
  bottom=0.5mm,              
  before skip=4pt,           
  after skip=4pt,
  #1
}

\newtcolorbox{definitionbox}[1][]{
  colframe=black!40,         % Grey border
  colback=black!3!white,     % Very light grey background
  boxrule=0.4mm,
  arc=1.5mm,
  fontupper=\small,          % Smaller body text
  boxsep=0.5mm,              % Reduces base padding
  left=1mm, right=1mm,       % Text pushes to edges
  top=1mm, bottom=1mm,       % Minimal padding top and bottom
  before skip=4pt,           
  after skip=4pt,
  #1
}

\title{On the Geometric Bottlenecks of Cross-Lingual Knowledge Editing}

\author{%
  Aastha A K Verma\thanks{Equal contribution; authors listed alphabetically by first name.}~\;~ 
    Anwoy Chatterjee$^{*}$~\;~ 
    Sparsh Jain~\;~ 
    Tanmoy Chakraborty \\\\
  Indian Institute of Technology Delhi\\
  New Delhi, India\\
  {\small \texttt{\{aastha.v1411, anwoychatterjee\}@gmail.com}}, {\small \texttt{\{sparsh, tanchak\}@ee.iitd.ac.in}}
}

\begin{document}

\maketitle
\setcounter{footnote}{0}

\input{sections/abstract}
\input{sections/intro}

\input{sections/related_works}

\input{sections/setup}
\input{sections/geometry}

\input{sections/impossibility}
\input{sections/alignment}
\input{sections/disc_and_conc}

\section*{Acknowledgements}
We thank the anonymous reviewers and the area chair for their thoughtful feedback. Anwoy Chatterjee acknowledges the support of the Google PhD Fellowship. Tanmoy Chakraborty acknowledges the support of the Rajiv Khemani Young Faculty Chair Professorship in Artificial Intelligence, and the ICMR Extramural Research (Centre for Advanced Research) Project Grant (EM/DEV/CAR-2025/00725). The authors declare no competing interests.

\bibliographystyle{plainnat}
\bibliography{references}

\appendix
\input{sections/appendix_toc}
\input{sections/appendix_faqs}
\input{sections/appendix_temp}

%%%%%%%%%%%%%%%%%%%%%%%%%%%%%%%%%%%%%%%%%%%%%%%%%%%%%%%%%%%%

% \newpage
% \input{checklist.tex}

\end{document}

%% file: sections/abstract.tex
\begin{abstract}
Transformer-based autoregressive multilingual language models (LMs) are often assumed to store factual knowledge in an \textit{interlingua}: a language-agnostic latent space. If this assumption holds, a parameter update applied in a source language during knowledge editing should inherently propagate to its semantic equivalents in target languages. However, empirical benchmarks show that cross-lingual propagation of edits often degrades substantially, a degradation that prior work treats largely as an algorithmic symptom rather than a structural bottleneck. In this work, we investigate whether representational anisotropy creates a structural bottleneck for source-only associative-memory editing, and perform a geometric dissection of the cross-lingual propagation gap. Using associative memory-based ``locate-then-edit'' methods, including ROME, MEMIT, and AlphaEdit, as analytical frameworks, we derive a closed-form decomposition of this gap into key-space propagation, parallel residual mismatch, and orthogonal residual mismatch. Our empirical findings show that cross-lingual propagation is primarily bottlenecked by a large orthogonal residual component, indicating that source-language edits often fail to span the target-language update directions required for successful transfer. Furthermore, we show that post-hoc correction using only source-language residual statistics encounters a large residual-prediction barrier. Finally, we perform a structural realignment via a targeted contrastive alignment objective and find that it substantially improves key-space propagation and partially improves cross-lingual edit transfer, while leaving value-space mismatch as a remaining bottleneck \footnote{Our source code is available at \url{https://github.com/C-anwoy/cross-lingual-edit-geometry}.}.
\end{abstract}

%% file: sections/intro.tex
\section{Introduction}\label{sec:intro}

Cognitive neuroscience studies of the bilingual human brain demonstrate that while conceptual semantics are largely shared across languages, the neural processing of distinct lexical forms remains highly language-specific \citep{doi:10.1073/pnas.2503721123, Correia332, KROLL1994149, Dijkstra_van_Heuven_2002, 10.1093/oso/9780195151770.003.0014}. This biological segregation raises a fundamental question for autoregressive multilingual language models (LMs): how and where is multilingual factual knowledge structurally organized within the parameter space? A prevailing hypothesis assumes that modern LMs converge on an ``interlingua'' --- a language-agnostic latent space where semantic facts are stored independent of the acquisition language \citep{zeng-etal-2025-converging, del2022similaritysentencerepresentationsmultilingual, pires-etal-2019-multilingual}. Probing the validity of this hypothesis is critical not only for structural interpretability but for the equitable deployment of LMs; if knowledge remains rigidly compartmentalized by language, ensuring factual consistency and alignment across diverse linguistic contexts becomes a fundamental challenge.

We investigate the interlingua hypothesis by probing the model's weight space through the lens of ``locate-then-edit'' knowledge editing methods \citep{rome, memit, alphaedit, yao-etal-2023-editing, zhang2024comprehensivestudyknowledgeediting}. If an LM utilizes a language-agnostic parameter space for factual storage, a targeted weight update applied to a fact in a source language (e.g., English) should inherently propagate to its semantic equivalents in all target languages. Under this assumption, cross-lingual transfer is an automatic geometric consequence of the update. However, empirical evaluations contradict this expectation. As demonstrated in Table \ref{tab:motivation_gap}, applying associative memory-based editing algorithms on the Llama-3.1-8B model reveals that while native-language edits generally succeed (except for MEMIT, which is weak even in English in our batch setting), cross-lingual propagation often decays sharply as typological distance increases. Although this degradation is consistent with observations in prior literature \cite{mlake, clmbench, durrani-etal-2025-editing, beniwal-etal-2024-cross, wang-etal-2024-cross}, the underlying mechanisms governing this failure remain understudied. This poses a critical question: \textit{is the lack of cross-lingual propagation driven by algorithmic failures specific to current editing methods, or are locate-then-edit paradigms inherently bottlenecked by the representational structure of LMs?} 

\begin{table*}[t]
\small
\centering
\setlength{\tabcolsep}{3pt} 
\begin{tabular}{l ccccc ccccc}
\toprule
& \multicolumn{5}{c}{\shortstack{\textbf{zsRE} \\ (Edit Language: English)}} & \multicolumn{5}{c}{\shortstack{\textbf{CounterFact} \\ (Edit Language: English)}} \\
\cmidrule(lr){2-6} \cmidrule(lr){7-11}
\textbf{Method} & \textbf{English} & \textbf{German} & \textbf{Hindi} & \textbf{Urdu} & \textbf{Chinese} & \textbf{English} & \textbf{German} & \textbf{Hindi} & \textbf{Urdu} & \textbf{Chinese} \\
\midrule
\textbf{ROME}      & 76.12 & \textbf{82.77} & 63.09 & 61.47 & 51.29 & 50.78 & \textbf{72.62} & 49.61 & 46.70 & 33.59 \\
\textbf{MEMIT}     & \textbf{27.37} & 19.12 & 19.94 & 15.41 & 2.20  & 5.15  & \textbf{13.02} & 7.09  & 8.65  & 0.87  \\
\textbf{AlphaEdit} & \textbf{75.58} & 59.29 & 36.64 & 38.81 & 36.64 & \textbf{96.99} & 89.61 & 58.74 & 54.08 & 57.86 \\
\bottomrule
\end{tabular}
\caption{\textbf{The cross-lingual transfer gap across editing methods in Llama-3.1-8B (ROME: layer 20; MEMIT and AlphaEdit: layers 16--20).} Values represent Efficacy ($\uparrow$): the proportion of cases where post-edit probability of the target answer exceeds that of the original answer. Native (English) and related (German) edits succeed, but transfer degrades sharply for typologically distant languages. Full evaluations, including Gemma-3-4B results, are detailed in Table~\ref{tab:baselines_numbers}.}
\vspace{-9.5pt}
\label{tab:motivation_gap}
\end{table*}

Recent empirical studies have observed that cross-lingual edit vectors reside in nearly orthogonal, disjoint subspaces \citep{clmbench}. However, prior work treats this language anisotropy largely as an observational symptom. In this work, we ground this observation within a formal geometric framework. Because editing methods like ROME \citep{rome}, MEMIT \citep{memit}, and AlphaEdit \citep{alphaedit} compute closed-form weight updates derived entirely from edit-language representations, they provide a precise analytical lens. 
By geometrically decomposing these updates and projecting them onto target-language spaces, we show that this orthogonal mismatch is not merely an empirical artifact, but a source-inaccessible component that limits zero-shot cross-lingual transfer under source-only associative-memory editing.

To this end, we investigate four primary research questions: (i) How can the cross-lingual transfer gap be decomposed within a closed-form weight update? (ii) What specific geometric bottlenecks dictate transfer failure as typological distance increases? 
(iii) Under what assumptions does source-only correction face a lower bound when using only edit-language statistics?
(iv) Can structural realignment of internal representations mitigate these transfer bottlenecks?

To answer these questions, we analyze the cross-lingual propagation of edits in Llama-3.1-8B and Gemma-3-4B across languages spanning distinct typologies (English, German, Hindi, Urdu, and Chinese), using multilingual versions \citep{bmike} of the zsRE \citep{zsre} and CounterFact \citep{rome} datasets.  
Through both theoretical and empirical analyses, we show that, for associative-memory editors, representational anisotropy
is a major factor underlying cross-lingual transfer failures. Our core contributions can be summarized as:
\begin{itemize}[nosep, wide, labelwidth=!, labelindent=0pt]
    \item \textbf{Exact decomposition of the propagation gap:} We derive a closed-form geometric breakdown of cross-lingual propagation, isolating key-space propagation, parallel residual mismatch, and orthogonal residual mismatch (\S\ref{sec:key_space_prop}, \S\ref{sec:value_space_barrier}).
    \item \textbf{Orthogonal mismatch as a dominant empirical bottleneck:} We show that target-language residuals contain large components orthogonal to the source-language residual span across the evaluated models, languages, and editing methods, limiting zero-shot transfer (\S\ref{sec:value_space_barrier}, \S\ref{sec:downstream_penalty}).
    \item \textbf{A conditional limit for source-only correction:} We formalize post-hoc correction as a residual prediction problem and show that, when correction is restricted to source residual information, the expected gap is lower-bounded by the private variance of the target residual (\S\ref{sec:impossibility}). Corrections that also use keys, prompts, or language identity are not covered by this bound.
    \item \textbf{Diagnostic validation via structural realignment:} We show that targeted contrastive alignment at mid-to-late layers substantially improves key-space propagation but yields mixed transfer gains at some cost to downstream performance, while the persistence of value-space mismatch clarifies why alignment alone is insufficient (\S\ref{sec:interventions}).
\end{itemize}

\begin{keytakeaway}
For locate-then-edit methods, cross-lingual editing fails when source-language updates do not align with target-language key and residual geometry. \textbf{Without additional alignment or multilingual edit information, zero-shot transfer through source-only associative-memory updates remains strongly constrained}.
\end{keytakeaway}

%% file: sections/related_works.tex
\section{Related Work}

\paragraph{Parameter-Based Editing and the Multilingual Gap.} 
The ``locate-then-edit'' paradigm -- particularly least-squares-based associative memory methods like ROME \citep{rome}, MEMIT \citep{memit}, and constraint-based variants like AlphaEdit \citep{alphaedit} -- has proven highly effective for modifying targeted facts within a single language. We specifically focus on this family of methods because their closed-form weight updates allow for exact geometric dissection, unlike opaque hypernetworks \citep{Sinitsin2020Editable, de-cao-etal-2021-editing, hase-etal-2023-methods, mitchell2022fast, pmlr-v162-mitchell22a} or heuristic neuron alterations \citep{memla, zhang-etal-2025-multilingual, 10.1609/aaai.v39i24.34807}. However, applying these parameter updates to multilingual contexts reveals noticeable performance degradation. Recent evaluation frameworks (e.g., MLaKE \citep{mlake}, CLM-Bench \citep{clmbench}) and extensive surveys \citep{durrani-etal-2025-editing} consistently document this cross-lingual transfer gap. 
These studies broadly attribute the failure to language anisotropy, noting that LMs tend to encode equivalent multilingual concepts into misaligned representational subspaces \citep{durrani-etal-2025-editing, mousi-etal-2024-exploring, clmbench}. \citet{ifergan-etal-2025-beneath} also find that consistent factual answers across languages do not necessarily imply shared representations, and that edit transfer is stronger between languages using the same script.

\textbf{Addressing Cross-Lingual Misalignment.} 
To mitigate these misalignments, recent works often rely on architectural workarounds. Approaches such as LU-LAFNs \citep{zhang-etal-2025-multilingual} and MEMLA \citep{memla} bypass weight-space transfer by routing updates through external caches or applying parameter masks. While other subspace-aware frameworks attempt to force cross-lingual synchronization \citep{sun-etal-2025-mitigating-negative, wu-etal-2025-edit, mela-etal-2024-mass}, the current literature primarily treats poor transfer as an algorithmic symptom to be engineered around. For instance, while CLM-Bench empirically observes that cross-lingual edit vectors reside in nearly orthogonal spaces, it leaves the underlying mechanics of the parameter transfer itself unexplored. Our work seeks to ground these empirical observations within the structure of the update. By geometrically dissecting the closed-form solutions of associative-memory models,
we show how this orthogonal mismatch arises within the update geometry and why it limits
source-only zero-shot transfer.

%% file: sections/setup.tex
\section{Background: Associative Memory in the Locate-Then-Edit Paradigm}\label{background}

\textbf{Associative Memory in Language Models.}
Multilayer Perceptron (MLP) sub-layers in Transformers are shown to function as linear associative key-value memories \citep{geva-etal-2021-transformer, rome}. For a given factual association $(s, r, o)$, the MLP processes an input subject $s$ by mapping it to a specific key vector $\bm{k} \in \mathbb{R}^{d_{in}}$ via its first linear layer and activation function. The second linear layer, parameterized by the down-projection weight matrix $W \in \mathbb{R}^{d_{out} \times d_{in}}$, retrieves a corresponding value vector $\bm{v} \in \mathbb{R}^{d_{out}}$ that encodes the target object $o$ \footnote{Note that the \textit{key} and \textit{value} vectors defined here are different from the key and value vectors in the attention layers.}. We use this formulation to describe what locate-then-edit methods write, not as a complete account of factual recall, which is a multi-stage process~\cite{geva-etal-2023-dissecting} in which subject and relation representations appear largely transferable across languages~\cite{fierro-etal-2025-multilingual}. In particular, the edit layer is where these methods write, not necessarily where a fact is stored~\cite{hase2023localization}.

\textbf{Formalizing Cross-Lingual Editing.}
Building on this memory formulation, we formally define the problem of cross-lingual knowledge editing \citep{wang-etal-2024-cross}. Let $(s_1, r_1, o_1)$ represent a fact in a source language $\ell_1$ (e.g., English), and let $(s_2, r_2, o_2)$ represent its exact semantic equivalent in a target evaluation language $\ell_2$. A successful edit modifies the original weights to $\hat{W} = W + \Delta W$, ensuring the source key $\bm{k}_1^*$ maps to a new target value $\bm{v}_1^*$. The goal of a \textit{cross-lingual} edit is for this update to inherently propagate the newly injected knowledge to the target language, successfully mapping the target-language key $\bm{k}_2^*$ to output $o_2$. Crucially, this transfer must occur zero-shot: the target-language fact $(s_2, r_2, o_2)$ is completely unseen during the optimization of $\Delta W$.

\textbf{A Unified Framework for Parameter Editing.}
To inject new knowledge without inducing catastrophic forgetting, current editing methods solve constrained optimization problems over the uncentered second-moment matrix $C = \mathbb{E}[\bm{k}\bm{k}^\top]$, which represents pre-existing key distributions over a large text corpus. We analyze three state-of-the-art frameworks that represent progressive geometric constraints on this update. \textbf{ROME} computes a minimum-disturbance rank-1 update for a single fact. The closed-form solution yields $\Delta W_{\text{ROME}} = \bm{r}_1 \frac{(\bm{k}_1^*)^\top C^{-1}}{(\bm{k}_1^*)^\top C^{-1}\bm{k}_1^*}$, where $\bm{r}_1 = \bm{v}_1^* - W\bm{k}_1^*$ is the \emph{edit residual}, i.e., the change that the edit writes at the source key ($\Delta W_{\mathrm{ROME}}\,\mathbf{k}^*_1 = \mathbf{r}_1$). \textbf{MEMIT} extends ROME to batch-edit $n$ facts across multiple critical layers. By aggregating keys and residuals into matrices $K_1$ and $R_1$, the update generalizes to $\Delta W_{\text{MEMIT}} = R_1 K_1^\top (C + K_1 K_1^\top)^{-1}$. \textbf{AlphaEdit} strictly preserves pre-existing knowledge by restricting the parameter update to the null space of $C$. This is achieved by constructing an orthogonal projection matrix $P_{\perp}$ onto this null space and applying it to the input key space via right-multiplication. The resulting constrained update is defined as $\Delta W_{\text{AE}} = \Delta W_{\text{MEMIT}} P_{\perp}$. More details about the individual editing methods, along with the implementation details and choice of hyperparameters, are provided in Appendix ~\ref{app:methods}.

\paragraph{Experimental Setup.}
A fundamental property shared across these associative memory frameworks is that the weight update $\Delta W$ is constructed entirely using source-language geometry ($\bm{k}_1^*, \bm{v}_1^*, C_1$). This deterministic reliance on source statistics allows us to project $\Delta W$ analytically onto target-language representations, providing a precise mechanism to trace cross-lingual signal decay. We evaluate these methods in Llama-3.1-8B \citep{grattafiori2024llama3herdmodels} and Gemma-3-4B \citep{gemmateam2025gemma3technicalreport} using the multilingual versions \citep{bmike} of the zsRE \citep{zsre} and CounterFact \citep{rome} datasets across English (\texttt{en}), German (\texttt{de}), Hindi (\texttt{hi}), Urdu (\texttt{ur}), and Chinese (\texttt{zh}). All our experiments were executed on a single NVIDIA H200 NVL (141GB) GPU. \textbf{Due to space constraints, we primarily present results for Llama-3.1-8B in the main paper; corresponding analyses for Gemma-3-4B are provided in the Appendix.} First, we geometrically decompose the cross-lingual transfer gap to establish the structural bottlenecks governing transfer failure. Second, we introduce a targeted intervention to correct these failing geometric components without requiring bilingual data at inference time.

%% file: sections/geometry.tex
\section{Geometric Dissection of the Cross-Lingual Transfer Gap}
\label{sec:gap_dissection}

Building upon our associative memory framework, we now derive a closed-form geometric decomposition of the error that arises when an edit optimized in a source language ($\ell_1$) is evaluated zero-shot in a target language ($\ell_2$). We isolate this transfer gap into key-space and value-space components, and empirically validate our derivations on Llama-3.1-8B (with supporting results for Gemma-3-4B deferred to Appendix). All formal proofs are provided in Appendix \ref{app:proofs}.

\input{sections/geometry_key}
\input{sections/geometry_residual}
\input{sections/geometry_transfer}

%% file: sections/geometry_key.tex
\subsection{Key-Space Propagation is Bottlenecked by Typological Distance}
\label{sec:key_space_prop}

When we edit a model using the source language $\ell_1$ and evaluate it using the target language $\ell_2$, the updated weight matrix $\hat{W}$ is applied to the target-language keys. We quantify the magnitude of the update signal that successfully survives this language barrier using the key-space propagation scalar.

\begin{definitionbox}
\textbf{Definition 1 (Key-Space Propagation Scalar):} The proportion of the source update signal that structurally propagates to the target-language representation.
\begin{itemize}[leftmargin=*, itemsep=2pt, nosep]
    \item \textbf{ROME ($\alpha_{1,2}$):} A scalar defined as $\alpha_{1,2} \coloneqq \frac{(\bm{k}_1^*)^\top C_1^{-1}\,\bm{k}_2^*}{(\bm{k}_1^*)^\top C_1^{-1}\,\bm{k}_1^*}$.
    \item \textbf{MEMIT ($\bm{A}_{1,2}$):} Generalizes to the batch alignment matrix $\bm{A}_{1,2} \coloneqq K_1^\top(C_1 + K_1 K_1^\top)^{-1}K_2 \in \mathbb{R}^{n \times n}$, where the mean self-propagation is $\bar{\alpha}_{\text{batch}} \coloneqq \frac{1}{n}\text{Tr}(\bm{A}_{1,2})$.
    \item \textbf{AlphaEdit ($\bm{A}_{1,2}^{\text{AE}}$):} Incorporates the null-space projection matrix $P_\perp$ via right-multiplication. This acts directly on the key space by projecting the target keys out of the dense source memory space, yielding the constrained alignment matrix $\bm{A}_{1,2}^{\text{AE}} \coloneqq K_1^\top(C_1 + K_1 K_1^\top)^{-1} (P_\perp K_2)$, with mean propagation $\bar{\alpha}_{\text{batch}}^{\text{AE}} \coloneqq \frac{1}{n}\text{Tr}(\bm{A}_{1,2}^{\text{AE}})$.
\end{itemize}
\end{definitionbox}

This propagation term mathematically depends on the alignment of the keys under the Mahalanobis-type inner product induced by $C_1^{-1}$. Applying the pseudo-whitening transform $\tilde{\bm{k}} = C_1^{-1/2}\bm{k}$, we can rewrite the ROME scalar as the standard Euclidean projection in the whitened space: 
$
    \alpha_{1,2} = \frac{\tilde{\bm{k}}_1^{*\top}\tilde{\bm{k}}_2^*}{\tilde{\bm{k}}_1^{*\top}\tilde{\bm{k}}_1^*}
$.

Assuming the cross-lingual key shift can be approximated by a linear transformation $\bm{k}_2^* = A_{1,2}\,\bm{k}_1^*$: a reasonable population-level approximation since MLP hidden states for semantically equivalent queries tend to lie in locally linear regimes (detailed in Appendix~\ref{app:proof_rayleigh}), we define the whitened transfer matrix $M_{1,2} \coloneqq C_1^{-1/2}A_{1,2}C_1^{1/2}$. This allows us to bound the propagation analytically.

\begin{proposition}[Rayleigh Bounds on Propagation]\label{prop:rayleigh}
    The propagation scalar $\alpha_{1,2}$ corresponds to the Rayleigh quotient $R(M_{1,2}, \tilde{\bm{k}}_1^*)$. By the min-max characterization applied to the symmetric part $\frac{1}{2}(M_{1,2} + M_{1,2}^\top)$, it is strictly bounded by its minimum and maximum eigenvalues, denoted as $\lambda_{\min}$ and $\lambda_{\max}$ respectively:
    \begin{equation}
    \lambda_{\min}\left(\frac{M_{1,2} + M_{1,2}^\top}{2}\right) \leq \alpha_{1,2} \leq \lambda_{\max}\left(\frac{M_{1,2} + M_{1,2}^\top}{2}\right)
    \end{equation}
\end{proposition}

\noindent The formal proof is provided in Appendix~\ref{app:proof_rayleigh}. Differences between the source- and target-language second-moment matrices ($C_1 \neq C_2$) typically result in weaker key-space propagation, as suggested by the experimental observations below.

\begin{figure}[!t]
    \centering
    \begin{subfigure}{0.32\textwidth}
        \centering
        \includegraphics[width=\linewidth]{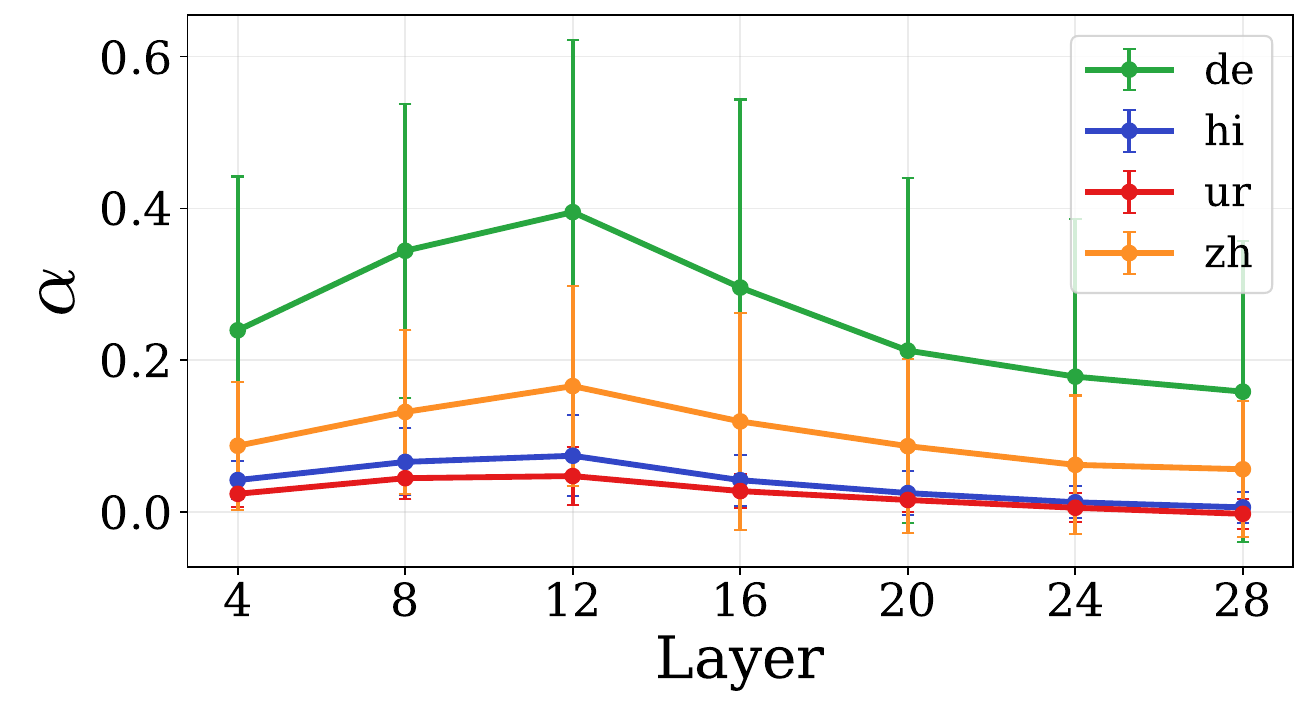}
        \caption{ROME}
        \label{fig:alpha_from_cases_by_language_llama_rome}
    \end{subfigure}
    \hfill
    \begin{subfigure}{0.32\textwidth}
        \centering
        \includegraphics[width=\linewidth]{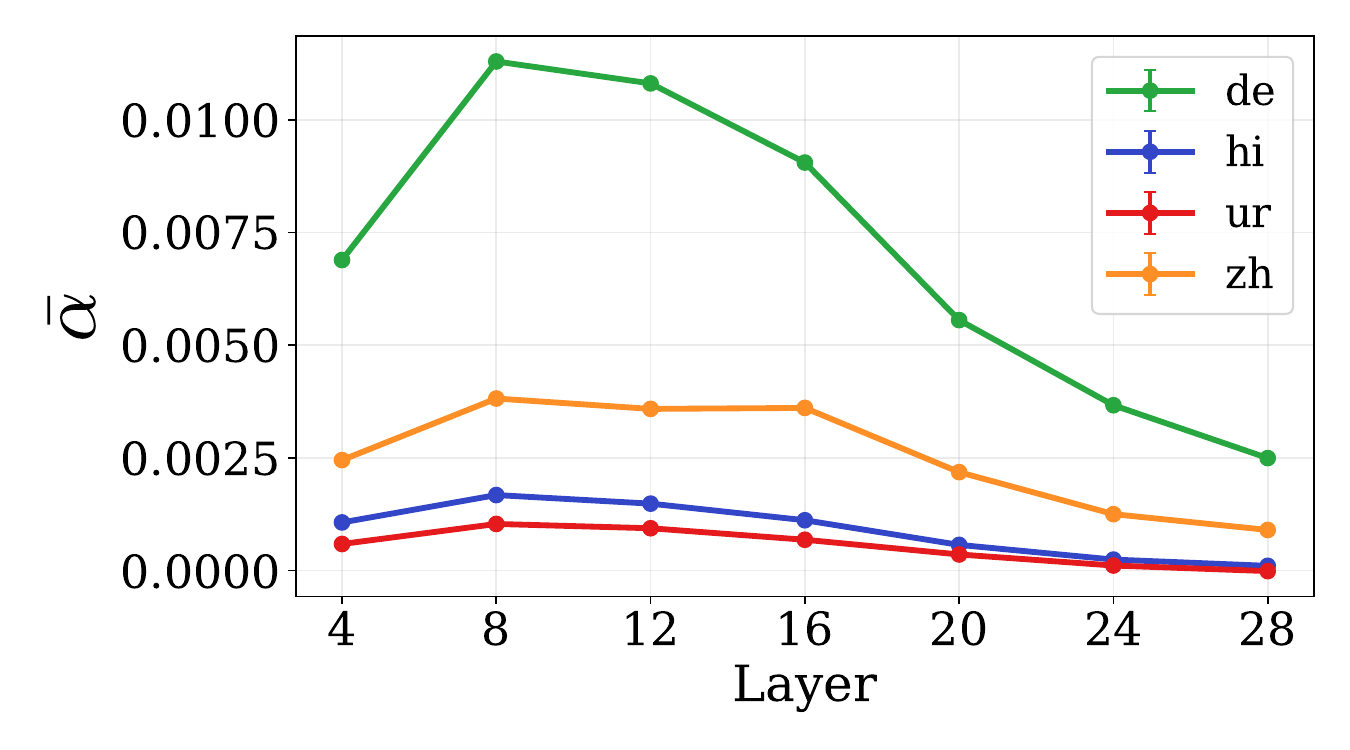}
        \caption{MEMIT}
        \label{fig:alpha_from_cases_by_language_llama_memit}
    \end{subfigure}
    \hfill
    \begin{subfigure}{0.32\textwidth}
        \centering
        \includegraphics[width=\linewidth]{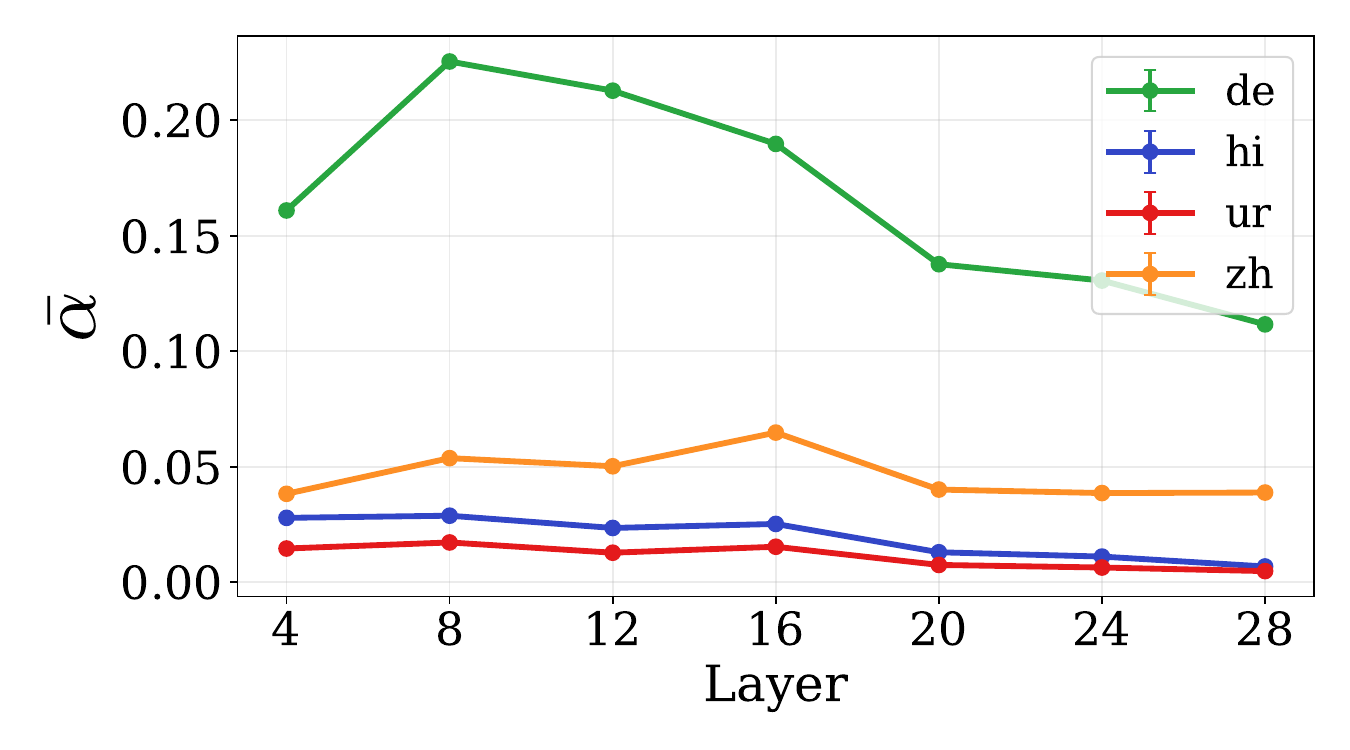}
        \caption{AlphaEdit}
        \label{fig:alpha_from_cases_by_language_llama_alphaedit}
    \end{subfigure}
    \caption{\textbf{Key-space propagation ($\alpha$) across editing methods (Llama-3.1-8B).} The propagation magnitude follows a typologically stratified pattern. German ($\alpha_{\text{de}}$) peaks at early-to-mid layers, while distant languages are uniformly suppressed across ROME, MEMIT, and AlphaEdit. Corresponding Gemma-3-4B traces are shown in Figure~\ref{fig:alpha_from_cases_by_language_gemma} of Appendix~\ref{app:extended_empirical}.}
    \label{fig:alpha_from_cases_by_language_llama}
\end{figure}

\begin{figure}[!t]
    \centering
\includegraphics[width=1\linewidth]{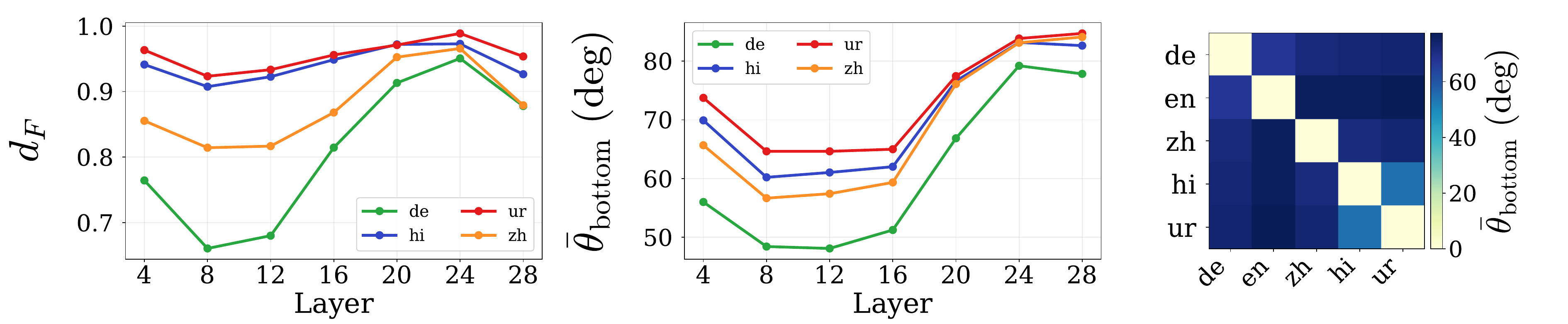}
    \caption{\textbf{Key-space covariance geometry across layers (Llama-3.1-8B).} Structural divergence is stratified by typological distance in the evaluated languages. (a)~Frobenius divergence and (b)~principal angles demonstrate that distant languages compress keys into misaligned, near-orthogonal subspaces ($>75^\circ$). (c)~Layer 20 heatmap confirms this misalignment is not English-centric. Gemma-3-4B results are detailed in Figure~\ref{fig:gemma_C_matrix_geomtry} of Appendix~\ref{app:extended_empirical}.}
    \label{fig:plot_triptych_fro_divergence_bottom_angle_k50_from_en_layer_20_llama}
\end{figure}

Empirically, this theoretical bound manifests as a language-stratified bottleneck (Figure~\ref{fig:alpha_from_cases_by_language_llama}). Across all methods, German ($\alpha_{1,\text{de}}$) peaks at ${\sim}0.4$ near layer 12, reflecting an early-to-mid layer subspace where English and German key geometries remain reasonably aligned. However, for distant languages like Hindi and Urdu, propagation remains uniformly suppressed ($< 0.08$) across all layers. 
This non-monotone, typological ordering (\texttt{de} $\gg$ \texttt{zh} $>$ \texttt{hi} $\gtrsim$ \texttt{ur}) is consistent with the
underlying covariance geometry shown in Figure~\ref{fig:plot_triptych_fro_divergence_bottom_angle_k50_from_en_layer_20_llama}.
Both the Frobenius divergence $d_F(C_{\text{en}}, C_j)$ and the principal angles of
the bottom eigenspaces follow a similar hierarchy, suggesting that distant languages compress key
representations into misaligned, nearly orthogonal subspaces, thereby suppressing the update signal
early in the forward pass.

\paragraph{Generalization to Batch Editing.} When moving to $n$ simultaneous edits, the batch alignment matrix $\bm{A}_{1,2}$ reveals an additional MEMIT-specific failure mode. We find that batching actively exacerbates the key-space bottleneck for distant languages by introducing substantial cross-fact leakage. The interference ratio ($\text{IR} \coloneqq \|\text{off-diag}(\bm{A}_{1,2})\|_F / \|\text{diag}(\bm{A}_{1,2})\|_F$) spikes to $8$--$20$ for Indic languages (compared to $4$--$5$ for German), meaning the weak cross-lingual signal that does survive is affected by off-diagonal interference from other misaligned facts in the batch.

\begin{takeaway}
\textbf{Key-space propagation is bottlenecked by the geometric divergence of the models' covariance matrices}
($C_1 \neq C_2$). Typological distance is associated with reduced signal magnitude across the language barrier, with b\textbf{atch editing compounding this failure via cross-fact interference}.
\end{takeaway}

%% file: sections/geometry_residual.tex
\subsection{Orthogonal Mismatch as a Value-Space Barrier}
\label{sec:value_space_barrier}

Even if a signal perfectly navigates the key-space gate ($\alpha_{1,2} \approx 1$), it must still align with the target language's optimal value-space geometry. We quantify this value-space alignment by defining the cross-lingual value gap, which measures the difference between the required target-language update and the actual signal delivered by the source-language edit.

\begin{definitionbox}
\label{def:gap}
\textbf{Definition 2 (Cross-Lingual Value Gap):} The exact mismatch in the edit residual at the edited layer when an update optimized in a source language is evaluated zero-shot in a target language.
\begin{itemize}[leftmargin=*, itemsep=2pt, nosep]
    \item \textbf{ROME ($\bm{g}_{1,2}$):} The vector gap defined as $\bm{g}_{1,2} \coloneqq \bm{v}_2^* - \hat{W}\bm{k}_2^* = \bm{r}_2 - \alpha_{1,2}\bm{r}_1$ (proof in Appendix~\ref{app:proof_gap}).
    \item \textbf{MEMIT ($\bm{G}_{1,2}$):} Generalizes to the batch residual gap $\bm{G}_{1,2} \coloneqq V_2^L - \hat{W}^L K_2^L = R_2 - R_1\bm{A}_{1,2}$ (\ref{app:memit_derivation}).
    \item \textbf{AlphaEdit ($\bm{G}_{1,2}^{\text{AE}}$):} Shares the batch residual structure, but filtered through the null-space constrained alignment matrix: $\bm{G}_{1,2}^{\text{AE}} \coloneqq R_2 - R_1\bm{A}_{1,2}^{\text{AE}}$ (\ref{app:alphaedit_derivation}).
\end{itemize}
\end{definitionbox}

To isolate the failure mode analytically, we decompose the required target residual $\bm{r}_2$ into a component parallel to the source residual $\bm{r}_1$ and its orthogonal complement: $\bm{r}_2 = \gamma\bm{r}_1 + \bm{r}_2^{\perp}$, where $\gamma \coloneqq (\bm{r}_2 \cdot \bm{r}_1)/\|\bm{r}_1\|^2$ and $\bm{r}_2^{\perp} \perp \bm{r}_1$.

\begin{proposition}[Parallel-Orthogonal Decomposition]\label{prop:decomp}
    Normalizing by $\|\bm{r}_1\|^2$, the relative squared value gap decomposes exactly into a scalar mismatch and a directional mismatch:
    \begin{equation}\label{eq:gapnorm_normalized}
        \frac{\|\bm{g}_{1,2}\|^2}{\|\bm{r}_1\|^2} = \underbrace{(\gamma - \alpha_{1,2})^2}_{\text{parallel mismatch}} + \underbrace{\frac{\|\bm{r}_2^{\perp}\|^2}{\|\bm{r}_1\|^2}}_{\text{orthogonal mismatch}}
    \end{equation}
\end{proposition}

\begin{figure}[h]
    \centering
    \includegraphics[width=1\linewidth]{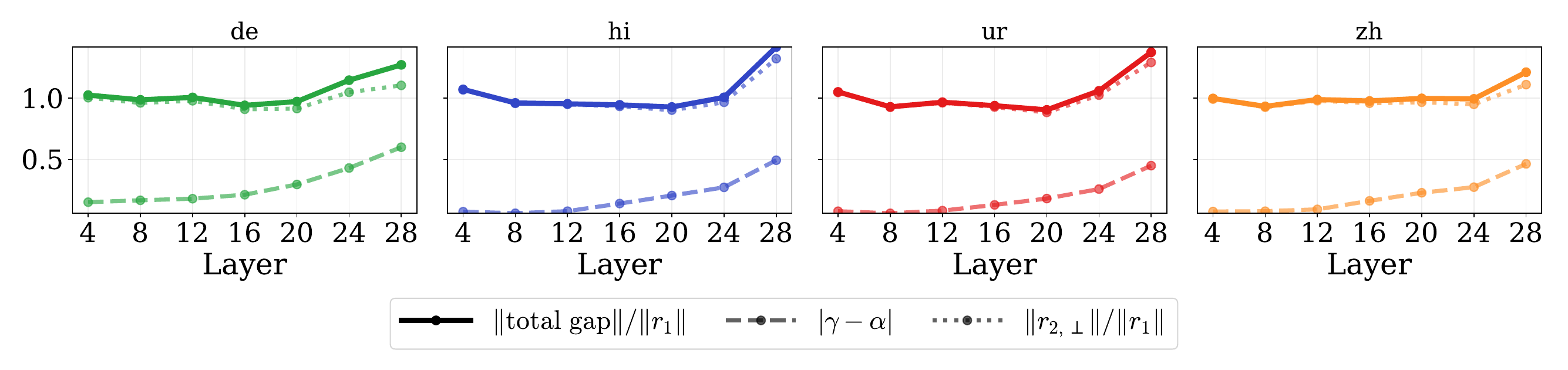}
    \caption{\textbf{Per-language gap decomposition across layers (ROME, Llama-3.1-8B).} The relative total gap $\|\bm{g}_{1,2}\|/\|\bm{r}_1\|$ (solid) is largely dominated by the orthogonal residual mismatch $\|\bm{r}_2^{\perp}\|/\|\bm{r}_1\|$
(dotted) in the evaluated language pairs. The parallel mismatch (dashed) acts as a visible secondary error only for typologically close languages (German). Gemma-3-4B results are shown in Figure~\ref{fig:gemma_residual_gap} of Appendix~\ref{app:extended_empirical}.}
    \label{fig:gap_decomp_lang_wise}
\end{figure}

\noindent The formal proof is detailed in Appendix~\ref{app:proof_decomp}.  
Extracting these independent components across layers
(Figure~\ref{fig:gap_decomp_lang_wise}) reveals a consistent pattern in our evaluated settings: the orthogonal mismatch dominates
the transfer error across language pairs.

Empirically, this decomposition suggests that the update directions required for equivalent
multilingual facts are substantially language-specific. At the optimal editing layers -- the ``semantic valley'' between layers 16 and 20 -- the relative gap norm reaches its minimum, yet the orthogonal component $\bm{r}_2^{\perp}$ still accounts for ${\sim}0.90$ of the total error. The target residual simply points in a direction that the source residual cannot reach. 

Crucially, while distance modulates the key-space scalar $\alpha_{1,2}$ (which dictates the severity of the secondary parallel mismatch, as seen with German in Figure~\ref{fig:gap_decomp_lang_wise}), the orthogonal mismatch remains large across the evaluated layers. Because methods like MEMIT and AlphaEdit compute updates bounded by the span of source-language residuals ($R_1$), they cannot supply target-specific directions outside this span. Thus, part of the target-language update remains inaccessible to a source-only residual update.

\paragraph{Is the mismatch an artifact of the edit layer?} The layer at which a fact is localized is not necessarily the best layer at which to edit it~\cite{hase2023localization}. We therefore selected, separately for each of 82 facts, the layer in $\{4, 8, \dots, 28\}$ that minimizes $\|\mathbf{r}_2^{\perp}\|/\|\mathbf{r}_1\|$. This per-fact oracle uses target-language information and is therefore optimistic. It reduces the normalized orthogonal mismatch in every language (by 0.09--0.14; paired $t$-tests, $p < 10^{-10}$), but the mismatch remains between 0.77 and 0.80, and the best layer varies widely across facts (Appendix~\ref{app:oracle-layer}). The mismatch is therefore not explained by a single poorly chosen edit layer.

\begin{takeaway}
Within associative-memory editing, \textbf{the transferred update is constrained by the source-language residual direction or residual span}. When the target-language residual has a large orthogonal component, this component cannot be supplied by a source-only edit. Empirically, this orthogonal component dominates the value-space gap across the evaluated models, languages, and editing layers.
\end{takeaway}

%% file: sections/geometry_transfer.tex
\subsection{Downstream Transfer Penalties Reflect Internal Misalignment}
\label{sec:downstream_penalty}

While the value gap ($\bm{g}_{1,2}$) defines the exact mathematical error at the edited layer $l^*$, we next examine
whether this internal geometry is reflected in observable model behavior. We map this gap to the output logits $\mathbf{y} \in \mathbb{R}^{|V|}$ in two stages. Let $\mathbf{z} \in \mathbb{R}^{d_{\mathrm{out}}}$ be the value representation at the edited layer $l^*$ and $\mathbf{h}(\mathbf{z}) \in \mathbb{R}^{d_{\mathrm{out}}}$ the final hidden state, so that $\mathbf{y} = W_U\,\mathbf{h}(\mathbf{z})$ with unembedding matrix $W_U \in \mathbb{R}^{|V| \times d_{\mathrm{out}}}$ (we absorb the local Jacobian of the final normalization layer into $W_U$). We write $J_h = \partial \mathbf{h} / \partial \mathbf{z} \in \mathbb{R}^{d_{\mathrm{out}} \times d_{\mathrm{out}}}$ for the downstream Jacobian.

To quantify this behaviorally, we define the performance drop between native and cross-lingual editing.

\begin{definitionbox}
\label{def:penalty}
\textbf{Definition 3 (Cross-Lingual Transfer Penalty):} The difference in editing success between native target-language editing and zero-shot cross-lingual transfer. Let $\text{NLL}^{(0)}$ be the pre-edit negative log-likelihood of the target answer $o_2$, and $\text{NLL}^{(\text{post})}$ be the post-edit NLL. Defining the reduction in NLL as $\Delta\text{NLL} \coloneqq \text{NLL}^{(0)} - \text{NLL}^{(\text{post})}$, the penalty is:
$
    \mathcal{P} \coloneqq \Delta\text{NLL}_{\text{self}} - \Delta\text{NLL}_{\text{en}}
$, where $\Delta\text{NLL}_{\text{self}}$ is the NLL reduction when editing natively in $\ell_2$, and $\Delta\text{NLL}_{\text{en}}$ is the reduction when editing in English and evaluating in $\ell_2$.
\end{definitionbox}

A higher positive penalty ($\mathcal{P}$) indicates that native editing is more effective than cross-lingual transfer.

\begin{proposition}[Downstream Logit Gap \& Transfer Penalty Bound]\label{prop:logit}
To a first-order Taylor approximation, the downstream logit error is the value gap mapped through the rest of the network: $\Delta\mathbf{y}_{\mathrm{gap}} \approx W_U J_h\,\mathbf{g}_{1,2}$. Consequently, the transfer penalty is approximated by this mapped gap:
\begin{equation}
P \approx \big(\mathbf{e}_{o_2} - \mathbf{q}^{(0)}\big)^{\top} W_U J_h\,\mathbf{g}_{1,2}
\end{equation}
where $\mathbf{q}^{(0)}$ is the pre-edit predictive distribution and $\mathbf{e}_{o_2}$ is the target one-hot vector. Their difference $(\mathbf{e}_{o_2} - \mathbf{q}^{(0)})$ geometrically defines the \emph{error direction} correcting the initial belief.
\end{proposition}

\noindent As derived in Appendix~\ref{app:penalty_proof}, because residual connections dominate the forward pass from mid-to-late layers ($J_h \approx I$), applying the Cauchy--Schwarz inequality yields a first-order \emph{upper} bound in terms of the gap norm: $|P| \lesssim \sqrt{2}\,\|W_U\|_2\,\|\mathbf{g}_{1,2}\|$. A small gap therefore implies a small penalty, but not conversely, so we do not claim a lower bound on $P$; whether the internal gap is reflected in behavior is examined empirically below.

\begin{figure}[h]
    \centering
    \begin{minipage}{0.5\textwidth}
        \centering
        \includegraphics[width=0.9\linewidth]{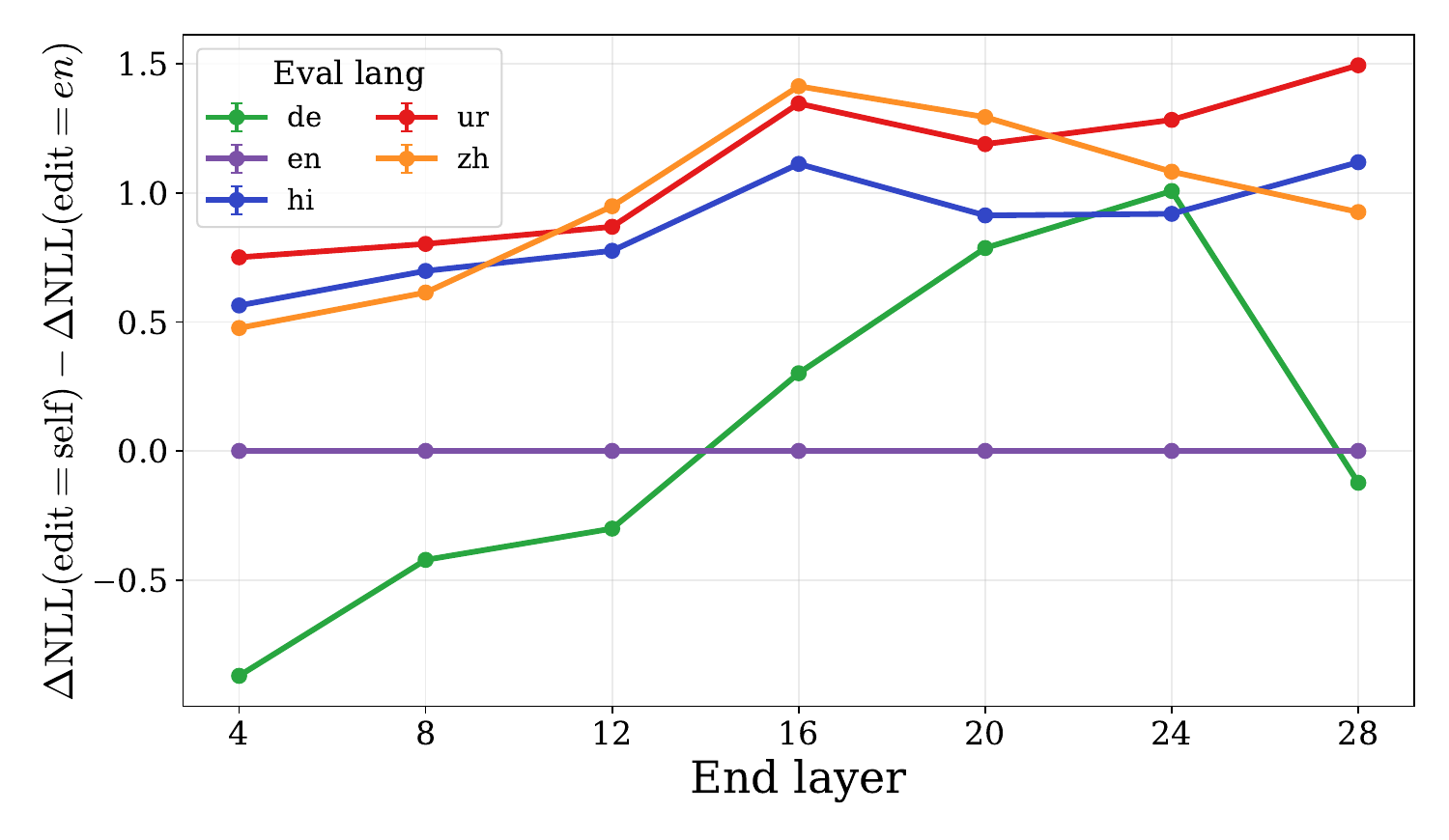}
        \captionof{figure}{\textbf{Cross-lingual transfer penalty (Llama-3.1-8B).} Positive values denote self-editing outperforms transfer. Distant languages suffer severe penalties peaking in the semantic valley.}
        \label{fig:nll_gap}
    \end{minipage}\hfill
    \begin{minipage}{0.45\textwidth}
        \centering
        
        \begin{tabular}{lcc}
            \toprule
            Lang Pair & Spearman $\rho$ & Pearson $r$ \\
            \midrule
            en $\to$ de & 0.221 & 0.152 \\
            en $\to$ hi & 0.119 & 0.098 \\
            en $\to$ ur & 0.198 & 0.178 \\
            en $\to$ zh & 0.114 & 0.099 \\
            \bottomrule
        \end{tabular}
        \captionof{table}{\textbf{Correlation between geometric gap and penalty.} Correlation between the relative gap norm $\|\bm{g}_{1,2}\|/\|\bm{r}_1\|$ and $\mathcal{P}$ across layers 12--24 ($p < 10^{-4}$).}
        \label{tab:corr_gap_nll}
    \end{minipage}
\end{figure}

Empirically (Figure~\ref{fig:nll_gap}), the transfer penalty follows the qualitative structure predicted by our decomposition. While German occasionally exhibits a negative penalty at early layers, distant languages suffer persistent penalties that peak in the ``semantic valley'' (layers 16--20). Furthermore, Table~\ref{tab:corr_gap_nll} shows statistically reliable but modest positive correlations between the relative gap norm and the transfer penalty, supporting our geometric decomposition as a useful population-level diagnostic for cross-lingual failure.

\begin{takeaway}
The internal geometric gap is behaviorally meaningful: it is associated with the observed penalty of cross-lingual transfer. The relationship is not one-to-one, since downstream decoding, tokenization, and layer interactions also affect final predictions, but the gap provides a useful diagnostic for transfer failure.
\end{takeaway}

%% file: sections/impossibility.tex
\section{A Conditional Limit for Source-Only Residual Correction}
\label{sec:impossibility}

Section~\ref{sec:value_space_barrier} shows that the cross-lingual gap is dominated by the orthogonal residual mismatch
($\bm{r}_2^\perp$). We now ask: could a correction strategy -- operating only with source-language data at test
time -- close this gap? We show that, under a source-residual-only correction model, the answer is quantitatively bounded.

\subsection{Reduction to a Residual Prediction Problem}
\label{sec:correction_problem}

At test time, a practitioner has access to the translated key $\bm{k}_2^*$ and the source-language edit components ($\bm{k}_1^*, \bm{v}_1^*, \bm{r}_1, \alpha_{1,2}$). Crucially, the target value $\bm{v}_2^*$ and residual $\bm{r}_2$ remain unobservable without native $\ell_2$ optimization. 
Modifying the covariance matrix $C$ only alters the scalar projection $\alpha_{1,2}$, which leaves the dominant orthogonal error $\|\mathbf{r}_2^{\perp}\|^2$ untouched, and even a per-fact oracle choice of the editing layer leaves most of it in place (Appendix~\ref{app:oracle-layer}). Therefore, addressing the structural gap requires replacing $\bm{r}_1$ with a corrected vector $f(\bm{r}_1)$ that delivers the proper signal to $\ell_2$. 
Under this source-residual-only correction model, reducing the gap amounts to learning a correction
function $f : \mathbb{R}^{D} \to \mathbb{R}^{D}$ such that $\alpha_{1,2}f(\bm{r}_1) \approx \bm{r}_2$ for unseen facts.

\subsection{The Private Fraction of Multilingual Residuals}
\label{sec:predictability}

The tractability of this restricted correction depends on how much variance in $r_2$ is predictable from $r_1$. 
We quantify this opacity via the \textbf{private fraction}:
\begin{equation}\label{eq:eta_generic}
    \eta_{1,2} \coloneqq \eta(\bm{r}_2 \mid \bm{r}_1) \coloneqq 1 - R^2_{\mathrm{pop}}(\bm{r}_2 \mid \bm{r}_1)
\end{equation}
where $R^2_{\mathrm{pop}}$ is the population coefficient of determination for the optimal nonlinear regression. 
When $\eta_{1,2} \approx 1$, predictors based only on the source residual explain little of the target residual variance. We estimate $\hat{\eta} = 1 - R^2_{\mathrm{best}}$ using Ridge (linear), two-layer MLP (nonlinear), and $k$-NN (nonparametric) estimators (Appendix~\ref{app:estimator_details}).
Applying this to the ROME components (Figure~\ref{fig:eta_plot_layer20_llama}) reveals a severe bottleneck. While keys are moderately predictable, target values and residuals are poorly predicted from source residuals ($\hat{\eta} \geq 0.96$).
Because $\bm{k}_2^*$ is observable, the unpredictability of $\bm{r}_2$ is inherited entirely from the value space ($\bm{v}_2^*$), driven by language-specific optimization landscapes and independent pre-existing model confidences.

\subsection{A Source-Only Residual Prediction Barrier}
\label{sec:floor}

This private fraction yields a lower bound for correction functions that use only the source residual. The optimal source-residual predictor $\mathbb{E}[\bm{r}_2 \mid \bm{r}_1]$ leaves behind the conditional variance:
\begin{equation}\label{eq:sigma}
    \sigma_{1,2}^2 \coloneqq \mathbb{E}\left[\left\|\bm{r}_2 - \mathbb{E}[\bm{r}_2 \mid \bm{r}_1]\right\|^2\right]
\end{equation}
By the law of total variance (Appendix~\ref{app:private_fraction}), this evaluates exactly to $\sigma_{1,2}^2 = \eta_{1,2} \cdot \mathrm{Var}(\bm{r}_2)$.

\begin{proposition}[Source-Only Residual Prediction Barrier]\label{prop:floor}
For any measurable correction function $f : \mathbb{R}^D \to \mathbb{R}^D$, the expected squared value gap is lower-bounded as
\begin{equation}
\mathbb{E}_{\mathrm{facts}}\big[\|\mathbf{r}_2 - \alpha_{1,2} f(\mathbf{r}_1)\|^2\big] \;\ge\; \eta(\mathbf{r}_2 \mid \mathbf{r}_1, \alpha_{1,2}) \cdot \operatorname{Var}(\mathbf{r}_2),
\end{equation}
where $\eta(\mathbf{r}_2 \mid \mathbf{r}_1, \alpha_{1,2}) \le \eta_{1,2}$ is the private fraction of Eq.~(\ref{eq:eta_generic}) conditioned on both $\mathbf{r}_1$ and the fact-specific scalar $\alpha_{1,2}$. If $\alpha_{1,2}$ carries no information about $\mathbf{r}_2$ beyond $\mathbf{r}_1$, the right-hand side equals the target residual's conditional variance $\eta_{1,2} \cdot \operatorname{Var}(\mathbf{r}_2)$.
\end{proposition}
The formal proof is provided in Appendix~\ref{app:impossibility_proofs}. This shows that, for correction functions whose input is restricted to the source residual, the remaining error is governed by the conditional variance of the target residual. Predictors that also use the keys, the prompts, language identity, or other internal states are not covered by this bound and may do better.

\begin{figure}[t]
    \centering
    \begin{minipage}{0.48\textwidth}
        \centering
        \includegraphics[width=\linewidth]{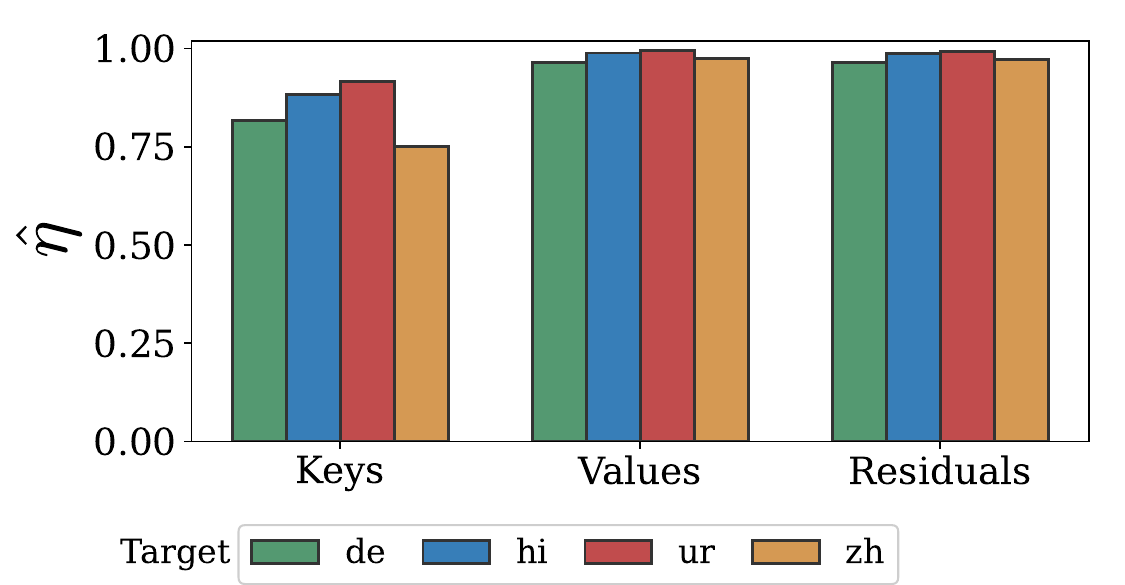}
        \captionof{figure}{\textbf{Private fraction ($\hat{\eta}$) at layer 20 (Llama-3.1-8B).} 
        While keys are more predictable, target values and residuals remain poorly predicted from source residuals across the evaluated languages.}
        \label{fig:eta_plot_layer20_llama}
    \end{minipage}
    \hfill
    \begin{minipage}{0.48\textwidth}
        \centering
        \small
        \begin{tabular}{lcccc}
        \toprule
        Lang Pair & $R^2_{\text{Ridge}}$ & $R^2_{\text{MLP}}$ & $R^2_{\text{kNN}}$ & $\hat{\eta}$ \\
        \midrule
        en $\to$ de & \textbf{0.0363} & 0.0322 & 0.0359 & 0.9637 \\
        en $\to$ hi & 0.0114 & 0.0085 & \textbf{0.0123} & 0.9877 \\
        en $\to$ ur & 0.0049 & \textbf{0.0079} & 0.0054 & 0.9921 \\
        en $\to$ zh & \textbf{0.0281} & 0.0244 & 0.0227 & 0.9719 \\
        \bottomrule
        \end{tabular}
        \vspace{0.5em}
        \captionof{table}{
        \textbf{Residual predictability (Llama-3.1-8B, layer 20).} Across three estimators, source residuals explain little target-residual variance, leaving ${\sim}96\%$ unexplained under the evaluated source-only predictors.}
        \label{tab:eta_table_layer20_llama}
    \end{minipage}
\end{figure}

Table~\ref{tab:eta_table_layer20_llama} reports the per-estimator $R^2$ at the selected editing layer. Across linear, nonlinear, and nonparametric estimators, the source residual explains little target-residual variance ($R^2 \leq 0.036$). This suggests that source residuals alone provide limited information about the target-language residual, making post-hoc source-only correction unlikely to fully close the transfer gap without additional target-language or representation-alignment signals.

\begin{takeaway}
Target-language residuals are poorly predicted from source residuals alone, suggesting a large practical barrier for source-only correction and motivating representation realignment or explicit multilingual edit information.
\end{takeaway}

%% file: sections/alignment.tex
\section{Contrastive Alignment Opens the Key-Space Gate}
\label{sec:interventions}

Since Section~\ref{sec:impossibility} identifies a source-only residual prediction barrier, we ask: can the gap be narrowed \emph{before} editing?
Our decomposition localizes the relative gap minimum to the semantic valley (layers 16--20 for Llama-3.1-8B; Section~\ref{sec:value_space_barrier}). We hypothesize that aligning latent representations across languages in and around this region improves zero-shot transfer.

\paragraph{Method.}
We perform supervised fine-tuning at layers 16--25, optimizing $\mathcal{L}_{\text{total}} = \mathcal{L}_{\text{contrast}} + \beta\,\mathcal{L}_{\text{LM}}$. The contrastive term aligns MLP hidden states ($\bm{h}$) at the final subject token across a batch of $B$ parallel zsRE facts via InfoNCE~\citep{oord2019representationlearningcontrastivepredictive}: $\mathcal{L}_{\text{contrast}} = -\frac{1}{B}\sum_{i=1}^{B} \log \bigl[\exp(\mathrm{sim}(\bm{h}_{\ell_1}^{(i)}, \bm{h}_{\ell_2}^{(i)})/\tau) / \sum_{j=1}^{B} \exp(\mathrm{sim}(\bm{h}_{\ell_1}^{(i)}, \bm{h}_{\ell_2}^{(j)})/\tau)\bigr]$, using cosine similarity and $\tau = 0.07$. 
The regularizer $\mathcal{L}_{\text{LM}}$ mitigates capability loss via Wikipedia next-token prediction; we monitor zero-shot cross-lingual performance on XNLI in Appendix~\ref{app:xnli}. We chose the window of layers 16--25 using the layer-wise diagnostics (key-space propagation in Figure~\ref{fig:alpha_from_cases_by_language_llama} and the low residual-gap region in Figure~\ref{fig:gap_decomp_lang_wise}), comparing candidate windows on representation alignment and post-edit transfer while monitoring language-modeling loss and downstream performance. Further details are provided in Appendix~\ref{app:contrastive_full}.

\paragraph{Results.}
Table~\ref{tab:alignment} reveals an important asymmetry between the two bottlenecks on CounterFact evaluations. The key-space propagation scalar $\alpha_{1,2}$ increases dramatically, indicating that target-language keys become more reachable from the source-language edit geometry. However, the orthogonal value-space mismatch ($\|\bm{r}_2^\perp\|/\|\bm{r}_1\|$) remains high and in some cases increases. This supports the central diagnostic claim of our decomposition: key-space alignment is necessary for transfer but not sufficient. The downstream effects are mixed: efficacy improves for Urdu and Chinese, is essentially unchanged for Hindi, and decreases for German, and results for other methods and for Gemma-3-4B are similarly mixed. Alignment also reduces XNLI accuracy by 4--8 points in every language (Table~\ref{tab:xnli}). The remaining value-space mismatch continues to limit cross-lingual efficacy.

\begin{table}[h]
    \centering
    \small
    \setlength{\tabcolsep}{4pt}
    \begin{tabular}{lcccccccc}
    \toprule
    & \multicolumn{2}{c}{$\alpha_{1,2}$ $\uparrow$} & \multicolumn{2}{c}{$\gamma$} & \multicolumn{2}{c}{$\|\bm{r}_2^\perp\|/\|\bm{r}_1\|$ $\downarrow$} & \multicolumn{2}{c}{Efficacy ($\uparrow$)} \\
    \cmidrule(lr){2-3} \cmidrule(lr){4-5} \cmidrule(lr){6-7} \cmidrule(lr){8-9}
    Pair & Base & Aligned & Base & Aligned & Base & Aligned & Base & Aligned \\
    \midrule
    en $\to$ de & 0.212 & \textbf{0.858} & 0.470 & 0.486 & \textbf{0.913} & 0.861 & \textbf{72.62} & 61.17 \\
    en $\to$ hi & 0.025 & \textbf{0.618} & 0.231 & 0.261 & \textbf{0.901} & 0.952 & 49.61 & \textbf{50.19} \\
    en $\to$ ur & 0.015 & \textbf{0.517} & 0.197 & 0.223 & \textbf{0.882} & 0.941 & 46.70 & \textbf{59.22} \\
    en $\to$ zh & 0.086 & \textbf{0.579} & 0.297 & 0.302 & 0.966 & \textbf{0.960} & 33.59 & \textbf{39.51} \\
    \bottomrule
    \end{tabular}
    \vspace{0.5em}
    \caption{\textbf{Effect of contrastive alignment (Llama-3.1-8B, ROME, layer 20).} The intervention dramatically improves key-space propagation ($\alpha_{1,2}$), but the orthogonal value-space mismatch persists. \emph{Base Efficacy reflects CounterFact metrics from Table~\ref{tab:motivation_gap}.} Results for other methods and Gemma-3-4B are presented in Tables~\ref{tab:sft_vs_base_nll} and \ref{tab:sft_vs_base_efficacy}.}
     \vspace{-0.5em}
    \label{tab:alignment}
\end{table}

%% file: sections/disc_and_conc.tex
\section{Conclusion}
\label{sec:discussion}

In this work, we studied why factual edits made in one language often fail to transfer to others using associative-memory-based editing methods in LMs. Our analysis identifies two distinct sources of failure: the update may have little effect on the target-language representation, or it may change that representation in a direction different from what a native-language edit would require. We find that the second problem persists across languages and editing layers, even when the layer is chosen separately for each fact. Our experiments further show that source-language residuals alone provide limited information for predicting the updates required in other languages. Aligning representations across languages improves the propagation of edits, but does not consistently improve editing success and can affect the downstream performance. These findings indicate that better cross-lingual editing requires more than aligning multilingual representations: editing methods must also account for differences in the updates required across languages. Our results characterize these limitations for associative-memory-based methods, while leaving open how other editing approaches might overcome them.

\section{Limitations and Broader Impact}
\label{sec:limitations}

Our analysis is limited to associative-memory-based editors (ROME, MEMIT, and AlphaEdit) and does not necessarily extend to methods with different update mechanisms. The identified bottlenecks neither explain all causes of cross-lingual editing failure nor rule out shared multilingual representations~\cite{fierro-etal-2025-multilingual}. Furthermore, the layers modified by editing methods need not be where factual information is retrieved~\cite{hase2023localization}. Our theoretical analysis assumes an approximately linear cross-lingual key mapping and uses first-order approximations; Proposition~\ref{prop:logit} provides only an upper bound on the behavioral transfer penalty, while Proposition~\ref{prop:floor} applies only to source-residual-based corrections. Empirically, we evaluate two models (4B--8B), estimate residual predictability using a limited set of predictors, and focus on short factual answers. Although the mismatch persists across the target lengths tested in zsRE (Appendix~\ref{app:target-length}), its behavior in long-form generation remains unknown~\cite{rosati-etal-2024-long}. We also leave the grammatical and linguistic side effects of editing for future work~\cite{niu2024knowledge}, as our experiments on logically equivalent formulations (Appendix~\ref{app:equivalent-formulations}) do not examine them.

Our findings highlight a practical risk in multilingual model editing: a factual correction made in one language may not transfer to others, potentially leaving users with inconsistent information. This is particularly concerning for safety-related edits, where a correction effective in one language may fail in another. Although we do not directly evaluate safety interventions, our results motivate testing the effectiveness and consistency of edits across the languages in which a model is deployed, rather than relying on success in a single language.

%% file: sections/appendix_toc.tex
\section*{Appendix Contents}
\label{app:toc}

Below we provide an overview of the appendix. These sections are intended to support the core claims of the main paper by providing rigorous mathematical proofs, extended empirical evaluations, and comprehensive methodological details to ensure full reproducibility.

\begin{itemize}[leftmargin=*, itemsep=3pt]

\item \hyperref[app:faq]{\textbf{Appendix A: Frequently Asked Questions (FAQs).}}  
\textit{Addresses anticipated queries regarding the mathematical framework, the interpretation of geometric misalignment, and the structural implications of our findings.}

\item \hyperref[app:methods]{\textbf{Appendix B: Detailed Breakdown of Editing Methods.}}  
\textit{Provides formal formulations for ROME, MEMIT, and AlphaEdit, including constrained optimization steps, key-space projection logic, and exact hyperparameter configurations in Table~\ref{tab:editing_hyperparams}.}

\item \hyperref[app:proofs]{\textbf{Appendix C: Formal Proofs for Geometric Dissections and Rayleigh Bounds.}}  
\textit{Contains rigorous derivations for the cross-lingual value gap, parallel-orthogonal decomposition, Rayleigh alignment bounds, and the downstream transfer penalty bound.}

\item \hyperref[app:extended_empirical]{\textbf{Appendix D: Extended Empirical Characterization of the Transfer Gap.}}  
\textit{Reports comprehensive layer-wise diagnostics for Gemma-3-4B, MEMIT/AlphaEdit batch interference ratios, the exhaustive cross-lingual $\Delta$NLL grid across all evaluated baseline models, a per-fact oracle analysis of the edit layer, the effect of target length, and logically equivalent formulations of edited facts.}

% \item \hyperref[app:impossibility_proofs]{\textbf{Appendix E: Impossibility: Proofs and Extended Experiments.}}  
% \textit{Details the optimality of conditional expectation, the law of total variance derivation for the private fraction ($\eta$), and the convergence of linear and nonlinear estimators establishing the irreducible transfer floor.}

\item \hyperref[app:impossibility_proofs]{\textbf{Appendix E: Source-Only Correction Bound: Proofs and Extended Experiments.}} Details the optimality of conditional expectation, the law of total variance derivation for the private fraction ($\eta$), and estimator-based evidence for a large source-only residual prediction barrier.

\item \hyperref[app:contrastive_full]{\textbf{Appendix F: Contrastive Alignment Experiment Details.}}  
\textit{Outlines the InfoNCE training objective, language modeling regularization to prevent catastrophic forgetting, XNLI reasoning preservation results, and perplexity monitoring used to determine the optimal training window.}

\end{itemize}

%% file: sections/appendix_faqs.tex
\section{Frequently Asked Questions (FAQs)}
\label{app:faq}

\begin{enumerate}[leftmargin=*, itemsep=6pt]

\item \textbf{How does our geometric framework relate to recent empirical observations of orthogonal subspaces?}  
While recent empirical works like CLM-Bench \cite{clmbench} have observed that cross-lingual edit vectors reside in nearly orthogonal subspaces, they primarily treat this as a descriptive symptom. Our work provides a mechanistic account of this observation. By deriving a closed-form
geometric decomposition of the update equations, we show how orthogonal residual components arise
within source-only associative-memory updates and why they limit zero-shot transfer.

\item \textbf{How can the key-space propagation scalar $\alpha_{1,2}$ be bounded by the eigenvalues of an asymmetric matrix?}  
The bound applies to the \textit{symmetric part} of the transfer matrix. Because the cross-lingual shift introduces asymmetry into the whitened transfer matrix $M_{1,2}$, its raw eigenvalues may be complex. However, as noted in Proposition~\ref{prop:rayleigh} and standard Rayleigh quotient properties for real asymmetric matrices, the scalar projection is bounded by the eigenvalues of the symmetric component: $\frac{1}{2}(M_{1,2} + M_{1,2}^\top)$.

\item \textbf{In Proposition~\ref{prop:logit}, how does the identity approximation ($J \approx I$) hold if the Jacobian maps to the vocabulary-sized logit space?}  
In Proposition~\ref{prop:logit}, $J$ does not denote the Jacobian mapping to the final $|V|$-dimensional logit vector. Rather, it denotes the Jacobian mapping the edited residual to the \textit{final hidden state} immediately preceding the unembedding matrix. Because dimensions are preserved across these hidden states ($d_{out} \to d_{out}$) and the forward pass is dominated by residual connections at mid-to-late layers, the approximation $J \approx I$ is a reasonable local approximation. We write this Jacobian as $J_h$ in Proposition~\ref{prop:logit} and keep the unembedding matrix $W_U$ explicit, which is why the bound there involves $\|W_U\|_2$.

\item \textbf{Why does Table~\ref{tab:corr_gap_nll} show a positive correlation between the geometric gap and the downstream evaluation metric?}  
A positive correlation is mathematically correct here because the downstream metric evaluated is the \textbf{transfer penalty} $\mathcal{P}$ (defined as native $\Delta\text{NLL}$ minus cross-lingual $\Delta\text{NLL}$). A higher transfer penalty indicates worse cross-lingual performance relative to native performance. Therefore, larger relative geometric gaps are expected to be associated with larger transfer penalties,
yielding a positive correlation.

\item \textbf{Is the typological ordering (\texttt{de} $\gg$ \texttt{zh} $>$ \texttt{hi} $\approx$ \texttt{ur}) contradictory to their genealogical language families?}  
While Hindi and Urdu share an Indo-European root with English and German, our typological distance metric is defined by structural and morphosyntactic divergence, not purely genealogical trees. For example, English and Chinese share strict \textit{Subject-Verb-Object (SVO)} ordering, leading to more aligned attention distributions and key-space geometries. In contrast, Hindi and Urdu are strictly \textit{Subject-Object-Verb (SOV)} and morphologically rich, causing their intermediate MLP key-spaces to geometrically diverge much earlier in the forward pass than Chinese.

\item \textbf{Does the source-only residual prediction bound in Proposition~\ref{prop:floor} assume conditional independence?}  
No. Proposition~\ref{prop:floor} conditions on both $r_1$ and $\alpha_{1,2}$, so the bound does not require conditional independence. The stronger form using $\eta_{1,2}$ applies only when $\alpha_{1,2}$ provides no additional information about $r_2$ beyond $r_1$. The bound is restricted to source-residual-based corrections; methods using keys, prompts, or other states are not covered.

\item \textbf{Does contrastive alignment mitigate MEMIT's severe cross-fact interference during batch editing?}  
Only partially. Our contrastive alignment intervention directly improves key-space propagation ($\alpha_{1,2}$) across all editing methods. However, MEMIT's cross-fact interference stems primarily from high \textbf{Interference Ratios (IR)} in the batch alignment matrix $\bm{A}_{1,2}$. Fully solving large-batch MEMIT interference requires complementary interventions that explicitly disentangle key-space representations.

\item \textbf{Could the structural realignment of mid-to-late layers induce catastrophic forgetting?}  
Per-epoch perplexity monitoring identifies an optimal training window  where alignment is maximized before catastrophic forgetting begins. Downstream evaluations on the XNLI benchmark show that contrastively aligned models lose 4--8 accuracy points relative to unaligned baselines (Table~\ref{tab:xnli}) while retaining non-trivial zero-shot performance, so better regularization is needed to avoid this cost.

\end{enumerate}

%% file: sections/appendix_temp.tex
\input{appendix/editing_methods}
\input{appendix/proofs_theory}

\input{appendix/empirical}
\input{appendix/proofs_impossibility}
\input{appendix/estimators}
\input{appendix/contrastive}

%% file: appendix/editing_methods.tex
\section{Detailed Breakdown of Editing Methods}
\label{app:methods}

\subsection{ROME: Rank-One Model Editing}
\label{app:rome}

ROME operates on the hypothesis that each MLP layer acts as a linear key--value store. For a given factual association $(s, r, o)$, the subject~$s$ is mapped to a key vector $\bm{k}^* \in \mathbb{R}^D$ (the MLP hidden-state activation averaged over diverse prefix contexts), and the desired factual completion is encoded as a target value vector $\bm{v}^* \in \mathbb{R}^H$.
 
\paragraph{Key Computation.}
Given a set of $N$ randomly sampled textual prefixes $\{x_j\}_{j=1}^{N}$, the key is computed as:
\begin{equation}
    \bm{k}^* = \frac{1}{N} \sum_{j=1}^{N} k(x_j + s), \quad \text{where } k(x) = \sigma\!\left(W_{fc}^{(l^*)} \gamma(a_{[x],i}^{(l^*)} + h_{[x],i}^{(l^*-1)})\right),
\end{equation}
where $\sigma$ is the activation function, $\gamma$ is the layer normalization, $a_{[x],i}^{(l^*)}$ is the attention output, and $h_{[x],i}^{(l^*-1)}$ is the residual stream at the subject's last token position $i$.
 
\paragraph{Target Value Computation.}
The target value $\bm{v}^*$ is obtained by optimizing:
\begin{equation}
    \bm{v}^* = \arg\min_{\bm{v}} \frac{1}{N} \sum_{j=1}^{N} -\log P_{G(m_l \leftarrow \bm{v})}[o \mid x_j \oplus s],
\end{equation}
where $G(m_l \leftarrow \bm{v})$ denotes the model with the MLP output at layer~$l^*$ clamped to $\bm{v}$ at the subject position.
 
\paragraph{Constrained Optimization.}
The weight update is obtained by solving:
\begin{equation}
    \min_{\Delta W} \|\Delta W K\|_F^2 \quad \text{subject to} \quad \Delta W \bm{k}^* = \bm{r},
\end{equation}
where $\bm{r} = \bm{v}^* - W\bm{k}^*$ is the residual and $K$ is a matrix of pre-existing keys. The constraint ensures the edit is exactly satisfied, while the objective minimizes disturbance to existing associations.
 
\paragraph{Lagrangian Formulation.}
The Lagrangian is:
\begin{equation}
    \mathcal{L}(\Delta W, \boldsymbol{\lambda}) = \|\Delta W K\|_F^2 + \boldsymbol{\lambda}^\top (\bm{r} - \Delta W \bm{k}^*).
\end{equation}
Setting $\nabla_{\Delta W} \mathcal{L} = 0$ gives $2 \Delta W C = \boldsymbol{\lambda} \bm{k}^{*\top}$, where $C = KK^\top = \mathbb{E}[\bm{k}\bm{k}^\top]$ is the uncentered second-moment (Gram) matrix computed empirically over a text corpus. Substituting back and applying the constraint $\Delta W \bm{k}^* = \bm{r}$ yields:
\begin{equation}
    \Delta W = \bm{r} \frac{(C^{-1}\bm{k}^*)^\top}{\bm{k}^{*\top} C^{-1} \bm{k}^*}.
\end{equation}
This establishes a rank-1 update. The updated weight matrix is $\hat{W} = W + \Delta W$.
 
\paragraph{Minimum-Disturbance Interpretation.}
The factor $C^{-1}$ reweights the key space by \emph{safety-to-edit}: directions along which many existing keys are concentrated (large eigenvalues of $C$) are suppressed under $C^{-1}$, while sparse directions are amplified. The resulting update preferentially modifies the weight matrix in directions that minimally disturb stored key--value associations (elaborated in Appendix~\ref{app:c_geometry}).
 
\subsection{MEMIT: Mass-Editing Memory in a Transformer}
\label{app:memit}

MEMIT generalizes ROME to batch-edit $n$ facts simultaneously across a set of critical layers $\mathcal{S} = \{l_1, \ldots, l_L\}$.
 
\paragraph{Expanded Objective.}
Because satisfying thousands of hard equality constraints would likely produce conflicting solutions or severe model damage, MEMIT relaxes the constraint to a least-squares objective:
\begin{equation}
    \hat{W}_l = \arg\min_{W} \left( \sum_{i=1}^{n} \|W\bm{k}_i - \bm{v}_i\|^2 + \sum_{c=n+1}^{n+q} \|W\bm{k}_c - \bm{v}_c\|^2 \right),
\end{equation}
where the first sum covers the $n$ edit targets and the second preserves $q$ randomly sampled existing associations.
 
\paragraph{Residual Distribution Across Layers.}
A critical design choice in MEMIT is that the target value $\bm{v}_i^L$ is computed at the \emph{last} critical layer $L$, and the required update is distributed evenly across all critical layers. For each fact $i$ at layer $l$, the per-layer residual is:
\begin{equation}
    \bm{r}_i^l = \frac{\bm{v}_i^L - \bm{h}_i^L}{L - l + 1} = \frac{\bm{s}_i^L}{L - l + 1},
\end{equation}
where $\bm{s}_i^L = \bm{v}_i^L - \bm{h}_i^L$ is the total value shortfall at the last critical layer. Crucially, the numerator $\bm{s}_i^L$ is shared across all critical layers; only the denominator changes. This ensures that the residual's direction depends on the language and fact, but not on the specific critical layer itself; only its magnitude scales with $1/(L - l + 1)$.
 
\paragraph{Batch Update Formula.}
Collecting keys $K^l = [\bm{k}_1^l, \ldots, \bm{k}_n^l]$ and residuals $R^l = [\bm{r}_1^l, \ldots, \bm{r}_n^l]$ into matrices, the per-layer update generalizes to:
\begin{equation}
    \Delta W^l = R^l (K^l)^\top \left(C^l + K^l (K^l)^\top\right)^{-1},
    \label{eq:memit-update}
\end{equation}
where $C^l = \lambda\, \mathbb{E}[\bm{k}\bm{k}^\top]$ is the scaled second-moment matrix at layer $l$. The addition of $K^l(K^l)^\top$ accounts for mutual interference among all $n$ edit keys in the batch.
 
\subsection{AlphaEdit: Null-Space Constrained Editing}
\label{app:alphaedit}

AlphaEdit augments MEMIT by projecting the weight update onto the null space of the second-moment matrix~$C$, thereby strictly preserving pre-existing knowledge.
 
\paragraph{Null-Space Projection.}
Let $C = U \Sigma U^\top$ be the eigendecomposition of the second-moment matrix. AlphaEdit constructs the projection matrix:
\begin{equation}
    P_\perp = I - U_{\text{top}} U_{\text{top}}^\top,
\end{equation}
where $U_{\text{top}}$ contains the eigenvectors corresponding to the largest eigenvalues (directions most densely populated by existing keys). The AlphaEdit update strictly right-multiplies the MEMIT update to project the incoming keys:
\begin{equation}
    \Delta W_{\text{AE}} = \Delta W_{\text{MEMIT}} P_\perp.
\end{equation}
 
\paragraph{What Is Preserved vs.\ Sacrificed.}
By restricting the update strictly to the null space of $C$, AlphaEdit guarantees that $\Delta W_{\text{AE}} \bm{k} = \Delta W_{\text{MEMIT}} (P_\perp \bm{k}) \approx 0$ for any key $\bm{k}$ that lies heavily in the column space of the existing key distribution. This prevents catastrophic forgetting. However, the projection reduces the effective magnitude of the update applied to the target facts, which can limit edit efficacy—particularly when the edit key $\bm{k}^*$ itself shares a significant component with the dense column space.
 
\subsection{The Second-Moment Matrix $C$: Geometric Interpretation}
\label{app:c_geometry}

The matrix $C = \sum_i \bm{k}_i \bm{k}_i^\top$ is the uncentered second-moment (Gram) accumulation of memory keys. It appears in the ROME solution through the \emph{linear-algebraic} structure of the constrained optimization, not purely statistical reasoning.
 
\paragraph{$C$-Norm as Memory Density.}
For any direction $\bm{x}$, the quadratic form $\bm{x}^\top C \bm{x} = \sum_i (\bm{x}^\top \bm{k}_i)^2$ equals the sum of squared projections of $\bm{x}$ onto the stored keys. A large $C$-norm indicates a direction heavily used by existing memory; a small value indicates a sparse direction. Thus, $C$ defines an ellipsoid encoding \emph{memory density}.
 
\paragraph{$C^{-1}$ as Safety-to-Edit Reweighting.}
The ROME solution uses $C^{-1}$ because the optimization minimizes disturbance under the $C$-norm. Geometrically, $C^{-1}$ inverts the density weighting: dense directions (large eigenvalues in $C$) become small under $C^{-1}$, suppressing updates along heavily protected memory directions, while sparse directions are amplified, permitting larger updates where they cause minimal interference.
 
\paragraph{The Mahalanobis Inner Product.}
The quantity $\bm{k}_1^\top C^{-1} \bm{k}_2$ defines a \emph{Mahalanobis-type} inner product. Using the pseudo-whitening transform $\tilde{\bm{k}} = C^{-1/2}\bm{k}$, we can write:
\begin{equation}
    \bm{k}_1^\top C^{-1} \bm{k}_2 = (C^{-1/2}\bm{k}_1)^\top (C^{-1/2}\bm{k}_2) = \tilde{\bm{k}}_1^\top \tilde{\bm{k}}_2.
\end{equation}
This is the standard Euclidean inner product in the whitened space. Thus, the propagation scalar $\alpha_{1,2}$ measures key alignment in a \emph{density-adjusted} coordinate system where alignment within low-memory-density subspaces is upweighted.
 
\paragraph{The Propagation Scalar $\alpha$.}
The scalar $\alpha_{1,2} = \frac{\bm{k}_1^\top C^{-1} \bm{k}_2}{\bm{k}_1^\top C^{-1} \bm{k}_1}$ is the $C$-metric projection coefficient. While raw cosine similarity asks ``are two keys aligned?'', the $C$-metric similarity asks ``are two keys aligned in directions that are safe to edit?'' Cross-lingual propagation depends exclusively on the latter.

\subsection{Experimental and Implementation Details}
\label{app:implementation}

\paragraph{Language-Specific Reconstruction of $C$ Matrices.}
To compute the uncentered covariance matrix $C$, which captures the second-moment statistics of the key vectors, we sourced foundational text from a recent dump of the English Wikipedia corpus. To ensure high-quality, continuous text representations, the raw data underwent rigorous regex preprocessing to filter out extraneous artifacts (reference tags, citation brackets, and markdown formatting). This yielded a refined dataset consisting of 1,134 documents, totaling approximately 400,000 tokens.

\paragraph{Multilingual Adaptation.}
Because our approach requires cross-lingual factual optimization, it was imperative that the key vector distributions accurately reflect the linguistic properties of the target languages. The entire cleaned English corpus was systematically translated into the target languages using the \texttt{google/translategemma-4b-it} model, ensuring that semantic integrity and contextual distribution were preserved across linguistic variants.

\paragraph{Matrix Construction and Dimensionality.}
The covariance matrix $C$ is constructed as $C \propto K K^\top$, where $K$ represents a matrix of concatenated hidden state activations. For a given target language, the translated documents were passed through the target language model up to its maximum context window. For every token, we collected the post-nonlinearity activation vector, $\bm{k} \in \mathbb{R}^D$, from the specific MLP module targeted for editing (where $D$ represents the intermediate hidden dimension of the MLP layer). By sampling these $\bm{k}$ vectors across the ${\sim}400,000$ token corpus, we constructed the matrix $C \in \mathbb{R}^{D \times D}$. This cached estimate $C \propto \mathbb{E}[\bm{k}\bm{k}^\top]$ is critical for solving the constrained optimization required to insert new factual associations while minimizing interference.

\paragraph{Hyperparameters.}
The complete set of hyperparameters utilized for each editing method and across both models is detailed in Table~\ref{tab:editing_hyperparams}.

\begin{table}[H]
    \centering
    \caption{Hyperparameters for editing methods across models. Edit layers were selected by a sweep on CounterFact with English edits, scoring each candidate by $\Delta$NLL (Definition~3) averaged over source and paraphrase prompts. ROME's layer $l^*$ maximizes this score in English among nine candidate layers spanning the network. For MEMIT and AlphaEdit, the five-layer window $S$ maximizes the score averaged over the four non-English languages among five to seven candidate windows spaced four layers apart, rather than being taken from localization results. The covariance $C$ is the mean second moment of the keys over about 400{,}000 tokens of English Wikipedia.}
    \label{tab:editing_hyperparams}
    \small
    \vspace{0.5em}
    \begin{tabular}{llcc}
    \toprule
    & & Llama-3.1-8B & Gemma-3-4B \\
    \midrule
    \multicolumn{2}{l}{\textbf{Shared}} \\
    & Hidden dim ($d_{\text{in}}$) & 14336 & 10240 \\
    & Output dim ($d_{\text{out}}$) & 4096 & 2560 \\
    & Num layers & 32 & 34 \\
    & $\bm{v}^*$ loss layer & 31 & 33 \\
    & $\bm{v}^*$ learning rate & $5 \times 10^{-1}$ & $5 \times 10^{-1}$ \\
    & $\bm{v}^*$ weight decay & $1 \times 10^{-3}$ & $1 \times 10^{-3}$ \\
    & KL factor ($\lambda_{\text{KL}}$) & 0.0625 & 0.0625 \\
    % & Num prefixes ($N$) & 10 & 10 \\
    & $C$ sample size & 400{,}000 & 400{,}000 \\
    \midrule
    \multicolumn{2}{l}{\textbf{ROME}} \\
    & Edit layer (en) ($l^*$) & 20 & 32 \\
    & Covariance $\lambda$ (Tikhonov) & $1 \times 10^{-5}$ & $1 \times 10^{-5}$ \\
    \midrule
    \multicolumn{2}{l}{\textbf{MEMIT}} \\
    & Critical layers (en) ($\mathcal{S}$) & [24, 25, 26, 27, 28] & [12, 13, 14, 15, 16] \\
    % & Last critical layer (en) ($L$) & 20 & 32 \\
    & Covariance scale ($\lambda$) & $1.5 \times 10^{4}$ & $1.5 \times 10^{4}$ \\
    & Batch size ($n$) & 50 & 50 \\
    \midrule
    \multicolumn{2}{l}{\textbf{AlphaEdit}} \\
    & Critical layers (en) ($\mathcal{S}$) & [16, 17, 18, 19, 20] & [28, 29, 30, 31, 32] \\
    % & Last critical layer (en) ($L$) & 20 & 32 \\
   & L2 regularization & 10 & 10 \\
& Null-space eigenvalue threshold & 0.02 & 0.02 \\
    & Batch size ($n$) & 50 & 50 \\
    \bottomrule
    \end{tabular}
\end{table}

%% file: appendix/proofs_theory.tex
\section{Formal Proofs for Geometric Dissections and Rayleigh Bounds}
\label{app:proofs}

\begin{tcolorbox}[colback=gray!5, colframe=gray!50, title=Notation]
Throughout this appendix, $\bm{k}_1^*$ and $\bm{k}_2^*$ denote the ROME keys for the \emph{same} subject computed in the source (edit) language $\ell_1$ and the target (evaluation) language $\ell_2$, respectively. Likewise, $\bm{v}_1^*$ and $\bm{v}_2^*$ are the corresponding optimized target values, and $\bm{r}_i = \bm{v}_i^* - W\bm{k}_i^*$ is the residual in language~$\ell_i$. The matrix $C = C_1$ is the second-moment matrix of $\ell_1$, $W$ is the pre-edit down-projection weight, and $\hat{W}$ is the weight matrix after the ROME update is applied in~$\ell_1$.
\end{tcolorbox}

\subsection{Cross-Lingual Value Gap (Definition 2)}
\label{app:proof_gap}

\begin{proof}
We edit in $\ell_1$ and evaluate in $\ell_2$. Applying the updated weight $\hat{W}$ to the target-language key $\bm{k}_2^*$:
\begin{equation}\label{eq:app_rome_applied}
    \hat{W}\bm{k}_2^* \;=\; W\bm{k}_2^* \;+\; \bm{r}_1 \cdot \frac{(\bm{k}_1^*)^\top C^{-1}\, \bm{k}_2^*}{(\bm{k}_1^*)^\top C^{-1}\, \bm{k}_1^*} 
    \;=\; W\bm{k}_2^* \;+\; \alpha_{1,2}\,\bm{r}_1.
\end{equation}
The first term is the pre-edit output for the $\ell_2$ key; the second is the fraction of the English update that survives the cross-lingual key-space gate, scaled by $\alpha_{1,2}$.

Now consider the value that $\ell_2$ \emph{would have needed} had it been edited natively: $\bm{v}_2^*$. The gap ($\bm{g}_{1,2}$) between this desired value and what the $\ell_1$ edit actually delivers is:
\begin{align}
    \bm{v}_2^* - \hat{W}\bm{k}_2^* &= \bm{v}_2^* - W\bm{k}_2^* - \alpha_{1,2}\,\bm{r}_1 \notag\\
    &= \underbrace{(\bm{v}_2^* - W\bm{k}_2^*)}_{\bm{r}_2} - \alpha_{1,2}\,\bm{r}_1 \notag\\
    &= \bm{r}_2 - \alpha_{1,2}\,\bm{r}_1. \qedhere
\end{align}
\end{proof}

\paragraph{Interpretation.}
The value gap $\bm{g}_{1,2} = \bm{r}_2 - \alpha_{1,2}\,\bm{r}_1$ has a precise geometric meaning: $\bm{r}_2$ is what the target language \emph{needs}---the full correction that a native $\ell_2$ edit would apply---while $\alpha_{1,2}\,\bm{r}_1$ is what the source-language edit actually \emph{delivers} after passing through the key-space gate. Perfect cross-lingual transfer requires these two vectors to coincide, which demands both that $\alpha_{1,2}$ matches the value-space projection $\gamma$ (magnitude matching) and that $\bm{r}_2$ is strictly parallel to $\bm{r}_1$ (directional alignment). 

\subsection{Proof of Proposition~\ref{prop:decomp} (Parallel--Orthogonal Decomposition)}
\label{app:proof_decomp}

\begin{proof}
Decompose $\bm{r}_2 = \gamma\, \bm{r}_1 + \bm{r}_2^\perp$ where $\gamma = \frac{\bm{r}_2 \cdot \bm{r}_1}{\|\bm{r}_1\|^2}$ and $\bm{r}_2^\perp \perp \bm{r}_1$.
Substituting into our gap definition:
\begin{equation}
    \bm{g}_{1,2} = \bm{r}_2 - \alpha_{1,2}\,\bm{r}_1 = \gamma\,\bm{r}_1 + \bm{r}_2^\perp - \alpha_{1,2}\,\bm{r}_1 = (\gamma - \alpha_{1,2})\,\bm{r}_1 + \bm{r}_2^\perp.
\end{equation}
Since $\bm{r}_2^\perp \perp \bm{r}_1$, the cross-term vanishes when computing the squared norm:
\begin{equation}
    \|\bm{g}_{1,2}\|^2 = \|(\gamma - \alpha_{1,2})\,\bm{r}_1\|^2 + \|\bm{r}_2^\perp\|^2 = (\gamma - \alpha_{1,2})^2 \|\bm{r}_1\|^2 + \|\bm{r}_2^\perp\|^2.
\end{equation}
Dividing both sides by $\|\bm{r}_1\|^2$ yields the normalized form in Equation~\ref{eq:gapnorm_normalized}.
\end{proof}

\subsection{Derivation of Key-Space Alignment Bounds (Proposition~\ref{prop:rayleigh})}
\label{app:proof_rayleigh}

\paragraph{Whitening Transform.}
\begin{proof}
Define the pseudo-whitened key $\tilde{\bm{k}} = C_1^{-1/2}\bm{k}$.
The propagation scalar can be rewritten as:
\begin{equation}
    \alpha_{1,2} = \frac{\bm{k}_1^{*\top} C_1^{-1} \bm{k}_2^*}{\bm{k}_1^{*\top} C_1^{-1} \bm{k}_1^*} = \frac{\tilde{\bm{k}}_1^\top \tilde{\bm{k}}_2}{\tilde{\bm{k}}_1^\top \tilde{\bm{k}}_1}.
\end{equation}
\end{proof}
 
\paragraph{Linear Key-Shift Assumption.}
We \textbf{assume} the cross-lingual key shift can be approximated by a linear transformation: $\bm{k}_2^* = A_{1,2}\, \bm{k}_1^*$, where $A_{1,2} \in \mathbb{R}^{D \times D}$ captures the systematic geometric relationship between source and target language keys. Under this assumption:
\begin{equation}
    \tilde{\bm{k}}_2 = C_1^{-1/2}\bm{k}_2^* = C_1^{-1/2} A_{1,2}\, \bm{k}_1^* = C_1^{-1/2} A_{1,2}\, C_1^{1/2} \tilde{\bm{k}}_1 = M_{1,2}\, \tilde{\bm{k}}_1,
\end{equation}
where $M_{1,2} \coloneqq C_1^{-1/2} A_{1,2}\, C_1^{1/2}$ is the key-space transfer matrix in the whitened coordinate system.
 
\paragraph{Rayleigh Quotient Identification.}
Substituting:
\begin{equation}
    \alpha_{1,2} = \frac{\tilde{\bm{k}}_1^\top M_{1,2}\, \tilde{\bm{k}}_1}{\tilde{\bm{k}}_1^\top \tilde{\bm{k}}_1},
\end{equation}
which is exactly the Rayleigh quotient $R(M_{1,2}) = \frac{\bm{x}^\top M \bm{x}}{\bm{x}^\top \bm{x}}$ evaluated at $\bm{x} = \tilde{\bm{k}}_1$.
 
\paragraph{Eigenvalue Bounds.}
By the min-max characterization of the Rayleigh quotient applied to the symmetric part $\frac{1}{2}(M_{1,2} + M_{1,2}^\top)$:
\begin{equation}
    \lambda_{\min}\left(\frac{M_{1,2} + M_{1,2}^\top}{2}\right) \leq \alpha_{1,2} \leq \lambda_{\max}\left(\frac{M_{1,2} + M_{1,2}^\top}{2}\right).
\end{equation}
 
\paragraph{Perfect Transfer Condition.}
If $A_{1,2} = I$ (i.e., the key representations are identical across languages), then $M_{1,2} = C_1^{-1/2} I\, C_1^{1/2} = I$, so $\alpha_{1,2} = 1$ (perfect propagation). As the geometries diverge, the spectrum of $M_{1,2}$ spreads, and $\alpha_{1,2}$ fluctuates depending on how $\tilde{\bm{k}}_1$ aligns with the principal directions of mismatch.
 
\paragraph{Assumption-Free Alternative.}
When the linear key-shift assumption is not warranted, the normalized Frobenius divergence $d_F(C_1, C_2) = \|C_1 - C_2\|_F / \|C_1\|_F$ provides a direct, assumption-free measure of the geometric mismatch between languages. As shown empirically, $d_F$ tracks the same typological ordering as $\alpha_{1,2}$ and serves as a reliable proxy. Moreover, later analysis using linear predictors shows there is non-trivial predictability of keys across languages, albeit small.
 
\paragraph{Validity of the Linear Assumption.}
We note that the linear key-shift model $\bm{k}_2 = A_{1,2}\,\bm{k}_1$ is an approximation. In practice, the mapping between cross-lingual key representations involves nonlinearities introduced by the activation function and the attention mechanism. However, for semantically equivalent fact queries, the MLP hidden states tend to lie in locally linear regimes, making the approximation reasonable for the population-level analysis we perform.

\subsection{Proof of Proposition~\ref{prop:logit} (Downstream Logit Gap)}
\label{app:proof_logit}

\begin{proof}
Let $\mathbf{z}_0 = W\mathbf{k}^*_2$ be the pre-edit value representation at the edited layer. Under the edit, the representation changes to $\hat{W}\mathbf{k}^*_2 = \mathbf{z}_0 + \alpha_{1,2}\mathbf{r}_1$. The desired representation is $\mathbf{v}^*_2 = \mathbf{z}_0 + \mathbf{r}_2$. Let $f$ denote the function mapping the value representation at layer $l^*$ to the output logits, so $\mathbf{y} = f(\mathbf{z}) = W_U\,\mathbf{h}(\mathbf{z})$, where $\mathbf{h}(\mathbf{z}) \in \mathbb{R}^{d_{\mathrm{out}}}$ is the final hidden state and $W_U \in \mathbb{R}^{|V| \times d_{\mathrm{out}}}$ is the unembedding matrix (we absorb the local Jacobian of the final normalization layer into $W_U$). By a first-order Taylor expansion around $\mathbf{z}_0$:
\begin{align}
\Delta\mathbf{y}_{\mathrm{desired}} &\approx W_U J_h\, \mathbf{r}_2, \\
\Delta\mathbf{y}_{\mathrm{actual}} &\approx W_U J_h\, \alpha_{1,2}\mathbf{r}_1,
\end{align}
where $J_h = \partial\mathbf{h}/\partial\mathbf{z}\,|_{\mathbf{z}_0} \in \mathbb{R}^{d_{\mathrm{out}} \times d_{\mathrm{out}}}$ is the Jacobian of the final hidden state, evaluated at the pre-edit representation. The logit gap is:
\begin{equation}
\Delta\mathbf{y}_{\mathrm{gap}} = \Delta\mathbf{y}_{\mathrm{desired}} - \Delta\mathbf{y}_{\mathrm{actual}} \approx W_U J_h\,(\mathbf{r}_2 - \alpha_{1,2}\mathbf{r}_1) = W_U J_h\, \mathbf{g}_{1,2}.
\end{equation}
\end{proof}

\paragraph{Caveat.} Strictly, the Jacobians for the desired and actual updates should be evaluated at different points ($\mathbf{z}_0 + \mathbf{r}_2$ and $\mathbf{z}_0 + \alpha_{1,2}\mathbf{r}_1$, respectively), making this a first-order approximation. For middle-layer edits where residual connections dominate and the downstream function is approximately linear, $J_h \approx I$, and the logit gap reduces to the value gap mapped through the unembedding, $W_U\,\mathbf{g}_{1,2}$. This approximation concerns the square Jacobian $J_h$, not the full map into the $|V|$-dimensional logits.

\subsection{Proof of Proposition~\ref{prop:logit} (Cross-Lingual Transfer Penalty Bound)}
\label{app:penalty_proof}

We show that the cross-lingual transfer penalty $\mathcal{P} \coloneqq \Delta\mathrm{NLL}_{\mathrm{self}} - \Delta\mathrm{NLL}_{\mathrm{en}}$ is governed by the residual gap $\bm{g}_{1,2} \coloneqq \bm{r}_2 - \alpha_{1,2}\bm{r}_1$.

\paragraph{Setup.} 
Let $\bm{y} \in \mathbb{R}^{|V|}$ denote the pre-softmax logit vector over the vocabulary $V$, and let $o_2$ be the index of the correct target-language answer token. Define the negative log-likelihood:
\begin{equation}
    \mathrm{NLL}(\bm{y}) = -\log p_{o_2}(\bm{y}) = -y_{o_2} + \log \sum_{j \in V} \exp(y_j).
\end{equation}

We consider three settings, all evaluated on the target-language prompt $(s_2, r_2)$:
\begin{enumerate}[label=(\roman*)]
    \item \textbf{Pre-edit:} logits $\bm{y}^{(0)}$, giving $\mathrm{NLL}^{(0)}$.
    \item \textbf{Cross-lingual edit (edit in $\ell_1$, eval in $\ell_2$):} The value representation at the edited layer shifts by $\alpha_{1,2}\bm{r}_1$. The logits become $\bm{y}^{(\mathrm{en})} \approx \bm{y}^{(0)} + W_U J_h \alpha_{1,2} \bm{r}_1$.
    \item \textbf{Native edit (edit and eval in $\ell_2$):} The value representation shifts by $\bm{r}_2$. The logits become $\bm{y}^{(\mathrm{self})} \approx \bm{y}^{(0)} + W_U J_h \bm{r}_2$.
\end{enumerate}

\paragraph{Transfer Penalty in Terms of the Gap.} 
Define the NLL reductions:
\begin{equation}
    \Delta\mathrm{NLL}_{\mathrm{en}} \coloneqq \mathrm{NLL}^{(0)} - \mathrm{NLL}^{(\mathrm{en})}, \qquad \Delta\mathrm{NLL}_{\mathrm{self}} \coloneqq \mathrm{NLL}^{(0)} - \mathrm{NLL}^{(\mathrm{self})}.
\end{equation}
A larger $\Delta\mathrm{NLL}$ indicates a more successful edit. The cross-lingual transfer penalty is:
\begin{equation}
    \mathcal{P} \coloneqq \Delta\mathrm{NLL}_{\mathrm{self}} - \Delta\mathrm{NLL}_{\mathrm{en}} = \mathrm{NLL}^{(\mathrm{en})} - \mathrm{NLL}^{(\mathrm{self})}.
    \label{eq:penalty_def}
\end{equation}

\begin{proof}
Apply a first-order Taylor expansion of $\mathrm{NLL}(\bm{y})$ around $\bm{y}^{(0)}$. The gradient is:
\begin{equation}
    \nabla_{\bm{y}} \mathrm{NLL}(\bm{y})\big|_{\bm{y}^{(0)}} = -\bm{e}_{o_2} + \mathrm{softmax}(\bm{y}^{(0)}) = \bm{q}^{(0)} - \bm{e}_{o_2}.
\end{equation}

For the native edit:
\begin{equation}
    \mathrm{NLL}^{(\mathrm{self})} \approx \mathrm{NLL}^{(0)} + (\bm{q}^{(0)} - \bm{e}_{o_2})^\top (\bm{y}^{(\mathrm{self})} - \bm{y}^{(0)}) = \mathrm{NLL}^{(0)} + (\bm{q}^{(0)} - \bm{e}_{o_2})^\top W_U J_h \bm{r}_2.
\end{equation}

For the cross-lingual edit:
\begin{equation}
    \mathrm{NLL}^{(\mathrm{en})} \approx \mathrm{NLL}^{(0)} + (\bm{q}^{(0)} - \bm{e}_{o_2})^\top W_U J_h \alpha_{1,2}\bm{r}_1.
\end{equation}

Subtracting the two equations yields the expanded transfer penalty:
\begin{align}
    \mathcal{P} &= \mathrm{NLL}^{(\mathrm{en})} - \mathrm{NLL}^{(\mathrm{self})} \nonumber \\
    &\approx (\bm{q}^{(0)} - \bm{e}_{o_2})^\top W_U J_h \bigl(\alpha_{1,2}\bm{r}_1 - \bm{r}_2\bigr) \nonumber \\
    &= (\bm{e}_{o_2} - \bm{q}^{(0)})^\top W_U J_h \underbrace{(\bm{r}_2 - \alpha_{1,2}\bm{r}_1)}_{\bm{g}_{1,2}}.
    \label{eq:penalty_derivation}
\end{align}
\end{proof}

\paragraph{Interpretation.} 
Eq.~\eqref{eq:penalty_derivation} reveals that the transfer penalty is the inner product of two vectors:
\begin{enumerate}[label=(\alph*)]
    \item The \emph{surprise direction} $(\bm{e}_{o_2} - \bm{q}^{(0)})$, pointing from the model's pre-edit belief toward the correct answer. Large when the model is uncertain about $o_2$.
    \item The \emph{mapped gap} $W_U J_h \bm{g}_{1,2}$, which is the value-space gap projected to the logit space. By the parallel--orthogonal decomposition, this decomposes as:
    \begin{equation}
        W_U J_h \bm{g}_{1,2} = (\gamma - \alpha_{1,2}) W_U J_h \bm{r}_1 + W_U J_h \bm{r}_2^\perp.
    \end{equation}
\end{enumerate}
The penalty $\mathcal{P}$ is large when (i)~the model is initially wrong about $o_2$, and (ii)~the gap $\bm{g}_{1,2}$ has a substantial component along the direction that would correct the prediction. Since $\bm{r}_2^\perp$ dominates $\bm{g}_{1,2}$, the penalty is expected to be driven primarily by directional misalignment.

\paragraph{Applying the Cauchy--Schwarz Bound.} 
Taking the absolute value and applying Cauchy--Schwarz:
\begin{equation}
    |\mathcal{P}| \;\leq\; \|\bm{e}_{o_2} - \bm{q}^{(0)}\| \;\cdot\; \|W_U J_h \bm{g}_{1,2}\|.
    \label{eq:penalty_cs}
\end{equation}
Since $\|\bm{e}_{o_2} - \bm{q}^{(0)}\| \leq \sqrt{2}$ (bounded by the simplex geometry), the penalty is strictly controlled by the norm of the mapped gap. In mid-to-late layers where $J_h \approx I$, this simplifies to:
\begin{equation}
    |\mathcal{P}| \;\lesssim\; \sqrt{2}\|W_U\|_2\,\;\|\bm{g}_{1,2}\| \;=\; \sqrt{2}\|W_U\|_2\,\;\sqrt{(\gamma - \alpha_{1,2})^2\|\bm{r}_1\|^2 + \|\bm{r}_2^\perp\|^2}.
    \label{eq:penalty_bound_norm}
\end{equation}

\paragraph{Connection to the Residual Prediction Barrier.} Taking expectations over facts and applying Proposition~4 under its stated assumption on $\alpha_{1,2}$:
\begin{equation}
\mathbb{E}\big[\|\mathbf{g}_{1,2}\|^2\big] \ge \sigma^2_{1,2} = \eta_{1,2} \cdot \operatorname{Var}(\mathbf{r}_2).
\label{eq:penalty_floor}
\end{equation}
This bounds the internal gap from below, but not the transfer penalty: the Cauchy--Schwarz bounds above are upper bounds, and to first order $P \approx \mathbf{u}^{\top}\mathbf{g}_{1,2}$ with $\mathbf{u} := J_h^{\top} W_U^{\top}(\mathbf{e}_{o_2} - \mathbf{q}^{(0)})$, so a large gap that is nearly orthogonal to $\mathbf{u}$ produces a small penalty. A lower bound on the penalty would need an additional assumption, for instance that for every fact and every source-residual-only correction $f$, the corrected gap $\mathbf{g}_f := \mathbf{r}_2 - \alpha_{1,2} f(\mathbf{r}_1)$ satisfies $|\mathbf{u}^{\top}\mathbf{g}_f| \ge c\,\|\mathbf{g}_f\|$ for some $c > 0$; the penalty $P_f$ of the corrected edit would then satisfy $\mathbb{E}[P_f^2] \gtrsim c^2\,\mathbb{E}\|\mathbf{g}_f\|^2 \ge c^2\,\eta_{1,2}\operatorname{Var}(\mathbf{r}_2)$. We do not verify this assumption, and we treat the relation between the internal gap and the behavioral penalty as empirical (Table~\ref{tab:corr_gap_nll}).

\subsection{MEMIT Extension: Batch Alignment Matrix Derivation}
\label{app:memit_derivation}

\paragraph{Setup.}
For a given critical layer $l$, the MEMIT update applied to the target-language key matrix $K_2$ produces:
\begin{equation}
    \hat{W}^l K_2^l = W^l K_2^l + R_1^l (K_1^l)^\top \left(C^l + K_1^l (K_1^l)^\top\right)^{-1} K_2^l.
\end{equation}
 
\paragraph{Batch Alignment Matrix.}
Define the batch alignment matrix:
\begin{equation}
    \bm{A}_{1,2}^l = (K_1^l)^\top \left(C^l + K_1^l (K_1^l)^\top\right)^{-1} K_2^l \in \mathbb{R}^{n \times n_2},
\end{equation}
so that the batch update on $\ell_2$ keys becomes $\hat{W}^l K_2^l = W^l K_2^l + R_1^l \bm{A}_{1,2}^l$. $\bar{\alpha}_{\text{batch}}$ is the mean of the diagonal of this matrix.
 
\paragraph{Batch Value Gap.}
At the last critical layer $L$, the gap becomes:
\begin{equation}
    \bm{G}_{1,2}^L = V_2^L - \hat{W}^L K_2^L = R_2^L - R_1^L \bm{A}_{1,2}^L,
\end{equation}
where $R_2^L$ is the target-language residual matrix.
 
\paragraph{Reduction to ROME.}
For $n = n_2 = 1$, the matrix $\bm{A}_{1,2}$ reduces to the scalar:
\begin{equation}
    \vec{\alpha}_{1,2} = \bm{k}_1^\top (C + \bm{k}_1 \bm{k}_1^\top)^{-1} \bm{k}_2.
\end{equation}
By the Sherman--Morrison identity, $(C + \bm{k}_1\bm{k}_1^\top)^{-1} = C^{-1} - \frac{C^{-1}\bm{k}_1\bm{k}_1^\top C^{-1}}{1 + \bm{k}_1^\top C^{-1}\bm{k}_1}$, which recovers $\alpha_{1,2}$ (up to a normalizing constant), confirming consistency with the ROME analysis.
 
\paragraph{Diagnostic Quantities.}
We compute four diagnostic quantities from $\bm{A}_{1,2}$:
\begin{enumerate}
    \item \textbf{Mean self-propagation} ($\bar{\alpha}_{\text{batch}}$): The mean of the diagonal entries $\frac{1}{n}\text{tr}(\bm{A}_{1,2})$ (when $n_2 = n$). This defines whether the average ``alpha gate'' follows the ROME hierarchy.
    \item \textbf{Interference Ratio} (IR): $\text{IR} \coloneqq \frac{\|\text{off-diag}(\bm{A}_{1,2})\|_F}{\|\text{diag}(\bm{A}_{1,2})\|_F}$. This quantifies how much an edit to one subject leaks into the representation of another during cross-lingual transfer. $\text{IR} \gg 1$ indicates severe cross-fact interference.
    \item \textbf{Batch-wise scalar projection} ($\Gamma$): $\Gamma = \frac{\text{Tr}(R_2^\top (R_1 \bm{A}_{1,2}))}{\|R_1 \bm{A}_{1,2}\|_F^2}$. Captures how much the English residual batch, filtered through the geometric gate $\bm{A}_{1,2}$, covers the required target-language updates.
    \item \textbf{Batch orthogonal norm} ($\Sigma$): $\Sigma = \|R_2 - \Gamma(R_1 \bm{A}_{1,2})\|_F$. Quantifies the structural gap that no amount of $\alpha$-scaling can bridge.
\end{enumerate}
 
\subsection{AlphaEdit Extension: Effect of Null-Space Projection}
\label{app:alphaedit_derivation}

\paragraph{AlphaEdit Gap Derivation.}
With the null-space projection $P_\perp$, the AlphaEdit update restricts parameter changes to the null space of $C$ by right-multiplying the update matrix: $\Delta W_{\text{AE}}^L = \Delta W_{\text{MEMIT}}^L P_\perp$. When applied to the target-language keys $K_2^L$, they are filtered through this projection matrix ($P_\perp K_2^L$). The cross-lingual value gap is therefore:
\begin{equation}
    \bm{G}_{\text{gap,AE}}^L = R_2^L - R_1^L \bm{A}_{1,2}^{\text{AE}}, \quad \text{where} \quad \bm{A}_{1,2}^{\text{AE}} = (K_1^L)^\top(C^L + K_1^L(K_1^L)^\top)^{-1}(P_\perp K_2^L).
\end{equation}
 
\paragraph{Why the Orthogonal Mismatch Persists.}
The projection matrix $P_\perp$ operates strictly on the \emph{target-language keys} $K_2$ in the key space. It mathematically constrains the incoming keys to the null space of the English key distribution, thereby protecting pre-existing knowledge. However, the orthogonal mismatch $\bm{r}_2^\perp$---the component of the target residual perpendicular to the source residual---exists entirely within the \emph{value space}. Because $P_\perp$ is a key-space projection, it has no mathematical mechanism to rotate or alter the value-space residual vectors ($R_1$ or $R_2$). Consequently, AlphaEdit perfectly safeguards existing knowledge, but leaves the fundamental directional misalignment $\bm{r}_2^\perp$ completely uncorrected.

%% file: appendix/empirical.tex
\section{Extended Empirical Characterization of the Transfer Gap.}
\label{app:extended_empirical}

\subsection{Full Layer-Wise Plots: Gemma-3-4B}
\label{app:gemma_full}

\begin{figure}[h]
    \centering
    \includegraphics[width=1\linewidth]{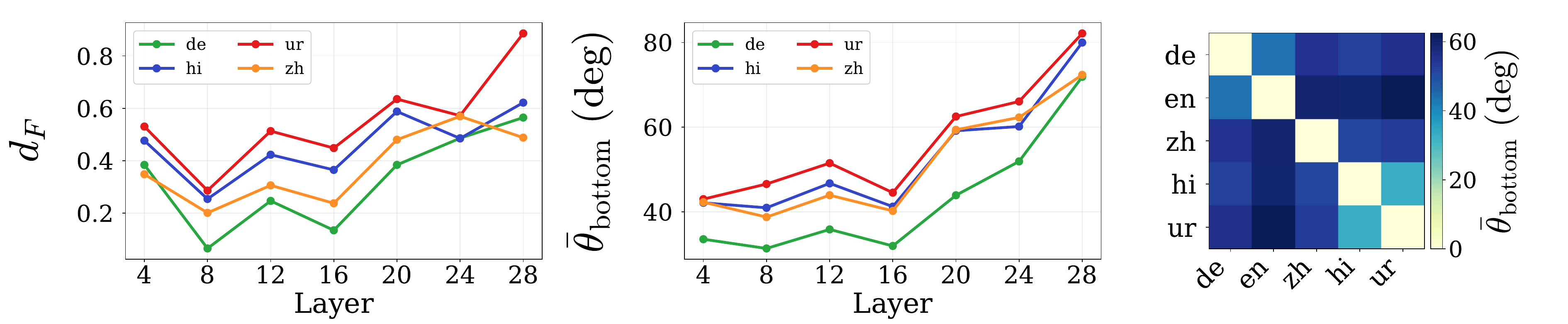}
    \caption{\textbf{The key-space geometry of Gemma-3-4B is governed by typological distance with a characteristic syntactic spike.} 
(a) Frobenius divergence $d_{F}(C_{\text{en}}, C_{j})$ remains relatively stable in early layers but increases steadily with layer depth, indicating the direction where internal key-space representations become most language-specific. 
(b) Principal angles between the bottom eigenspaces of English and target languages show that Indic languages (Hindi, Urdu) approach near-orthogonality ($>70^{\circ}$) more rapidly than in Llama-3.1-8B. 
(c) Pairwise heatmap at layer 16 confirms that geometric misalignment is universal across all typologically distant pairs and is not restricted to English-centric transfers. The gap from English is much more prominent for Gemma-3-4B than Llama-3.1-8B.}
    \label{fig:gemma_C_matrix_geomtry}
\end{figure}

\begin{figure}[h]
    \centering
    \begin{subfigure}{0.32\textwidth}
        \centering
        \includegraphics[width=\linewidth]{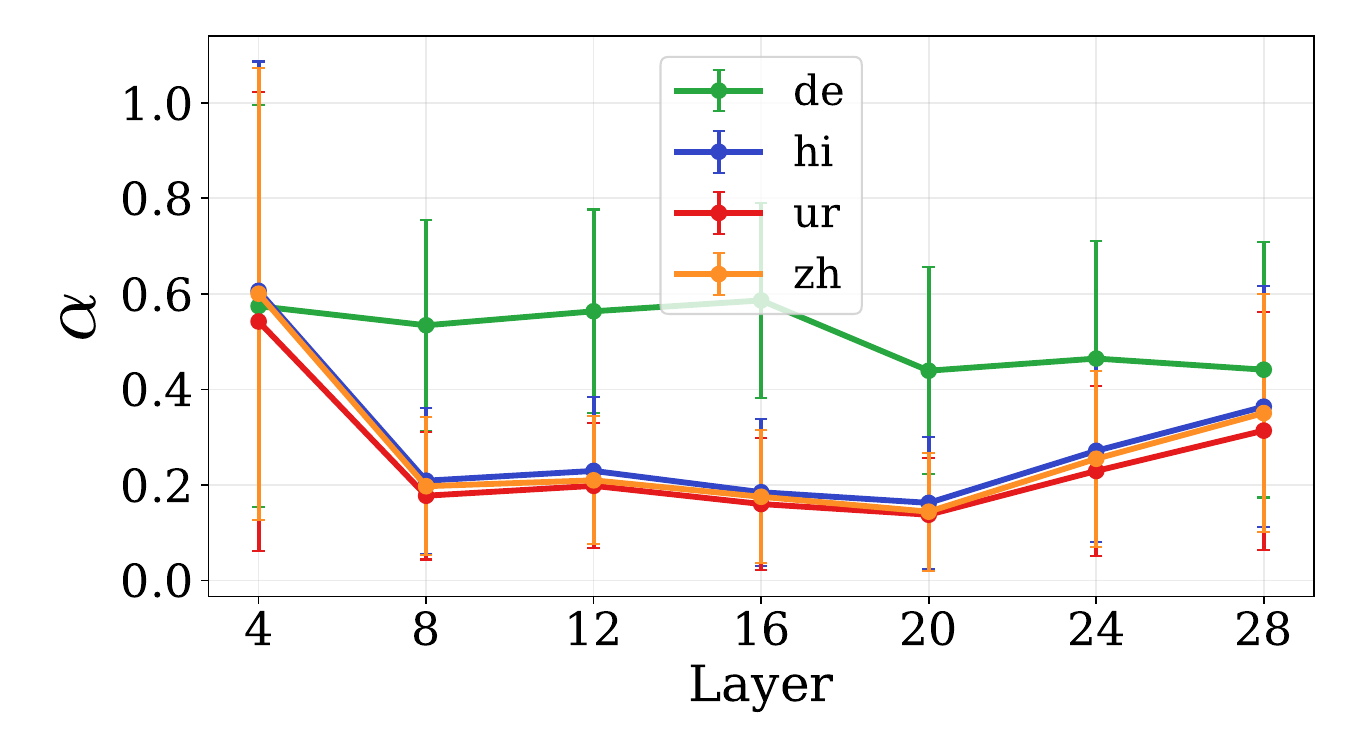}
        \caption{ROME}
        \label{fig:alpha_from_cases_by_language_gemma_rome}
    \end{subfigure}
    \hfill
    \begin{subfigure}{0.32\textwidth}
        \centering
        \includegraphics[width=\linewidth]{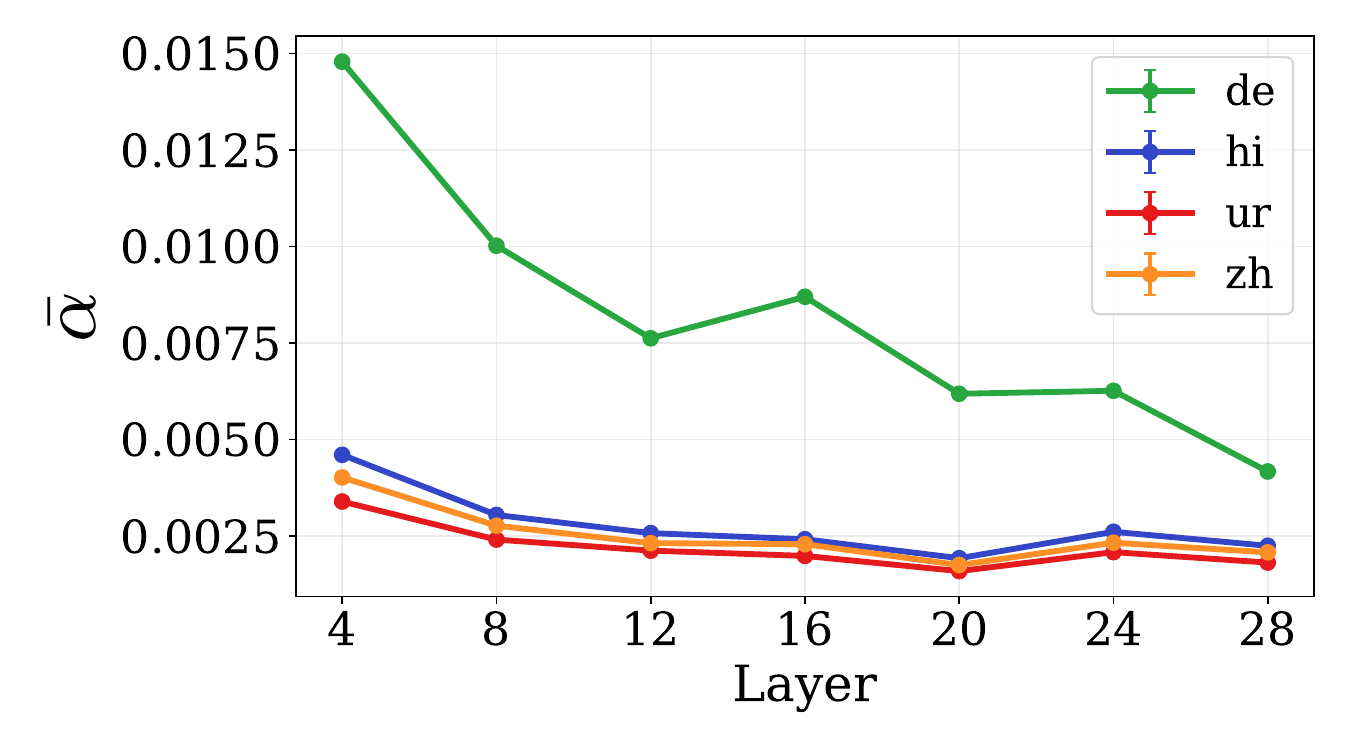}
        \caption{MEMIT}
        \label{fig:alpha_from_cases_by_language_gemma_memit}
    \end{subfigure}
    \hfill
    \begin{subfigure}{0.32\textwidth}
        \centering
        \includegraphics[width=\linewidth]{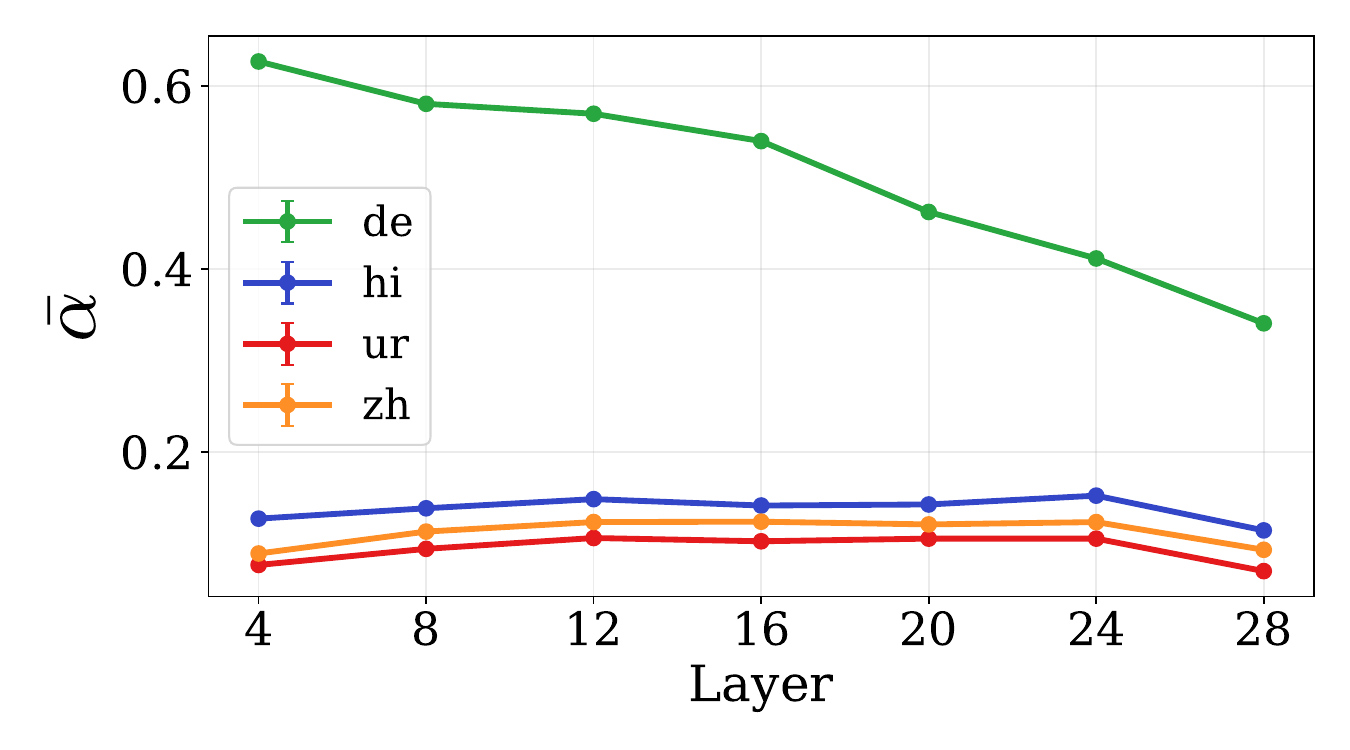}
        \caption{AlphaEdit}
        \label{fig:alpha_from_cases_by_language_llama_alphaedit}
    \end{subfigure}
    
    \caption{\textbf{Layer-wise trend for the mean key-space propagation scalar $\bar{\alpha}$ on Gemma-3-4B.} 
    The propagation magnitude follows a typologically stratified pattern, with the hierarchy \texttt{de} $\gg$ \texttt{zh} $>$ \texttt{hi} $\approx$ \texttt{ur}. Compared to Llama-3.1-8B, the absolute magnitudes are generally lower, suggesting stronger cross-lingual key-space separation in Gemma-3-4B and weaker update propagation early in the forward pass.}
    \label{fig:alpha_from_cases_by_language_gemma}
\end{figure}

\begin{figure}[h]
    \centering
    \includegraphics[width=1\linewidth]{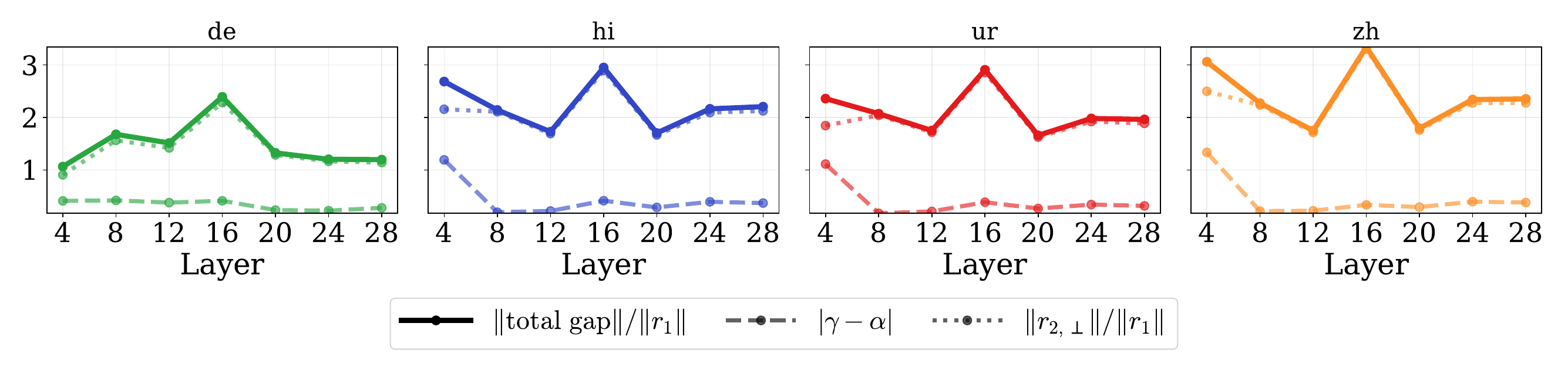}
    \caption{\textbf{Value-space gap decomposition for Gemma-3-4B.} The relative total gap (solid line) is universally dominated by the orthogonal residual mismatch component $\bm{r}_{2}^{\perp}$ (dotted line). Similar to the Llama-3.1-8B results, this directional misalignment accounts for the vast majority of the transfer failure, particularly at the optimal editing layers (28--32).}
    \label{fig:gemma_residual_gap}
\end{figure}

To verify that the structural bottlenecks identified in the main text generalize across architectures, we replicate our full layer-wise geometric analysis on Gemma-3-4B. Figure~\ref{fig:gemma_C_matrix_geomtry} illustrates the key-space covariance geometry, confirming that Gemma also exhibits severe spatial fracturing along typological lines. This geometric divergence directly dictates the surviving update signal, as evidenced by the severe, language-stratified decay of the propagation scalar ($\bar{\alpha}$) across all editing methods in Figure~\ref{fig:alpha_from_cases_by_language_gemma}. Finally, Figure~\ref{fig:gemma_residual_gap} provides the relative gap decomposition, showing that the orthogonal value-space mismatch ($\mathbf{r}_2^{\perp}$) remains the dominant failure mode for zero-shot cross-lingual transfer and accounts for the vast majority of the error at the optimal editing layers.

\subsection{MEMIT Batch Analysis}
\label{app:memit_full}

\begin{figure}[h]
    \centering
    \begin{subfigure}{0.48\textwidth}
        \centering
        \includegraphics[width=\linewidth]{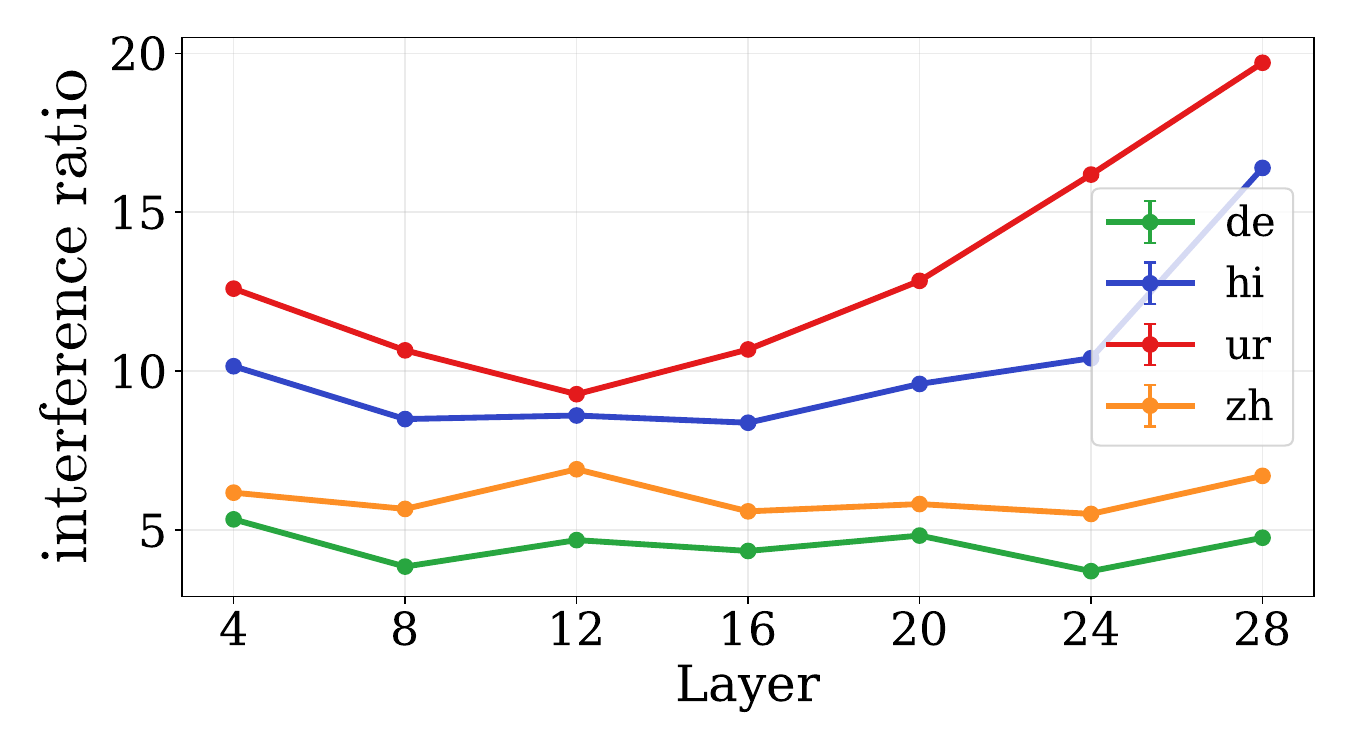}
        \caption{Llama-3.1-8B}
    \end{subfigure}
    \hfill
    \begin{subfigure}{0.48\textwidth}
        \centering
        \includegraphics[width=\linewidth]{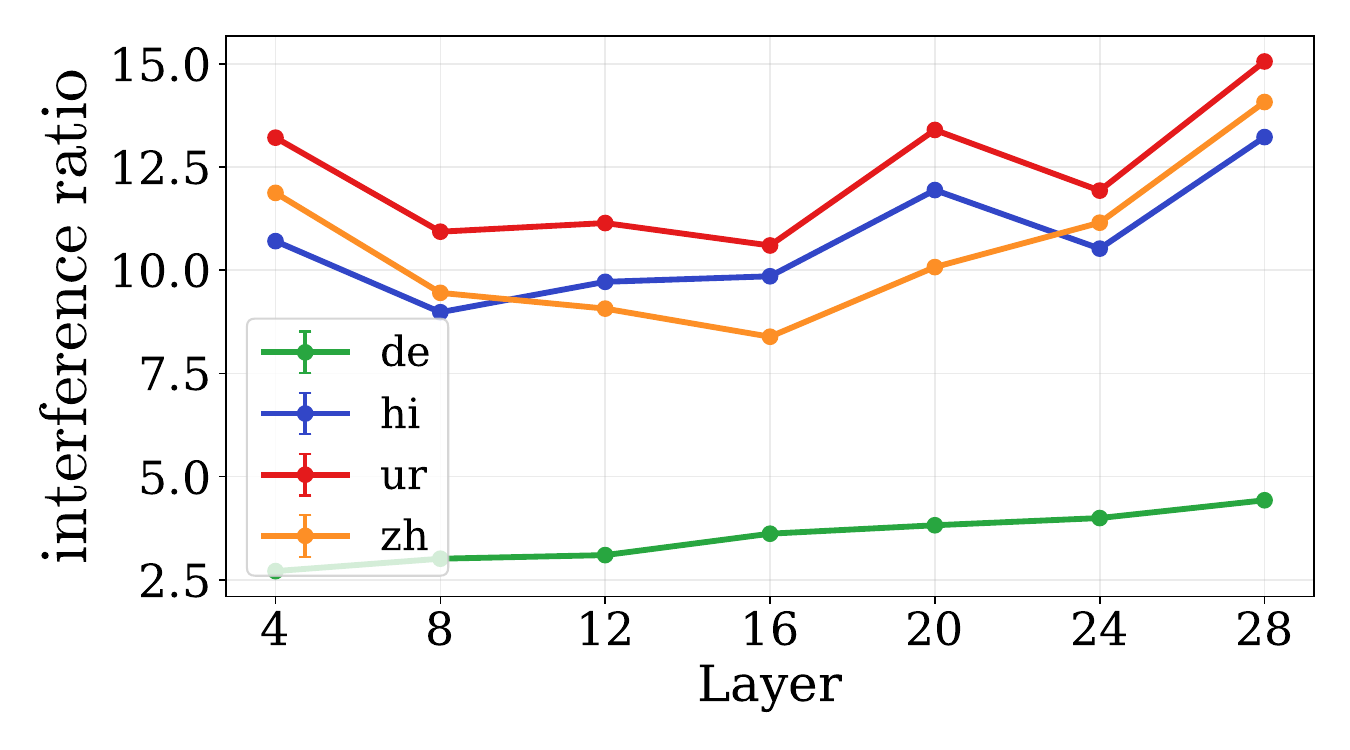}
        \caption{Gemma-3-4B}
    \end{subfigure}
    \caption{\textbf{MEMIT Interference Ratio (IR) across layers.} When editing in English, distant target languages suffer from severe off-diagonal cross-fact interference, corrupting the surviving update signal.}
    \label{fig:memit_ir}
\end{figure}

As discussed in Section~\ref{sec:key_space_prop}, batch editing introduces cross-fact interference that can compound the
cross-lingual transfer gap. Figure~\ref{fig:memit_ir} illustrates the Interference Ratio ($\text{IR}$) across layers for MEMIT when edited natively in English. 
For both models, typologically distant languages exhibit large interference spikes ($\text{IR} > 10$),
suggesting that batching can corrupt the weak transfer signal for these languages via off-diagonal leakage.

\begin{figure}[h]
    \centering
    \begin{subfigure}{0.48\textwidth}
        \centering
        \includegraphics[width=\linewidth]{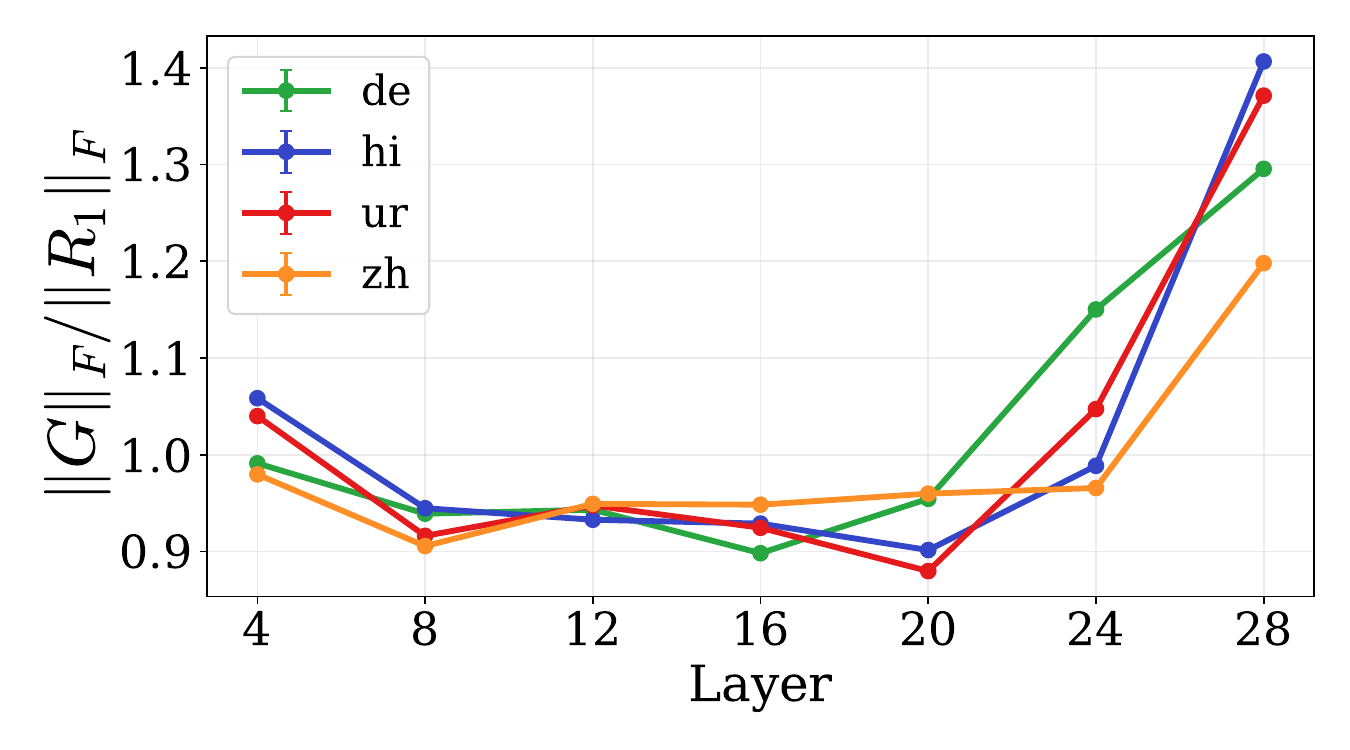}
        \caption{Llama-3.1-8B}
    \end{subfigure}
    \hfill
    \begin{subfigure}{0.48\textwidth}
        \centering
        \includegraphics[width=\linewidth]{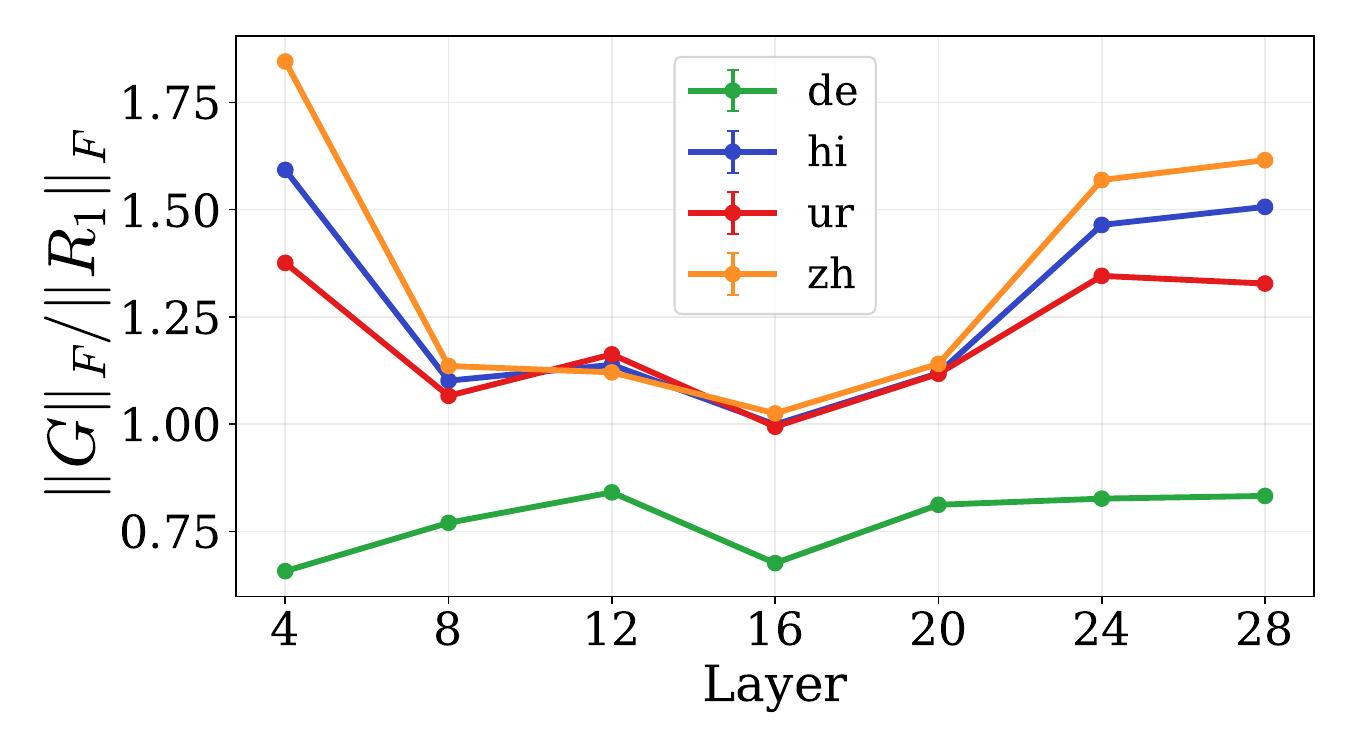}
        \caption{Gemma-3-4B}
    \end{subfigure}
    \caption{\textbf{MEMIT Relative Batch Value Gap.} The relative Frobenius gap $\|\bm{G}_{1,2}\|_F / \|R_1\|_F$ remains consistently high, mimicking the single-edit ROME failure pattern across all target languages.}
    \label{fig:memit_rel_gap}
\end{figure}

Furthermore, Figure~\ref{fig:memit_rel_gap} plots the relative Frobenius norm of the batch value gap ($\|\bm{G}_{1,2}\|_F / \|R_1\|_F$). The relative error remains persistently high across all layers and diverges sharply at the later layers, indicating that aggregating updates into a batch does not remove the observed value-space mismatch.

\subsection{AlphaEdit Batch Analysis}
\label{app:alphaedit_results}

\begin{figure}[!th]
    \centering
    \begin{subfigure}{0.48\textwidth}
        \centering
        \includegraphics[width=\linewidth]{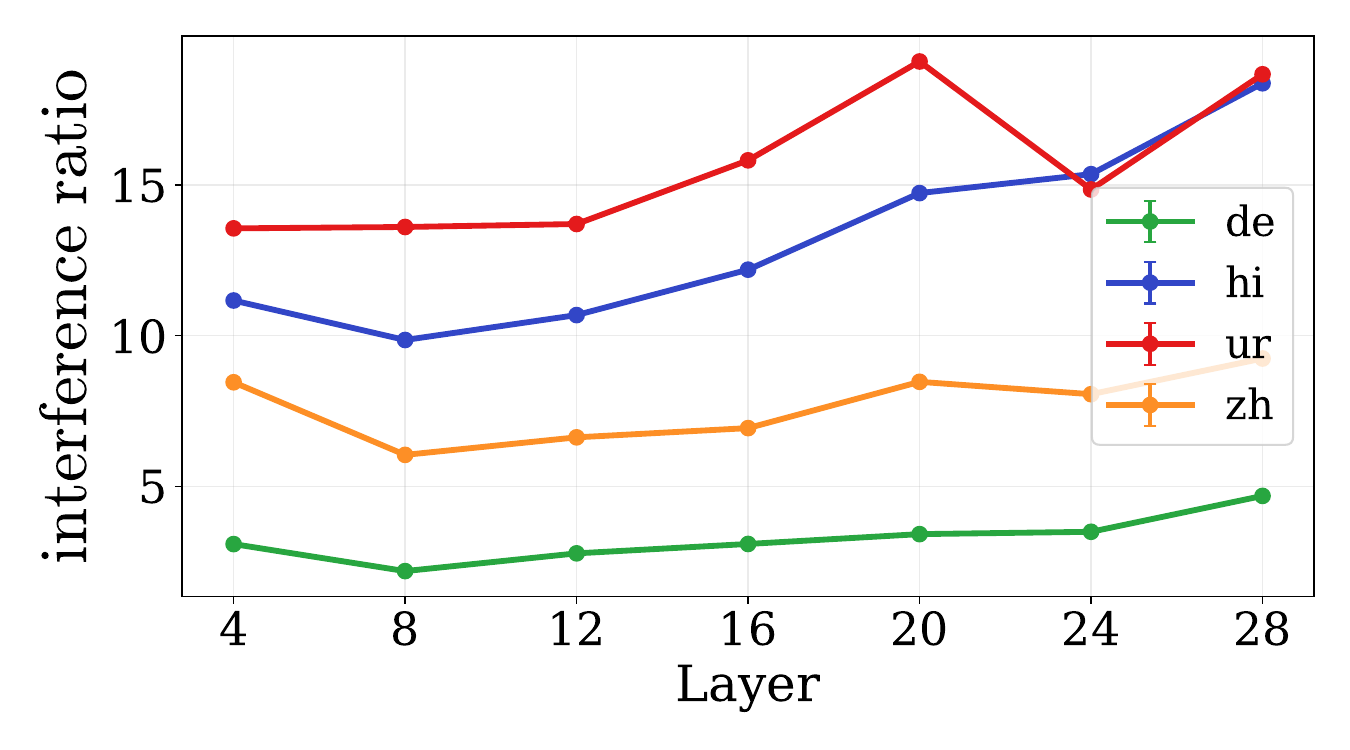}
        \caption{Llama-3.1-8B}
    \end{subfigure}
    \hfill
    \begin{subfigure}{0.48\textwidth}
        \centering
        \includegraphics[width=\linewidth]{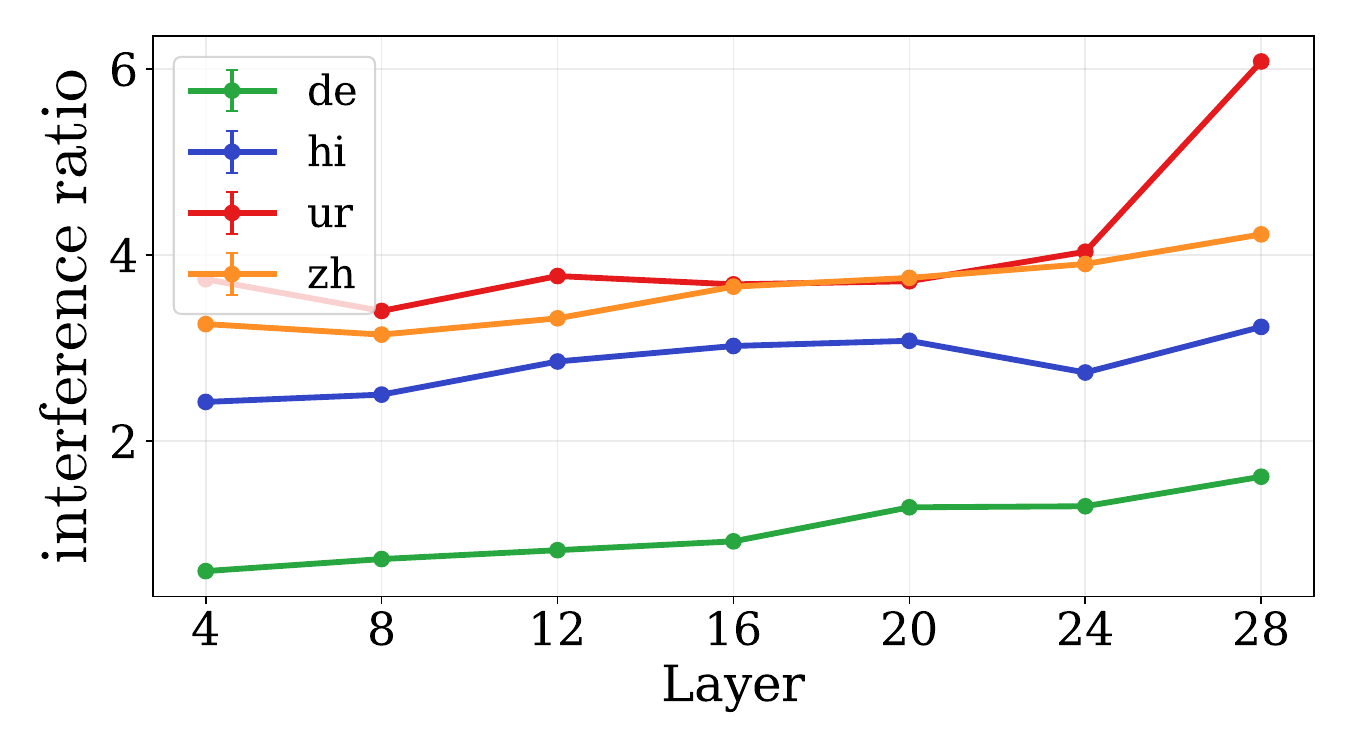}
        \caption{Gemma-3-4B}
    \end{subfigure}
    \caption{\textbf{AlphaEdit Interference Ratio (IR) across layers.} Despite the null-space projection, cross-fact interference remains heavily stratified by typological distance from English.}
    \label{fig:alphaedit_ir}
\end{figure}

AlphaEdit modifies MEMIT by strictly projecting the incoming target keys into the null space of the source key distribution to protect pre-existing knowledge. Figure~\ref{fig:alphaedit_ir} demonstrates the Interference Ratio for AlphaEdit. While the projection inherently alters the specific trajectory compared to unconstrained MEMIT, the core typological hierarchy persists: distant languages still suffer from significantly higher cross-fact leakage than German.

\begin{figure}[!th]
    \centering
    \begin{subfigure}{0.48\textwidth}
        \centering
        \includegraphics[width=\linewidth]{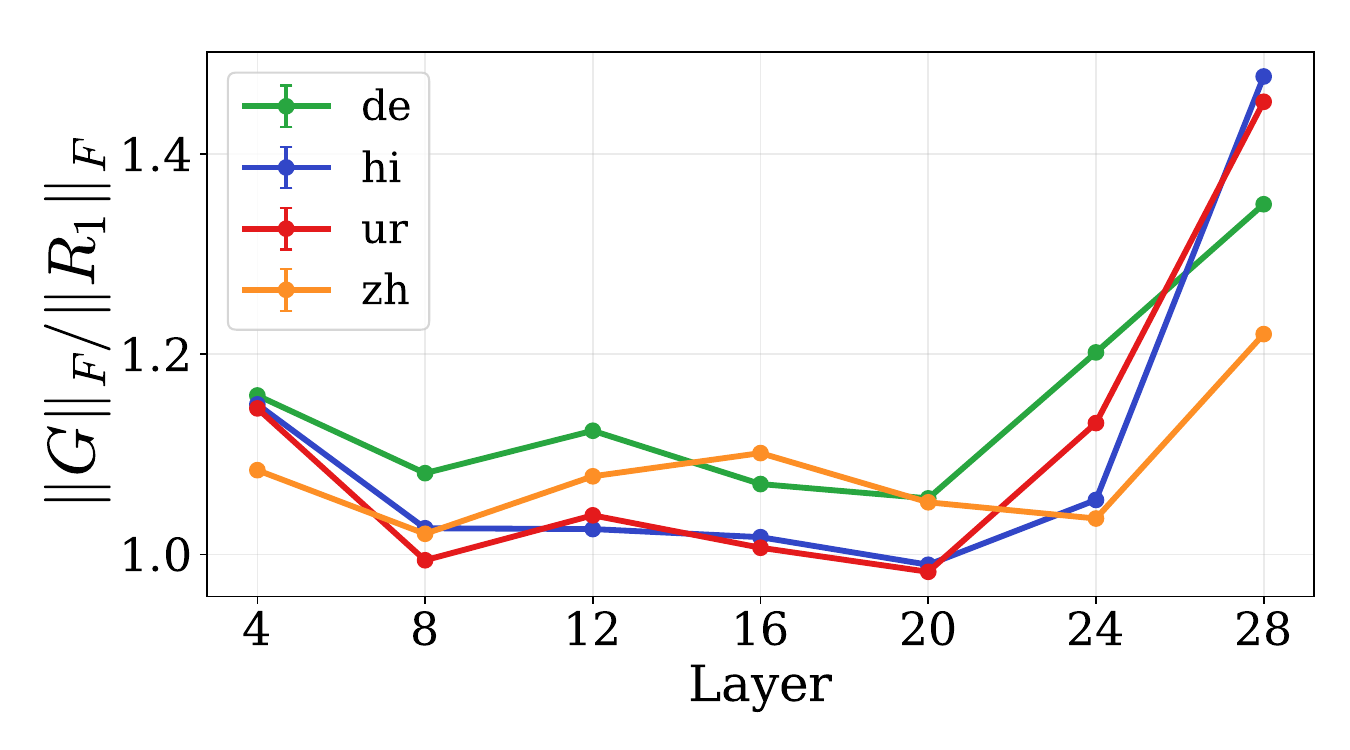}
        \caption{Llama-3.1-8B}
    \end{subfigure}
    \hfill
    \begin{subfigure}{0.48\textwidth}
        \centering
        \includegraphics[width=\linewidth]{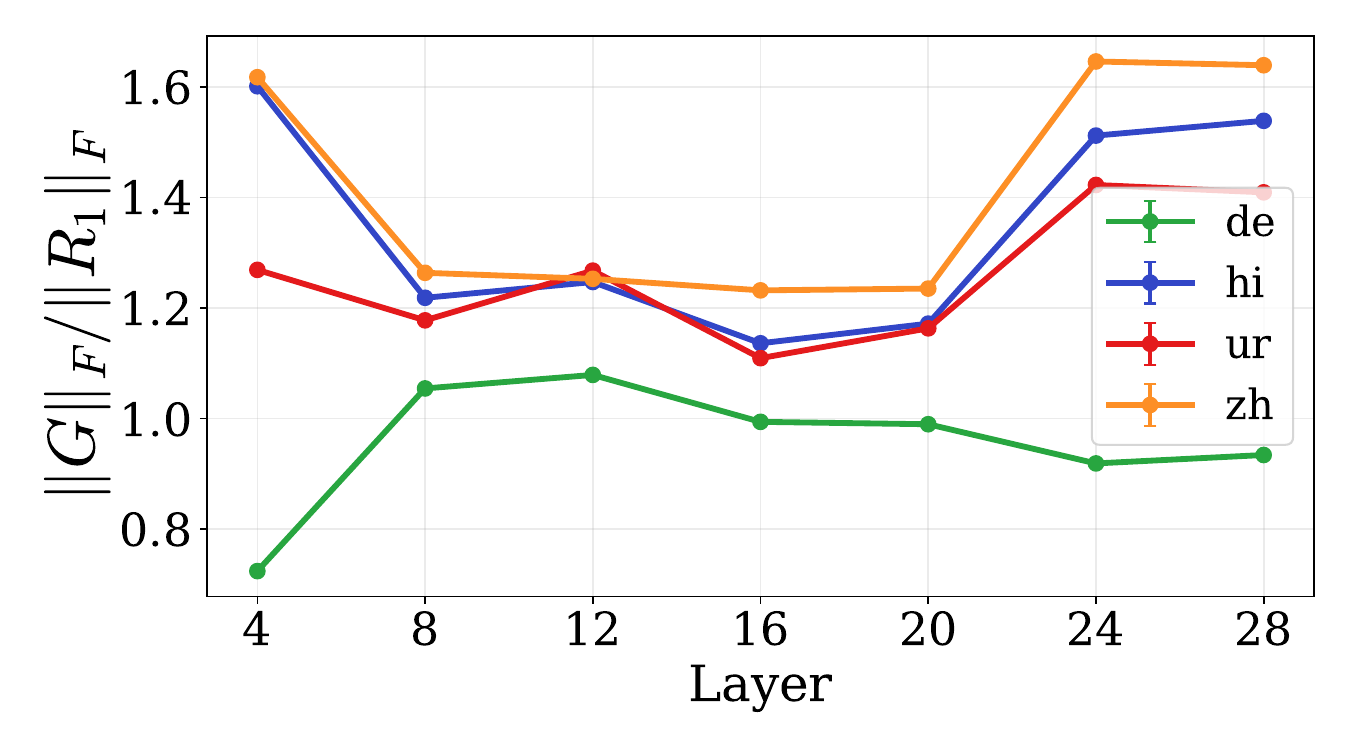}
        \caption{Gemma-3-4B}
    \end{subfigure}
    \caption{\textbf{AlphaEdit Relative Batch Value Gap.} The relative Frobenius gap $\|\bm{G}_{1,2}^{\text{AE}}\|_F / \|R_1\|_F$ confirms that orthogonal misalignment persists despite strict null-space parameter constraints.}
    \label{fig:alphaedit_rel_gap}
\end{figure}

Figure~\ref{fig:alphaedit_rel_gap} is consistent with our analysis that AlphaEdit's key-space projection does not directly
resolve the value-space orthogonal mismatch. The relative batch value gap
($\|\bm{G}_{1,2}^{\text{AE}}\|_F / \|R_1\|_F$) closely follows the MEMIT baseline, suggesting that the orthogonal
term remains a major contributor to transfer failure across both constraint-based and unconstrained
batch editing methods.

\subsection{Cross-Lingual Performance of Edit Methods}
\label{app:nll-margin}
To evaluate editing performance, we measure the change in the log-likelihood margin between the new target answer ($o^*$) and the original answer ($o^c$), following the relative-probability comparison used in ROME~\citep{rome}. To distinguish this from the target-answer NLL reduction in Definition~3, we denote this metric by $\Delta\mathrm{NLL}_{\mathrm{margin}}$:

\begin{equation}
\begin{aligned}
\Delta\mathrm{NLL}_{\mathrm{margin}}
&= \left[\mathrm{NLL}_{\mathrm{pre}}(o^*)-\mathrm{NLL}_{\mathrm{pre}}(o^c)\right]\\
&\quad-\left[\mathrm{NLL}_{\mathrm{post}}(o^*)-\mathrm{NLL}_{\mathrm{post}}(o^c)\right]\\
&= \log\frac{P_{\mathrm{post}}(o^*\mid s,r)}{P_{\mathrm{post}}(o^c\mid s,r)}
-\log\frac{P_{\mathrm{pre}}(o^*\mid s,r)}{P_{\mathrm{pre}}(o^c\mid s,r)}.
\end{aligned}
\end{equation}

A positive value indicates that editing increases the model's relative preference for the target answer over the original answer. Table~\ref{tab:baselines_numbers} provides an exhaustive cross-lingual grid of $\Delta\mathrm{NLL}_{\mathrm{margin}}$ scores across all evaluated methods (ROME, MEMIT, AlphaEdit), architectures, and language pairs for the CounterFact and zsRE datasets. 

These results empirically validate our theoretical framework. For ROME on both models and for AlphaEdit on Llama-3.1-8B, native target language edits (the diagonal) generally yield the largest $\Delta$NLL gains, while, zero-shot cross-lingual edits from English to distant languages (off-diagonal entries, e.g., en $\to$ zh or en $\to$ ur) suffer from catastrophic signal decay. This decay frequently results in near-zero or negative efficacy, confirming that the key-space gating and orthogonal value-space mismatch effectively nullify the parameter update before it can mathematically shift the target language output distribution.
\clearpage
\begin{table}[H]
\centering
\scriptsize
\begin{tabular}{@{}ccccccccc@{}}
\toprule
Method                                           & Model                                                  & Dataset                                           & \textit{\begin{tabular}[c]{@{}c@{}}\textit{Edit Lang ↓} \textbackslash \\ \textbf{Eval Lang →} \end{tabular}} & \textbf{en}      & \textbf{de}      & \textbf{hi}      & \textbf{ur}      & \textbf{zh}      \\ \midrule
\multicolumn{1}{c|}{\multirow{20}{*}{\rotatebox{90}{ROME}}}      & \multicolumn{1}{c|}{\multirow{10}{*}{\textit{Gemma3-4B}}} & \multicolumn{1}{c|}{\multirow{5}{*}{CounterFact}} & \textit{en}                                                                                                   & \textbf{2.5839}  & 0.6413           & 0.2455           & 0.0368           & 0.0251          \\
\multicolumn{1}{c|}{}                            & \multicolumn{1}{c|}{}                                  & \multicolumn{1}{c|}{}                             & \textit{de}                                                                                                   & 2.0933           & \textbf{2.7755}  & 0.0401           & 0.0009           & -0.0265         \\
\multicolumn{1}{c|}{}                            & \multicolumn{1}{c|}{}                                  & \multicolumn{1}{c|}{}                             & \textit{hi}                                                                                                   & 0.4531           & 0.3637           & \textbf{0.7743}  & 0.0872           & -0.0208         \\
\multicolumn{1}{c|}{}                            & \multicolumn{1}{c|}{}                                  & \multicolumn{1}{c|}{}                             & \textit{ur}                                                                                                   & 0.1092           & 0.1264           & 0.0531           & \textbf{0.3397}  & -0.0277         \\
\multicolumn{1}{c|}{}                            & \multicolumn{1}{c|}{}                                  & \multicolumn{1}{c|}{}                             & \textit{zh}                                                                                                   & 0.8474           & 0.5782           & 0.1661           & 0.0272           & \textbf{0.9829} \\ \cmidrule(l){3-9} 
\multicolumn{1}{c|}{}                            & \multicolumn{1}{c|}{}                                  & \multicolumn{1}{c|}{\multirow{5}{*}{zsRE}}        & \textit{en}                                                                                                   & \textbf{1.2839}  & 0.4874           & -0.0158          & -0.0933          & -0.0891         \\
\multicolumn{1}{c|}{}                            & \multicolumn{1}{c|}{}                                  & \multicolumn{1}{c|}{}                             & \textit{de}                                                                                                   & 0.7927           & \textbf{1.2738}  & -0.0675          & -0.0320          & -0.0810         \\
\multicolumn{1}{c|}{}                            & \multicolumn{1}{c|}{}                                  & \multicolumn{1}{c|}{}                             & \textit{hi}                                                                                                   & 0.2418           & 0.2881           & \textbf{0.5410}  & 0.0824           & 0.0076          \\
\multicolumn{1}{c|}{}                            & \multicolumn{1}{c|}{}                                  & \multicolumn{1}{c|}{}                             & \textit{ur}                                                                                                   & 0.2106           & 0.1966           & 0.1181           & \textbf{0.3078}  & 0.0859          \\
\multicolumn{1}{c|}{}                            & \multicolumn{1}{c|}{}                                  & \multicolumn{1}{c|}{}                             & \textit{zh}                                                                                                   & 0.3059           & 0.2733           & -0.0539          & -0.0527          & \textbf{0.7325} \\ \cmidrule(l){2-9} 
\multicolumn{1}{c|}{}                            & \multicolumn{1}{c|}{\multirow{10}{*}{\textit{Llama3.1-8B}}} & \multicolumn{1}{c|}{\multirow{5}{*}{CounterFact}} & \textit{en}                                                                                                   & 0.2435           & \textbf{0.3758}  & -0.0200          & -0.0344          & -0.1970         \\
\multicolumn{1}{c|}{}                            & \multicolumn{1}{c|}{}                                  & \multicolumn{1}{c|}{}                             & \textit{de}                                                                                                   & 0.8795           & \textbf{1.2959}  & 0.0923           & 0.0569           & -0.0039         \\
\multicolumn{1}{c|}{}                            & \multicolumn{1}{c|}{}                                  & \multicolumn{1}{c|}{}                             & \textit{hi}                                                                                                   & 0.0081           & 0.0307           & \textbf{0.7345}  & 0.0666           & -0.0019         \\
\multicolumn{1}{c|}{}                            & \multicolumn{1}{c|}{}                                  & \multicolumn{1}{c|}{}                             & \textit{ur}                                                                                                   & 0.0130           & 0.0124           & 0.0757           & \textbf{1.1242}  & -0.0016         \\
\multicolumn{1}{c|}{}                            & \multicolumn{1}{c|}{}                                  & \multicolumn{1}{c|}{}                             & \textit{zh}                                                                                                   & 0.1186           & 0.0700           & 0.0200           & 0.0189           & \textbf{0.4986} \\ \cmidrule(l){3-9} 
\multicolumn{1}{c|}{}                            & \multicolumn{1}{c|}{}                                  & \multicolumn{1}{c|}{\multirow{5}{*}{zsRE}}        & \textit{en}                                                                                                   & \textbf{0.9411}  & 0.3737           & 0.0504           & 0.0106           & -0.0240         \\
\multicolumn{1}{c|}{}                            & \multicolumn{1}{c|}{}                                  & \multicolumn{1}{c|}{}                             & \textit{de}                                                                                                   & 0.5673           & \textbf{1.2357}  & 0.1026           & 0.0675           & 0.0581          \\
\multicolumn{1}{c|}{}                            & \multicolumn{1}{c|}{}                                  & \multicolumn{1}{c|}{}                             & \textit{hi}                                                                                                   & 0.0108           & 0.0090           & \textbf{1.0800}  & 0.0596           & 0.0013          \\
\multicolumn{1}{c|}{}                            & \multicolumn{1}{c|}{}                                  & \multicolumn{1}{c|}{}                             & \textit{ur}                                                                                                   & 0.0119           & 0.0051           & 0.0794           & \textbf{1.3608}  & 0.0052          \\
\multicolumn{1}{c|}{}                            & \multicolumn{1}{c|}{}                                  & \multicolumn{1}{c|}{}                             & \textit{zh}                                                                                                   & 0.1208           & 0.1032           & 0.0706           & 0.0588           & \textbf{1.0878} \\ \midrule
\multicolumn{1}{c|}{\multirow{20}{*}{\rotatebox{90}{MEMIT}}}     & \multicolumn{1}{c|}{\multirow{10}{*}{\textit{Gemma3-4B}}} & \multicolumn{1}{c|}{\multirow{5}{*}{CounterFact}} & \textit{en}                                                                                                   & \textbf{0.0155}  & -0.3353          & -0.6230          & -0.6626          & -1.0715         \\
\multicolumn{1}{c|}{}                            & \multicolumn{1}{c|}{}                                  & \multicolumn{1}{c|}{}                             & \textit{de}                                                                                                   & \textbf{-0.2431} & -0.8298          & -1.4111          & -1.3304          & -1.7529         \\
\multicolumn{1}{c|}{}                            & \multicolumn{1}{c|}{}                                  & \multicolumn{1}{c|}{}                             & \textit{hi}                                                                                                   & \textbf{-0.0270} & -0.7947          & -1.0354          & -0.9675          & -1.7719         \\
\multicolumn{1}{c|}{}                            & \multicolumn{1}{c|}{}                                  & \multicolumn{1}{c|}{}                             & \textit{ur}                                                                                                   & \textbf{0.4849}  & -0.5782          & -1.0932          & 0.0586           & -1.1830         \\
\multicolumn{1}{c|}{}                            & \multicolumn{1}{c|}{}                                  & \multicolumn{1}{c|}{}                             & \textit{zh}                                                                                                   & -0.7105          & \textbf{-0.2396} & -2.8411          & -2.9700          & -0.9154         \\ \cmidrule(l){3-9} 
\multicolumn{1}{c|}{}                            & \multicolumn{1}{c|}{}                                  & \multicolumn{1}{c|}{\multirow{5}{*}{zsRE}}        & \textit{en}                                                                                                   & -0.9670          & -0.9641          & -1.1780          & \textbf{-0.7509} & -1.2210         \\
\multicolumn{1}{c|}{}                            & \multicolumn{1}{c|}{}                                  & \multicolumn{1}{c|}{}                             & \textit{de}                                                                                                   & -1.9817          & -1.9181          & -1.9502          & \textbf{-1.6717} & -2.1525         \\
\multicolumn{1}{c|}{}                            & \multicolumn{1}{c|}{}                                  & \multicolumn{1}{c|}{}                             & \textit{hi}                                                                                                   & -0.7664          & -0.7329          & -0.6153          & \textbf{-0.4490} & -1.1768         \\
\multicolumn{1}{c|}{}                            & \multicolumn{1}{c|}{}                                  & \multicolumn{1}{c|}{}                             & \textit{ur}                                                                                                   & -1.1290          & -1.2117          & -0.9521          & \textbf{-0.5387} & -1.1125         \\
\multicolumn{1}{c|}{}                            & \multicolumn{1}{c|}{}                                  & \multicolumn{1}{c|}{}                             & \textit{zh}                                                                                                   & \textbf{-0.8242} & -0.9036          & -1.7648          & -1.4006          & -1.2735         \\ \cmidrule(l){2-9} 
\multicolumn{1}{c|}{}                            & \multicolumn{1}{c|}{\multirow{10}{*}{\textit{Llama3.1-8B}}} & \multicolumn{1}{c|}{\multirow{5}{*}{CounterFact}} & \textit{en}                                                                                                   & -8.9126          & -1.6937          & \textbf{-0.0167} & -0.0490          & -0.0940         \\
\multicolumn{1}{c|}{}                            & \multicolumn{1}{c|}{}                                  & \multicolumn{1}{c|}{}                             & \textit{de}                                                                                                   & \textbf{0.1211}  & -6.7239          & 0.0569           & -0.0034          & -0.3460         \\
\multicolumn{1}{c|}{}                            & \multicolumn{1}{c|}{}                                  & \multicolumn{1}{c|}{}                             & \textit{hi}                                                                                                   & 0.1544           & \textbf{0.2201}  & -0.0053          & 0.1629           & 0.2112          \\
\multicolumn{1}{c|}{}                            & \multicolumn{1}{c|}{}                                  & \multicolumn{1}{c|}{}                             & \textit{ur}                                                                                                   & \textbf{0.0323}  & 0.0182           & -0.1078          & -0.6198          & -0.0408         \\
\multicolumn{1}{c|}{}                            & \multicolumn{1}{c|}{}                                  & \multicolumn{1}{c|}{}                             & \textit{zh}                                                                                                   & 0.2963           & \textbf{0.3314}  & -0.1068          & -0.0329          & -0.6674         \\ \cmidrule(l){3-9} 
\multicolumn{1}{c|}{}                            & \multicolumn{1}{c|}{}                                  & \multicolumn{1}{c|}{\multirow{5}{*}{zsRE}}        & \textit{en}                                                                                                   & -3.2067          & -0.4168          & \textbf{0.0362}  & 0.0359           & 0.0336          \\
\multicolumn{1}{c|}{}                            & \multicolumn{1}{c|}{}                                  & \multicolumn{1}{c|}{}                             & \textit{de}                                                                                                   & \textbf{0.0786}  & -2.4212          & 0.0146           & 0.0000           & 0.0417          \\
\multicolumn{1}{c|}{}                            & \multicolumn{1}{c|}{}                                  & \multicolumn{1}{c|}{}                             & \textit{hi}                                                                                                   & 0.0084           & \textbf{0.0491}  & -0.0096          & 0.0108           & -0.0070         \\
\multicolumn{1}{c|}{}                            & \multicolumn{1}{c|}{}                                  & \multicolumn{1}{c|}{}                             & \textit{ur}                                                                                                   & 0.0091           & \textbf{0.0322}  & -0.0423          & 0.0235           & -0.0151         \\
\multicolumn{1}{c|}{}                            & \multicolumn{1}{c|}{}                                  & \multicolumn{1}{c|}{}                             & \textit{zh}                                                                                                   & 0.0367           & \textbf{0.1039}  & 0.0948           & 0.0475           & 0.0740          \\ \midrule
\multicolumn{1}{c|}{\multirow{20}{*}{\rotatebox{90}{AlphaEdit}}} & \multicolumn{1}{c|}{\multirow{10}{*}{\textit{Gemma3-4B}}} & \multicolumn{1}{c|}{\multirow{5}{*}{CounterFact}} & \textit{en}                                                                                                   & \textbf{13.8083} & 12.6151          & 8.6437           & 9.8539           & 6.0643          \\
\multicolumn{1}{c|}{}                            & \multicolumn{1}{c|}{}                                  & \multicolumn{1}{c|}{}                             & \textit{de}                                                                                                   & 2.9956           & 3.8075           & 3.5700           & 3.1938           & \textbf{4.1834} \\
\multicolumn{1}{c|}{}                            & \multicolumn{1}{c|}{}                                  & \multicolumn{1}{c|}{}                             & \textit{hi}                                                                                                   & \textbf{5.9524}  & 4.8222           & 5.6935           & 1.7496           & 2.2863          \\
\multicolumn{1}{c|}{}                            & \multicolumn{1}{c|}{}                                  & \multicolumn{1}{c|}{}                             & \textit{ur}                                                                                                   & 12.7915          & \textbf{13.1668} & 8.2788           & 1.9107           & 4.6223          \\
\multicolumn{1}{c|}{}                            & \multicolumn{1}{c|}{}                                  & \multicolumn{1}{c|}{}                             & \textit{zh}                                                                                                   & 8.2578           & 8.4854           & \textbf{9.4360}  & 8.0885           & 4.3435          \\ \cmidrule(l){3-9} 
\multicolumn{1}{c|}{}                            & \multicolumn{1}{c|}{}                                  & \multicolumn{1}{c|}{\multirow{5}{*}{zsRE}}        & \textit{en}                                                                                                   & -0.0662          & -0.0444          & \textbf{0.7284}  & 0.4518           & 0.1774          \\
\multicolumn{1}{c|}{}                            & \multicolumn{1}{c|}{}                                  & \multicolumn{1}{c|}{}                             & \textit{de}                                                                                                   & 3.9829           & 4.8624           & \textbf{6.8815}  & 3.8235           & 6.3252          \\
\multicolumn{1}{c|}{}                            & \multicolumn{1}{c|}{}                                  & \multicolumn{1}{c|}{}                             & \textit{hi}                                                                                                   & 5.6332           & 4.9963           & \textbf{5.9510}  & 4.4652           & 5.3812          \\
\multicolumn{1}{c|}{}                            & \multicolumn{1}{c|}{}                                  & \multicolumn{1}{c|}{}                             & \textit{ur}                                                                                                   & 3.7817           & \textbf{4.6340}  & 4.6209           & 3.4223           & 3.6607          \\
\multicolumn{1}{c|}{}                            & \multicolumn{1}{c|}{}                                  & \multicolumn{1}{c|}{}                             & \textit{zh}                                                                                                   & 1.8118           & 2.3947           & \textbf{3.4995}  & 3.3888           & 2.7104          \\ \cmidrule(l){2-9} 
\multicolumn{1}{c|}{}                            & \multicolumn{1}{c|}{\multirow{10}{*}{\textit{Llama3.1-8B}}} & \multicolumn{1}{c|}{\multirow{5}{*}{CounterFact}} & \textit{en}                                                                                                   & \textbf{7.8645}  & 4.6289           & 0.2879           & 0.1759           & 0.1612          \\
\multicolumn{1}{c|}{}                            & \multicolumn{1}{c|}{}                                  & \multicolumn{1}{c|}{}                             & \textit{de}                                                                                                   & 1.2134           & \textbf{2.0783}  & 0.4059           & 0.5619           & 0.5905          \\
\multicolumn{1}{c|}{}                            & \multicolumn{1}{c|}{}                                  & \multicolumn{1}{c|}{}                             & \textit{hi}                                                                                                   & 0.2679           & 0.2755           & \textbf{0.8161}  & 0.7054           & 0.4045          \\
\multicolumn{1}{c|}{}                            & \multicolumn{1}{c|}{}                                  & \multicolumn{1}{c|}{}                             & \textit{ur}                                                                                                   & 0.3079           & 0.1155           & 0.3128           & \textbf{0.6811}  & 0.1063          \\
\multicolumn{1}{c|}{}                            & \multicolumn{1}{c|}{}                                  & \multicolumn{1}{c|}{}                             & \textit{zh}                                                                                                   & 0.7634           & 0.6974           & 0.4956           & 0.7086           & \textbf{1.1890} \\ \cmidrule(l){3-9} 
\multicolumn{1}{c|}{}                            & \multicolumn{1}{c|}{}                                  & \multicolumn{1}{c|}{\multirow{5}{*}{zsRE}}        & \textit{en}                                                                                                   & \textbf{1.1832}  & 0.8488           & 0.1030           & 0.2201           & 0.3089          \\
\multicolumn{1}{c|}{}                            & \multicolumn{1}{c|}{}                                  & \multicolumn{1}{c|}{}                             & \textit{de}                                                                                                   & 0.6164           & \textbf{1.2629}  & 0.0446           & 0.0342           & 0.1625          \\
\multicolumn{1}{c|}{}                            & \multicolumn{1}{c|}{}                                  & \multicolumn{1}{c|}{}                             & \textit{hi}                                                                                                   & 0.0468           & 0.0598           & \textbf{0.3043}  & 0.1750           & -0.0214         \\
\multicolumn{1}{c|}{}                            & \multicolumn{1}{c|}{}                                  & \multicolumn{1}{c|}{}                             & \textit{ur}                                                                                                   & 0.1600           & 0.0859           & 0.0543           & \textbf{0.3009}  & 0.0342          \\
\multicolumn{1}{c|}{}                            & \multicolumn{1}{c|}{}                                  & \multicolumn{1}{c|}{}                             & \textit{zh}                                                                                                   & 0.2800           & 0.3090           & 0.0671           & 0.0773           & \textbf{0.6793} \\  
\bottomrule
\end{tabular}
\vspace{6pt}
\caption{$\Delta\mathrm{NLL}_{\mathrm{margin}}$ across knowledge editing methods, architectures, and layer configurations. Values measure the change in the log-likelihood margin between the new target answer and the original answer, as defined in Appendix~\ref{app:nll-margin}. Higher values indicate a greater shift toward the target answer.}
\label{tab:baselines_numbers}
\end{table}

\subsection{Is the Orthogonal Mismatch an Artifact of the Edit Layer?}
\label{app:oracle-layer}

The layer at which factual information is localized need not be the best layer at which to edit it~\cite{hase2023localization}. To test whether our conclusions depend on the choice of edit layer, we performed a per-fact oracle analysis for ROME on Llama-3.1-8B with English as the edit language. For 82 facts, we computed the normalized orthogonal mismatch $\|\mathbf{r}_2^{\perp}\|/\|\mathbf{r}_1\|$ at layers 4, 8, 12, 16, 20, 24, and 28, and selected, separately for each fact and target language, the layer that minimizes it. This oracle uses the target-language residual, which is unavailable in source-only editing, so it gives an optimistic estimate of what layer selection alone can achieve.

Table~\ref{tab:oracle-layer} compares the oracle with the fixed layer 20. The oracle reduces the mismatch significantly in every language (paired $t$-tests, all $p < 10^{-10}$), by 0.090 to 0.138, but the mismatch remains large, between 0.774 and 0.801. The best layer also varies widely across facts: for Hindi, it ranges from layer 4 to layer 24, and the most frequently selected layer is best for only 27 of the 82 facts. The orthogonal mismatch is therefore not explained by a single poorly chosen edit layer.

\begin{table}[h]
\centering
\begin{tabular}{lccc}
\toprule
Target & Fixed layer 20 & Per-fact oracle & Improvement \\
\midrule
German  & 0.864 & 0.774 & 0.090 \\
Hindi   & 0.895 & 0.801 & 0.094 \\
Urdu    & 0.872 & 0.775 & 0.097 \\
Chinese & 0.933 & 0.795 & 0.138 \\
\bottomrule
\end{tabular}
\vspace{6pt}
\caption{Per-fact oracle layer versus fixed layer 20 (ROME, Llama-3.1-8B, edit language English, 82 facts). Entries are $\|\mathbf{r}_2^{\perp}\|/\|\mathbf{r}_1\|$ (lower is better). The oracle picks, for each fact, the layer in $\{4, 8, \dots, 28\}$ with the lowest value, using target-language information.}
\label{tab:oracle-layer}
\end{table}

\subsection{Does the Mismatch Depend on Target Length?}
\label{app:target-length}

CounterFact and zsRE contain short answers, and we know of no open parallel multilingual benchmark for long-form knowledge editing that provides semantically matched edits and evaluation prompts across our five languages. We therefore test the closest question that the available data support: whether the orthogonal mismatch changes with target length. We repeated the layer-20 residual-gap analysis (ROME, Llama-3.1-8B) on the largest zsRE subset shared across all five languages: 382 facts whose targets range from 1 to 16 tokens (median 3), grouped by target length (Table~\ref{tab:target-length}).

The difference between the longest and shortest bins is positive for every language and significant only for Chinese (Mann--Whitney test, $p = 0.02$). The correlation between target length and the orthogonal ratio is non-negative for every language ($+0.05$ for German, $+0.03$ for Hindi, $+0.04$ for Urdu, and $+0.11$ for Chinese) and is significant only for Chinese ($p = 0.02$). The parallel coefficient $\gamma$ is likewise stable, rising slightly with length for Hindi (0.24 to 0.28, $p = 0.03$) and Urdu (0.22 to 0.26, $p = 0.06$). Within the 1--16 token range covered by zsRE, the orthogonal mismatch therefore does not decrease for longer targets. This is still a short-answer analysis and does not establish that the same holds for paragraph-length generation, which can reveal properties such as factual drift, internal consistency, and discourse quality that our benchmarks do not measure~\cite{rosati-etal-2024-long}.

\begin{table}[h]
\centering
\begin{tabular}{lccccc}
\toprule
en$\to$ & 1 token (49) & 2 tokens (69) & 3--4 tokens (188) & 5+ tokens (76) & 5+ minus 1 token \\
\midrule
de & 0.827 & 0.906 & 0.904 & 0.854 & $+0.027$ ($p = 0.28$) \\
hi & 0.869 & 0.918 & 0.902 & 0.884 & $+0.015$ ($p = 0.49$) \\
ur & 0.845 & 0.892 & 0.885 & 0.877 & $+0.032$ ($p = 0.21$) \\
zh & 0.892 & 0.947 & 0.953 & 0.972 & $+0.080$ ($p = 0.02$) \\
\bottomrule
\end{tabular}
\vspace{6pt}
\caption{Orthogonal mismatch $\|\mathbf{r}_2^{\perp}\|/\|\mathbf{r}_1\|$ by target length (zsRE, ROME, Llama-3.1-8B, layer 20, 382 facts). Bin sizes are in parentheses. The last column is the difference between the longest and shortest bins, with a Mann--Whitney test.}
\label{tab:target-length}
\end{table}

\subsection{Logically Equivalent Formulations}
\label{app:equivalent-formulations}

\citet{niu2024knowledge} raise two distinct issues for edits to MLP components: they may fail to carry over to logically equivalent expressions of a fact, such as its inverse (``The capital of $X$ is $Y$'' and ``$Y$ is the capital of $X$''), and they can also affect linguistic patterns. We ran a matched experiment that addresses the first issue. For the 21 facts with available bijective formulations, we compared prompts that all express the same edited fact: an English paraphrase, the logically equivalent English formulation obtained by swapping subject and object and using the inverse relation, and translations into the four target languages. Table~\ref{tab:equivalent-formulations} reports $\gamma$ and the orthogonal ratio for each condition.

Reversing the relation preserves the underlying fact, yet the orthogonal ratio rises from 0.718 for an English paraphrase to 0.829 for the inverse formulation. This is a geometric counterpart to the symmetry failure identified by \citet{niu2024knowledge}: even logically equivalent formulations can require substantially different residual directions. The same-language controls help separate sensitivity to the formulation from the additional cross-lingual mismatch. Changing the wording or the direction of the relation within English already produces a sizeable mismatch, and transfer to Hindi, Urdu, and Chinese produces a significantly larger one: pooling the 63 paired observations across these three languages, the mean increase over the English paraphrase is $+0.246$ (Wilcoxon signed-rank test, $p = 1.9 \times 10^{-6}$; bootstrap 95\% CI $[+0.177, +0.322]$).

This experiment does not test whether an edit degrades grammatical or morphological behavior in the target language. We therefore cannot rule out a second pathway, in which an edit alters linguistic behavior and thereby affects cross-lingual transfer~\cite{niu2024knowledge}. Evaluating such side effects directly is left to future work.

\begin{table}[h]
\centering
\begin{tabular}{lccc}
\toprule
Condition & $\gamma$ & $\|\mathbf{r}_2^{\perp}\|/\|\mathbf{r}_1\|$ & Difference from paraphrase \\
\midrule
English paraphrase & $0.556 \pm 0.039$ & $0.718 \pm 0.031$ & -- \\
English inversion  & $0.522 \pm 0.021$ & $0.829 \pm 0.023$ & $+0.111$ ($p = 4.9 \times 10^{-3}$) \\
German             & $0.591 \pm 0.019$ & $0.826 \pm 0.020$ & $+0.108$ ($p = 3.8 \times 10^{-3}$) \\
Urdu               & $0.190 \pm 0.013$ & $0.920 \pm 0.044$ & $+0.202$ ($p = 4.8 \times 10^{-6}$) \\
Hindi              & $0.229 \pm 0.013$ & $0.941 \pm 0.047$ & $+0.223$ ($p = 9.5 \times 10^{-6}$) \\
Chinese            & $0.400 \pm 0.023$ & $1.031 \pm 0.058$ & $+0.313$ ($p = 9.5 \times 10^{-7}$) \\
\bottomrule
\end{tabular}
\vspace{6pt}
\caption{Logically equivalent and translated formulations of the same edited fact (21 facts). Differences are relative to the English paraphrase, with paired Wilcoxon signed-rank tests.}
\label{tab:equivalent-formulations}
\end{table}

%% file: appendix/proofs_impossibility.tex
\section{Source-Only Correction Bound: Proofs and Extended Experiments}
\label{app:impossibility_proofs}

\subsection{Proof of Proposition~\ref{prop:floor} (Source-Only Residual Prediction Barrier)}
\label{app:floor_proof}

\begin{proof}
\textbf{Condition on $\alpha_{1,2}$ as well.} Since $f : \mathbb{R}^D \to \mathbb{R}^D$ is an arbitrary measurable function and $\alpha_{1,2}$ is a scalar determined by the keys and $C$ that varies across facts, $\alpha_{1,2} \cdot f(\mathbf{r}_1)$ is in general not a function of $\mathbf{r}_1$ alone, so $\alpha_{1,2}$ cannot be absorbed into $f$. It is, however, a measurable function of $\mathbf{x} := (\mathbf{r}_1, \alpha_{1,2})$; define $g(\mathbf{x}) = \alpha_{1,2} \cdot f(\mathbf{r}_1)$. As $f$ ranges over all measurable functions, $g$ ranges over a subset of the measurable functions of $\mathbf{x}$. Therefore:
\[
\inf_f \mathbb{E}\big[\|\mathbf{r}_2 - \alpha_{1,2} f(\mathbf{r}_1)\|^2\big] \;\ge\; \inf_g \mathbb{E}\big[\|\mathbf{r}_2 - g(\mathbf{x})\|^2\big].
\]

\textbf{Optimality of conditional expectation.} It is a well known result that the conditional expectation is the optimal estimator minimizing the mean squared error for any random variable (Proposition~5). For any measurable $g$, write:
\[
\mathbf{r}_2 - g(\mathbf{x}) = \underbrace{\big(\mathbf{r}_2 - \mathbb{E}[\mathbf{r}_2 \mid \mathbf{x}]\big)}_{\varepsilon} + \underbrace{\big(\mathbb{E}[\mathbf{r}_2 \mid \mathbf{x}] - g(\mathbf{x})\big)}_{\delta}.
\]
Taking the squared norm and expectation:
\[
\mathbb{E}\big[\|\mathbf{r}_2 - g(\mathbf{x})\|^2\big] = \mathbb{E}\big[\|\varepsilon\|^2\big] + \mathbb{E}\big[\|\delta\|^2\big] + 2\,\mathbb{E}\big[\varepsilon^{\top}\delta\big].
\]
The term $\delta = \mathbb{E}[\mathbf{r}_2 \mid \mathbf{x}] - g(\mathbf{x})$ is a deterministic function of $\mathbf{x}$. The term $\varepsilon = \mathbf{r}_2 - \mathbb{E}[\mathbf{r}_2 \mid \mathbf{x}]$ satisfies $\mathbb{E}[\varepsilon \mid \mathbf{x}] = \mathbf{0}$ by the definition of conditional expectation. Therefore:
\[
\mathbb{E}[\varepsilon^{\top}\delta] = \mathbb{E}\big[\mathbb{E}[\varepsilon^{\top}\delta \mid \mathbf{x}]\big] = \mathbb{E}\big[\mathbb{E}[\varepsilon \mid \mathbf{x}]^{\top}\delta\big] = \mathbb{E}[\mathbf{0}^{\top}\delta] = 0.
\]

\textbf{Final bound.} Since $\mathbb{E}[\|\delta\|^2] \ge 0$:
\[
\mathbb{E}\big[\|\mathbf{r}_2 - g(\mathbf{x})\|^2\big] = \mathbb{E}\big[\|\varepsilon\|^2\big] + \mathbb{E}\big[\|\delta\|^2\big] \;\ge\; \mathbb{E}\big[\|\varepsilon\|^2\big] = \eta(\mathbf{r}_2 \mid \mathbf{r}_1, \alpha_{1,2}) \cdot \operatorname{Var}(\mathbf{r}_2).
\]
Equality holds if and only if $g = \mathbb{E}[\mathbf{r}_2 \mid \mathbf{x}]$ almost surely, i.e., at the optimal predictor. Because $\mathbb{E}[\mathbf{r}_2 \mid \mathbf{r}_1]$ is itself a function of $\mathbf{x}$, the same argument gives $\mathbb{E}[\|\varepsilon\|^2] \le \sigma^2_{1,2}$, i.e., $\eta(\mathbf{r}_2 \mid \mathbf{r}_1, \alpha_{1,2}) \le \eta_{1,2}$. If $\alpha_{1,2}$ carries no information about $\mathbf{r}_2$ beyond $\mathbf{r}_1$, that is, $\mathbb{E}[\mathbf{r}_2 \mid \mathbf{x}] = \mathbb{E}[\mathbf{r}_2 \mid \mathbf{r}_1]$ almost surely, then $\mathbb{E}[\|\varepsilon\|^2] = \sigma^2_{1,2}$, and the identity $\sigma^2_{1,2} = \eta_{1,2} \cdot \operatorname{Var}(\mathbf{r}_2)$ follows from the law of total variance (Appendix~E.4).
\end{proof}

\paragraph{Remark.} The assumption is needed for the form with $\eta_{1,2}$. For example, if $\mathbf{r}_2 = \alpha_{1,2}\mathbf{r}_1$ with $\alpha_{1,2}$ independent of $\mathbf{r}_1$ and not constant, the choice $f(\mathbf{r}_1) = \mathbf{r}_1$ gives zero error, whereas $\sigma^2_{1,2} = \operatorname{Var}(\alpha_{1,2})\,\mathbb{E}\|\mathbf{r}_1\|^2 > 0$.

\subsection{Sample and Population $R^2$}
\label{app:sample_pop_r2}

We distinguish two notions of the coefficient of determination that play different roles in our analysis.

\paragraph{Sample $R^2$ (for a specific estimator).}
Given a held-out test set $\{(\bm{r}_1^{(i)}, \bm{r}_2^{(i)})\}_{i=1}^{n}$ and a trained estimator $\hat{f}: \mathbb{R}^D \to \mathbb{R}^D$, the sample coefficient of determination is
\begin{equation}\label{eq:sample_r2}
    R^2_{\hat{f}} \;=\; 1 \;-\; \frac{\sum_{i=1}^{n} \left\|\bm{r}_2^{(i)} - \hat{f}(\bm{r}_1^{(i)})\right\|^2}{\sum_{i=1}^{n} \left\|\bm{r}_2^{(i)} - \bar{\bm{r}}_2\right\|^2},
\end{equation}
where $\bar{\bm{r}}_2 = \frac{1}{n}\sum_{i}\bm{r}_2^{(i)}$ is the sample mean of the target residuals. This quantity is computed on finite held-out data and measures how well $\hat{f}$ predicts $\bm{r}_2$ relative to the trivial baseline of always predicting $\bar{\bm{r}}_2$. When $R^2_{\hat{f}} \approx 0$, the estimator $\hat{f}$ explains essentially no more variance than the mean; when $R^2_{\hat{f}} \approx 1$, the predictions are near-perfect.

\paragraph{Population $R^2$ (theoretical optimum).}
The population coefficient of determination is defined over the true joint distribution of $(\bm{r}_1, \bm{r}_2)$:
\begin{equation}\label{eq:pop_r2}
    R^2_{\mathrm{pop}} \;=\; 1 \;-\; \frac{\mathbb{E}\!\left[\left\|\bm{r}_2 - \mathbb{E}[\bm{r}_2 \mid \bm{r}_1]\right\|^2\right]}{\mathbb{E}\!\left[\left\|\bm{r}_2 - \mathbb{E}[\bm{r}_2]\right\|^2\right]} \;=\; 1 \;-\; \frac{\sigma_{1,2}^2}{\mathrm{Var}(\bm{r}_2)}.
\end{equation}
This uses the \emph{optimal} predictor $g^*(\bm{r}_1) = \mathbb{E}[\bm{r}_2 \mid \bm{r}_1]$ (the conditional expectation; see Appendix~\ref{app:optimal_estimator}), which minimizes the expected squared error over \emph{all} measurable functions. The population $R^2$ is a theoretical quantity that cannot be computed exactly, since neither the true conditional expectation nor the true joint distribution is available.

\paragraph{Key relationship: $R^2_{\hat f,\mathrm{pop}}\leq R^2_{\mathrm{pop}}$.}
For any fitted predictor $\hat f$, its population mean squared error cannot be lower than that of the optimal conditional expectation. Restricted function classes and finite training data may further limit predictive performance. More precisely, at the population level:

\begin{equation}
\label{eq:r2-pop}
    R^2_{\hat{f},\mathrm{pop}} \;:=\; 1 - \frac{\mathbb{E}\!\left[\left\|\bm{r}_2 - \hat{f}(\bm{r}_1)\right\|^2\right]}{\mathrm{Var}(\bm{r}_2)} \;\leq\; 1 - \frac{\mathbb{E}\!\left[\left\|\bm{r}_2 - \mathbb{E}[\bm{r}_2 \mid \bm{r}_1]\right\|^2\right]}{\mathrm{Var}(\bm{r}_2)} \;=\; R^2_{\mathrm{pop}},
\end{equation}
where the inequality follows from the optimality of the conditional expectation (Appendix~\ref{app:optimal_estimator}). The sample $R^2_{\hat{f}}$ computed on held-out data is a consistent estimate of $R^2_{\hat{f},\mathrm{pop}}$ as $n \to \infty$, so for large enough test sets we have $R^2_{\hat{f}} \approx R^2_{\hat{f},\mathrm{pop}} \leq R^2_{\mathrm{pop}}$.

\paragraph{Consequence for the private fraction.}
For a fixed fitted predictor $\hat f$, Eq.~(\ref{eq:r2-pop}) implies the following population-level relationship:
\begin{equation}
\eta_{\hat f,\mathrm{pop}}
:= 1-R^2_{\hat f,\mathrm{pop}}
\geq 1-R^2_{\mathrm{pop}}
=\eta.
\end{equation}
The empirical quantity $\hat{\eta}=1-R^2_{\hat f}$ estimates $\eta_{\hat f,\mathrm{pop}}$ on finite held-out data. Due to sampling variation, the inequality need not hold for every observed $\hat{\eta}$. Thus, a high $\hat{\eta}$ indicates limited predictability under the evaluated estimator, but does not establish a numerical lower bound on the population-optimal unexplained fraction $\eta$.

\paragraph{Why convergence across estimators suggests tightness.}
A single estimator yielding high $\hat{\eta}$ could, in principle, simply be the wrong function class for the problem. To guard against this, we employ three estimators with fundamentally different inductive biases: Ridge regression (global linear structure), a two-layer MLP (smooth nonlinear structure via universal approximation), and $k$-nearest neighbors (nonparametric, locally adaptive). If one class were missing exploitable structure that another could capture, their $R^2$ values would diverge. Instead, all three converge to nearly identical values, with $R^2 \le 0.04$ at the editing layer (layer 20) and $R^2 \le 0.09$ at every layer (Table~\ref{tab:r2_layerwise_llama}). While this convergence does not constitute a formal proof that $R^2_{\mathrm{pop}} \approx R^2_{\hat{f}}$ --it is logically possible that some exotic function class could achieve higher $R^2$ -- it provides strong empirical evidence. The agreement across linear, nonlinear, and nonparametric approaches indicates that the weak predictability is not an artifact of any particular modeling choice.

\subsection{The Optimal Estimator Theorem}
\label{app:optimal_estimator}

The following is a classical result in estimation theory that forms the foundation for Proposition~\ref{prop:floor}. We state and prove it here for completeness.

\begin{proposition}[Optimality of the conditional expectation]\label{thm:optimal_estimator}
    Let $\bm{r}_1 \in \mathbb{R}^D$ and $\bm{r}_2 \in \mathbb{R}^D$ be random vectors with $\mathbb{E}[\|\bm{r}_2\|^2] < \infty$. Among all measurable functions $g: \mathbb{R}^D \to \mathbb{R}^D$, the conditional expectation $g^*(\bm{r}_1) = \mathbb{E}[\bm{r}_2 \mid \bm{r}_1]$ is the unique minimizer of the mean squared error:
    \begin{equation}
        g^* \;=\; \arg\min_{g} \;\mathbb{E}\!\left[\left\|\bm{r}_2 - g(\bm{r}_1)\right\|^2\right],
    \end{equation}
    and the minimum value is the conditional variance:
    \begin{equation}
        \min_{g}\;\mathbb{E}\!\left[\left\|\bm{r}_2 - g(\bm{r}_1)\right\|^2\right] \;=\; \mathbb{E}\!\left[\left\|\bm{r}_2 - \mathbb{E}[\bm{r}_2 \mid \bm{r}_1]\right\|^2\right] \;=:\; \sigma_{1,2}^2.
    \end{equation}
\end{proposition}

\begin{proof}
    Let $g: \mathbb{R}^D \to \mathbb{R}^D$ be any measurable function. Define:
    \begin{align}
        \bm{\epsilon} &\;:=\; \bm{r}_2 - \mathbb{E}[\bm{r}_2 \mid \bm{r}_1], \label{eq:eps_def}\\
        \bm{\delta} &\;:=\; \mathbb{E}[\bm{r}_2 \mid \bm{r}_1] - g(\bm{r}_1). \label{eq:delta_def}
    \end{align}
    Note that $\epsilon$ is the part of $\mathbf{r}_2$ that no function of $\mathbf{r}_1$ can predict (the deviation of $\mathbf{r}_2$ from its conditional mean), while $\delta$ is the suboptimality of $g$ relative to the oracle. We can write the total error as:
    \begin{equation}
        \bm{r}_2 - g(\bm{r}_1) \;=\; \bm{\epsilon} + \bm{\delta}.
    \end{equation}
    Expanding the squared norm:
    \begin{equation}\label{eq:mse_expansion}
        \mathbb{E}\!\left[\left\|\bm{r}_2 - g(\bm{r}_1)\right\|^2\right] \;=\; \mathbb{E}\!\left[\|\bm{\epsilon}\|^2\right] + \mathbb{E}\!\left[\|\bm{\delta}\|^2\right] + 2\,\mathbb{E}\!\left[\bm{\epsilon}^\top \bm{\delta}\right].
    \end{equation}
    
    \emph{The cross-term vanishes.}\quad The key observation is that $\bm{\delta} = \mathbb{E}[\bm{r}_2 \mid \bm{r}_1] - g(\bm{r}_1)$ is a deterministic function of $\bm{r}_1$ (since both $\mathbb{E}[\bm{r}_2 \mid \bm{r}_1]$ and $g(\bm{r}_1)$ are $\bm{r}_1$-measurable). Meanwhile, by the defining property of the conditional expectation:
    \begin{equation}
        \mathbb{E}[\bm{\epsilon} \mid \bm{r}_1] \;=\; \mathbb{E}[\bm{r}_2 \mid \bm{r}_1] - \mathbb{E}[\bm{r}_2 \mid \bm{r}_1] \;=\; \bm{0}.
    \end{equation}
    Therefore, applying the tower property (law of iterated expectations):
    \begin{equation}
        \mathbb{E}\!\left[\bm{\epsilon}^\top \bm{\delta}\right] \;=\; \mathbb{E}\!\left[\mathbb{E}\!\left[\bm{\epsilon}^\top \bm{\delta} \mid \bm{r}_1\right]\right] \;=\; \mathbb{E}\!\left[\mathbb{E}\!\left[\bm{\epsilon} \mid \bm{r}_1\right]^\top \bm{\delta}\right] \;=\; \mathbb{E}\!\left[\bm{0}^\top \bm{\delta}\right] \;=\; 0,
    \end{equation}
    where we used the fact that $\bm{\delta}$ can be pulled out of the inner conditional expectation since it is $\bm{r}_1$-measurable.
    
    \emph{Conclusion.}\quad Substituting back into Eq.~\eqref{eq:mse_expansion}:
    \begin{equation}
        \mathbb{E}\!\left[\left\|\bm{r}_2 - g(\bm{r}_1)\right\|^2\right] \;=\; \underbrace{\mathbb{E}\!\left[\|\bm{\epsilon}\|^2\right]}_{=\;\sigma_{1,2}^2} \;+\; \underbrace{\mathbb{E}\!\left[\|\bm{\delta}\|^2\right]}_{\geq\; 0} \;\geq\; \sigma_{1,2}^2.
    \end{equation}
    Equality holds if and only if $\mathbb{E}[\|\bm{\delta}\|^2] = 0$, i.e., $g(\bm{r}_1) = \mathbb{E}[\bm{r}_2 \mid \bm{r}_1]$ almost surely.
\end{proof}

\subsection{Private Fraction: Derivation from the Law of Total Variance}
\label{app:private_fraction}

We derive the identity $\eta_{1,2} = 1 - R^2_{\mathrm{pop}}$ from the law of total variance, connecting the private fraction to the population coefficient of determination.

\paragraph{The law of total variance.}
For any random vectors $X, Y$ with finite second moments, the law of total variance (also known as Eve's law) states:
\begin{equation}\label{eq:total_var}
    \mathrm{Var}(Y) \;=\; \mathbb{E}\!\left[\mathrm{Var}(Y \mid X)\right] \;+\; \mathrm{Var}\!\left(\mathbb{E}[Y \mid X]\right),
\end{equation}
where all variance and expectation operators are defined component-wise for vector-valued $Y$. In our setting, we use the trace (total) variance $\mathrm{Var}(Y) := \mathbb{E}[\|Y - \mathbb{E}[Y]\|^2]$.

\paragraph{Application to $Y = \bm{r}_2$, $X = \bm{r}_1$.}
Setting $Y = \bm{r}_2$ and $X = \bm{r}_1$, Eq.~\eqref{eq:total_var} becomes:
\begin{equation}\label{eq:total_var_applied}
    \underbrace{\mathbb{E}\!\left[\left\|\bm{r}_2 - \mathbb{E}[\bm{r}_2]\right\|^2\right]}_{\mathrm{Var}(\bm{r}_2)} \;=\; \underbrace{\mathbb{E}\!\left[\left\|\bm{r}_2 - \mathbb{E}[\bm{r}_2 \mid \bm{r}_1]\right\|^2\right]}_{\text{unexplained variance} \;=\; \sigma^2_{1,2}} \;+\; \underbrace{\mathbb{E}\!\left[\left\|\mathbb{E}[\bm{r}_2 \mid \bm{r}_1] - \mathbb{E}[\bm{r}_2]\right\|^2\right]}_{\text{explained variance}}.
\end{equation}

We briefly verify the second term. By the tower property of conditional expectation, $\mathbb{E}\!\left[\mathbb{E}[\bm{r}_2 \mid \bm{r}_1]\right] = \mathbb{E}[\bm{r}_2]$. Therefore, defining $\bm{m}(\bm{r}_1) := \mathbb{E}[\bm{r}_2 \mid \bm{r}_1]$, the explained variance is $\mathbb{E}[\|\bm{m}(\bm{r}_1) - \mathbb{E}[\bm{m}(\bm{r}_1)]\|^2] = \mathrm{Var}(\bm{m}(\bm{r}_1))$, confirming the standard form.

\paragraph{Deriving the identity.}
Dividing both sides of Eq.~\eqref{eq:total_var_applied} by $\mathrm{Var}(\bm{r}_2)$:
\begin{equation}
    1 \;=\; \frac{\sigma_{1,2}^2}{\mathrm{Var}(\bm{r}_2)} \;+\; \frac{\text{explained variance}}{\mathrm{Var}(\bm{r}_2)}.
\end{equation}
The second term on the right is precisely the population $R^2$ (the fraction of variance in $\bm{r}_2$ that is explained by the optimal predictor $\mathbb{E}[\bm{r}_2 \mid \bm{r}_1]$):
\begin{equation}
    R^2_{\mathrm{pop}} \;=\; \frac{\mathrm{Var}\!\left(\mathbb{E}[\bm{r}_2 \mid \bm{r}_1]\right)}{\mathrm{Var}(\bm{r}_2)} \;=\; 1 - \frac{\sigma_{1,2}^2}{\mathrm{Var}(\bm{r}_2)}.
\end{equation}
Rearranging:
\begin{equation}\label{eq:eta_identity}
    \boxed{\eta_{1,2} \;:=\; \frac{\sigma_{1,2}^2}{\mathrm{Var}(\bm{r}_2)} \;=\; 1 - R^2_{\mathrm{pop}}}
\end{equation}

\paragraph{Interpretation.}
Eq.~\eqref{eq:eta_identity} decomposes the total variance of $\bm{r}_2$ into two complementary fractions:
\begin{equation}
    \underbrace{1}_{\text{total}} \;=\; \underbrace{\eta_{1,2}}_{\text{private (unexplained by}  {\hspace{2pt}\mathbf{r}_1)}} \;+\; \underbrace{R^2_{\mathrm{pop}}}_{\text{explainable from }\bm{r}_1}.
\end{equation}
The private fraction $\eta_{1,2}$ measures the share of $\mathbf{r}_2$'s variance that is invisible to any function of the source residual $\mathbf{r}_1$; it is private only relative to $\mathbf{r}_1$, since information carried by other variables, such as the keys, the prompts, or language identity, is not accounted for. When $\eta_{1,2} \approx 1$, as our estimates suggest, no correction strategy operating on $\mathbf{r}_1$ alone can substantially reduce the value gap.

The denominator $\mathrm{Var}(\bm{r}_2)$ serves as a normalization, measuring how spread out the target residuals are across facts. A higher $\mathrm{Var}(\bm{r}_2)$ means the target updates are more diverse (different facts require geometrically different corrections in $\ell_2$), which---when combined with high $\eta_{1,2}$---implies a larger absolute floor $\sigma^2_{1,2} = \eta_{1,2}\operatorname{Var}(\mathbf{r}_2)$ on the error of any source-residual-only correction.

%% file: appendix/estimators.tex
\subsection{Estimator Specifications and Diagnostics}
\label{app:estimator_details}

All three estimators operate on PCA-reduced representations of the residual (or key/value) vectors. We first describe the shared PCA reduction procedure, then specify each estimator. Throughout, $D$ denotes the full vector dimension (e.g., 4096 for Llama-3.1-8B) and $N$ denotes the number of facts.

% ------------------------------------------------------------------
\subsubsection{PCA Reduction Procedure}
\label{app:pca_procedure}

\paragraph{Independence across languages.}
PCA bases are computed \emph{independently per language}, not jointly across the language pair. For a source--target pair $(\ell_1, \ell_2)$ at a given layer, we obtain separate bases $U_1, \bm{\mu}_1$ and $U_2, \bm{\mu}_2$ from the respective training sets. This ensures that each language's PCA basis captures its own dominant variance directions, without contamination from the other language's geometry.

\paragraph{Procedure.}
For language $\ell$ at layer $l^*$:
\begin{enumerate}[leftmargin=2em, itemsep=2pt]
    \item \textbf{Collect training vectors.} Assemble the training residuals into a matrix $X_\ell^{\mathrm{train}} \in \mathbb{R}^{D \times N_{\mathrm{train}}}$.
    \item \textbf{Center.} Compute the column-wise mean $\bm{\mu}_\ell = \frac{1}{N_{\mathrm{train}}}\sum_i \bm{x}_\ell^{(i)}$ and form $\tilde{X}_\ell = X_\ell^{\mathrm{train}} - \bm{\mu}_\ell \mathbf{1}^\top$.
    \item \textbf{SVD.} Compute the thin SVD $\tilde{X}_\ell = U \Sigma V^\top$, retaining up to $\min(200, \mathrm{rank}(\tilde{X}_\ell))$ components.
    \item \textbf{Projection.} For a target dimension $p$, the PCA code of a vector $\bm{x}$ is $\bm{z} = U_{:,1:p}^\top (\bm{x} - \bm{\mu}_\ell) \in \mathbb{R}^p$.
    \item \textbf{Reconstruction.} The approximate reconstruction is $\hat{\bm{x}} = U_{:,1:p}\,\bm{z} + \bm{\mu}_\ell$.
\end{enumerate}

\paragraph{Cumulative explained variance.}
For each language and layer, we report the cumulative explained variance as a function of the number of retained components:
\begin{equation}
    \mathrm{CEV}(p) \;=\; \frac{\sum_{i=1}^{p} \sigma_i^2}{\sum_{j=1}^{\mathrm{rank}} \sigma_j^2},
\end{equation}
where $\sigma_i$ are the singular values from the SVD. These curves are stored per language pair and layer.

\paragraph{PCA ceiling.}
For each reduced dimension $p \in \{25, 50, 100\}$, we compute a \emph{PCA reconstruction ceiling}: the $R^2$ in the full $D$-dimensional target space between the true test vectors $X_2^{\mathrm{test}}$ and their PCA reconstruction $\hat{X}_2^{\mathrm{test}} = U_{2,:,1:p}\,U_{2,:,1:p}^\top(X_2^{\mathrm{test}} - \bm{\mu}_2) + \bm{\mu}_2$. This ceiling quantifies the maximum $R^2$ achievable by \emph{any} estimator operating in the $p$-dimensional target PCA space, since even a perfect predictor of the codes $\bm{z}_2$ cannot recover information discarded by the PCA truncation.

\paragraph{Dimension selection.}
All estimators are evaluated at $p \in \{25, 50, 100\}$ (filtered to $p \leq \min(\mathrm{rank}_1, \mathrm{rank}_2, 200)$). We report the best $R^2$ across dimensions for each estimator. In practice, $R^2$ saturates early: increasing $p$ beyond 50 yields negligible improvement, consistent with the intrinsic low-dimensionality of the predictable component.

% ------------------------------------------------------------------
\subsubsection{Ridge Regression (E1)}
\label{app:ridge_details}

\paragraph{Model.}
Ridge regression fits a linear map $T \in \mathbb{R}^{p \times p}$ between source and target PCA codes:
\begin{equation}
    \hat{\bm{z}}_2 \;=\; T\,\bm{z}_1, \qquad T \in \mathbb{R}^{p \times p}.
\end{equation}
Here $\bm{z}_1 = U_1^\top(\bm{r}_1 - \bm{\mu}_1) \in \mathbb{R}^p$ and $\hat{\bm{z}}_2 \in \mathbb{R}^p$; we use $q = p$ (same reduced dimension for source and target).

\paragraph{Closed-form solution.}
Given training code matrices $Z_1, Z_2 \in \mathbb{R}^{p \times N_{\mathrm{train}}}$, the Ridge solution with Tikhonov regularization is:
\begin{equation}
    T \;=\; (Z_2\,Z_1^\top)\,(Z_1\,Z_1^\top + \lambda I)^{-1},
\end{equation}
where $\lambda$ is set automatically as:
\begin{equation}
    \lambda \;=\; 0.01 \cdot \frac{\mathrm{tr}(Z_1 Z_1^\top)}{p}.
\end{equation}
This scales the regularization proportionally to the average eigenvalue of the source Gram matrix, avoiding both under- and over-regularization without cross-validation.

\paragraph{Parameter count.} $p^2$ entries in $T$ (e.g., $2{,}500$ for $p = 50$).

\paragraph{Reconstruction and evaluation.}
Predicted target codes $\hat{Z}_2^{\mathrm{test}} = T\,Z_1^{\mathrm{test}}$ are reconstructed to full space via $\hat{X}_2 = U_2\,\hat{Z}_2 + \bm{\mu}_2$. The reported $R^2$ is computed in the full $D$-dimensional target space:
\begin{equation}
    R^2_{\mathrm{full}} \;=\; 1 - \frac{\sum_i \|\bm{r}_2^{(i)} - \hat{\bm{r}}_2^{(i)}\|^2}{\sum_i \|\bm{r}_2^{(i)} - \bar{\bm{r}}_2\|^2}.
\end{equation}

\paragraph{Shuffled-pair control.}
As a sanity check, we refit $T$ after randomly permuting the columns of $Z_2^{\mathrm{train}}$ relative to $Z_1^{\mathrm{train}}$, breaking the case alignment. This control yields $R^2 \approx 0$ on held-out data, confirming that the (already small) unshuffled $R^2$ reflects genuine cross-lingual structure rather than spurious dimensionality effects.

% ------------------------------------------------------------------
\subsubsection{Two-Layer MLP (E2)}
\label{app:mlp_details}

\paragraph{Architecture.}
A two-layer feedforward network maps source codes to target codes:
\begin{equation}
    \hat{\bm{z}}_2 \;=\; W_2\,\mathrm{ReLU}(W_1\,\bm{z}_1 + \bm{b}_1) + \bm{b}_2,
\end{equation}
with $W_1 \in \mathbb{R}^{h \times p}$, $W_2 \in \mathbb{R}^{p \times h}$, hidden dimension $h = \max(4, \lfloor 2p \rfloor)$ (i.e., $h = 2p$ for all $p \geq 2$). By the universal approximation theorem, this architecture can capture smooth nonlinear structure in the $\bm{z}_1 \to \bm{z}_2$ mapping, complementing the linear Ridge estimator.

\paragraph{Parameter count.} $h(p + 1) + p(h + 1) = 2hp + h + p$ (e.g., $10{,}150$ for $p = 50$, $h = 100$).

\paragraph{Training.}
\begin{itemize}[leftmargin=2em, itemsep=2pt]
    \item \textbf{Loss:} Mean squared error on PCA codes.
    \item \textbf{Optimizer:} Adam with learning rate $3 \times 10^{-4}$ and weight decay $10^{-3}$.
    \item \textbf{Epochs:} Up to 500.
    \item \textbf{Validation and early stopping:} 20\% of the training set is held out for validation ($\max(1, \lfloor 0.2 N_{\mathrm{train}} \rfloor)$ samples). Training halts if validation MSE fails to improve for 30 consecutive epochs, and the best-validation-loss weights are restored.
    \item \textbf{Device:} CUDA if available, otherwise CPU.
\end{itemize}

\paragraph{Seed aggregation.}
To reduce variance from random initialization and train/val splits, each configuration is trained with three seeds $\{0, 1, 2\}$. We report the \emph{median} $R^2_{\mathrm{full}}$ over seeds for both training and test sets.

\paragraph{Overfitting diagnostics.}
Two safeguards are employed. First, validation-based early stopping (above) prevents overfitting to training noise. Second, a \emph{shuffled-target control}: for each seed, we retrain the MLP with the rows of $Z_2^{\mathrm{train}}$ randomly permuted, breaking the fact-level correspondence. The median training $R^2$ under this shuffle is recorded; values near zero confirm that the MLP's capacity does not allow it to memorize arbitrary input--output pairings, and that the (already small) unshuffled $R^2$ reflects genuine structure.

% ------------------------------------------------------------------
\subsubsection{$k$-Nearest Neighbors (E3)}
\label{app:knn_details}

\paragraph{Prediction.}
For each test source code $\bm{z}_1^{\mathrm{test}}$, the $k$-NN estimator identifies the $k$ nearest training source codes and averages their corresponding target codes:
\begin{equation}
    \hat{\bm{z}}_2 \;=\; \frac{1}{k}\sum_{j \in \mathcal{N}_k(\bm{z}_1^{\mathrm{test}})} \bm{z}_2^{(j)},
\end{equation}
where $\mathcal{N}_k(\bm{z}_1^{\mathrm{test}})$ denotes the index set of the $k$ nearest neighbors in the training set. Reconstruction to full space follows the same procedure as Ridge ($\hat{\bm{r}}_2 = U_2\,\hat{\bm{z}}_2 + \bm{\mu}_2$).

\paragraph{Distance metrics.}
We evaluate two metrics:
\begin{itemize}[leftmargin=2em, itemsep=2pt]
    \item \textbf{Euclidean} (primary): $d(\bm{a}, \bm{b}) = \|\bm{a} - \bm{b}\|_2$.
    \item \textbf{Cosine} (secondary): $d(\bm{a}, \bm{b}) = 1 - \frac{\bm{a}^\top \bm{b}}{\|\bm{a}\|\,\|\bm{b}\|}$, with a floor of $10^{-12}$ on norms to avoid division by zero.
\end{itemize}
% Neighbor search uses \texttt{argpartition} (NumPy) or \texttt{topk} (PyTorch/CUDA) for efficiency.

\paragraph{Choice of $k$.}
We sweep over $k \in \{5, 10, 20\}$ (capped at $\min(k, N_{\mathrm{train}})$). Each $(k, \mathrm{metric})$ combination is evaluated independently; the best configuration is selected by highest $R^2_{\mathrm{full}}$ on the test set.

\paragraph{Parameter count.}
Zero learned parameters. The $k$-NN estimator is nonparametric and, under standard regularity conditions, is consistent for estimating $\mathbb{E}[r_2\mid r_1]$ when $N\to\infty$, $k\to\infty$, and $k/N\to0$. It provides a complementary approach to the parametric Ridge and MLP estimators. However, it is subject to the curse of dimensionality: at $p>50$, the effective neighborhood structure degrades, consistent with the lack of improvement over Ridge at higher PCA dimensions.

% ------------------------------------------------------------------
\paragraph{Train/test split (shared across estimators).}
For all three estimators, the $N$ available case ids are sorted, then permuted with a fixed random seed (default 42). The first $\lfloor 0.6 N \rfloor$ cases form the training set; the remainder form the test set. For each source--target language pair, train and test sets are intersected with the case ids present in both languages to ensure alignment. If this intersection is too small but at least 8 shared cases exist, the split is recomputed over the shared ids with the same seed and fraction. We require $\geq 8$ source cases and a non-empty test set to proceed.

\subsection{Extended $R^2$ Results}
\label{app:r2_extended}

\begin{figure}[h]
    \centering
    \begin{subfigure}{0.32\textwidth}
        \centering
        \includegraphics[width=\linewidth]{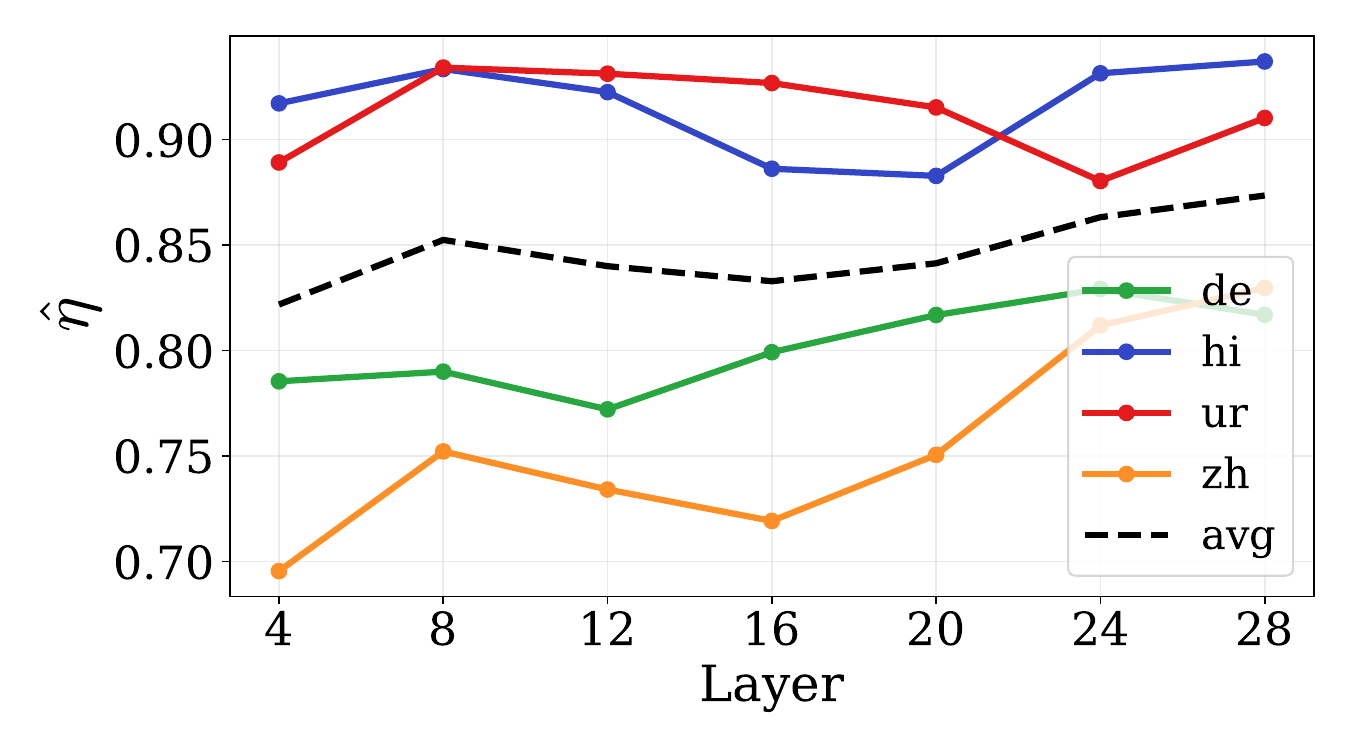}
        \caption{Keys}
        \label{fig:eta_kvr_keys_llama}
    \end{subfigure}
    \hfill
    \begin{subfigure}{0.32\textwidth}
        \centering
        \includegraphics[width=\linewidth]{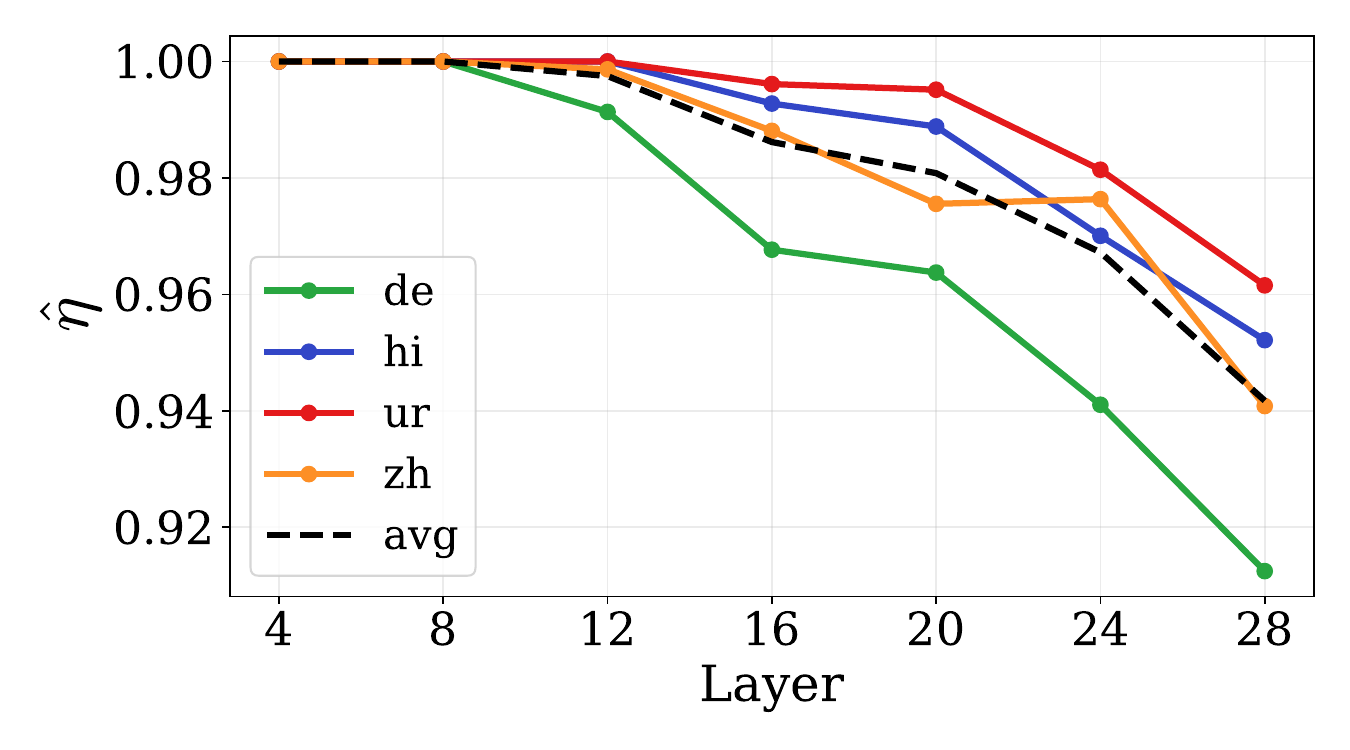}
        \caption{Values}
        \label{fig:eta_kvr_values_llama}
    \end{subfigure}
    \hfill
    \begin{subfigure}{0.32\textwidth}
        \centering
        \includegraphics[width=\linewidth]{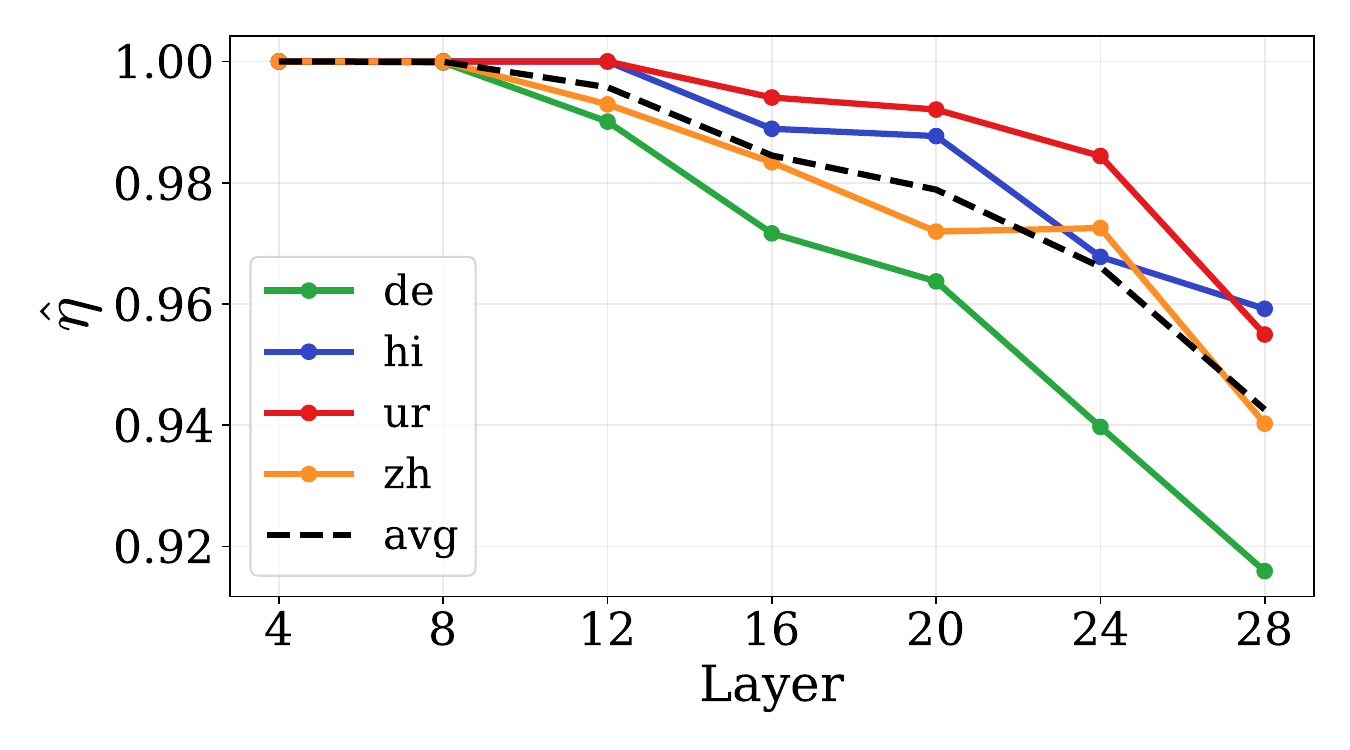}
        \caption{Residuals}
        \label{fig:eta_kvr_residual_llama}
    \end{subfigure}
    \caption{Estimated private fraction $\hat{\eta}$ across layers for keys (left), values (middle), and residuals (right) on LLaMA-3.1-8B. Values and residuals are both severely opaque ($\hat{\eta} \geq 0.9$), confirming the value space as the fundamental bottleneck.}
    \label{fig:eta_kvr_llama}
\end{figure}

\paragraph{$\hat{\eta}$ vs.\ layer.}
Figure~\ref{fig:eta_kvr_llama} shows $\hat{\eta} = 1 - R^2_{\mathrm{best}}$ across all layers for each language pair, for keys, values and residuals. The residual's private fraction remains above 0.9 at all layers for all language pairs, and decreases with layer depth. German shows the lowest $\hat{\eta}$ ($\approx 0.964$ at the optimal layer 20), while Urdu remains the highest ($\hat{\eta} \approx 0.992$). The consistency of high $\hat{\eta}$ across layers rules out the possibility that the residual-prediction barrier is an artifact of a suboptimal layer choice. Table~\ref{tab:r2_layerwise_llama} details the layerwise results for LLaMA-3.1-8B residuals. We have a clear trend of increasing, but still low, residual predictability. Note that we cap $R^2_{\mathrm{best}}$ at 0 (so that $\hat{\eta} \le 1$) because negative $R^2$ implies that the predictor is worse than the mean of the sample population, and hence not a good option for us.

\begin{table}[H]
\centering
\small
\begin{tabular}{cl rrrrr}
\toprule
\textbf{Layer} & \textbf{Target} & \textbf{Ridge} & \textbf{MLP} & \textbf{KNN} & \textbf{Best $R^2$} & \textbf{PVT Frac.} \\
\midrule
\multirow{4}{*}{4}
 & de & \textbf{$-$0.0096} & $-$0.0109 & $-$0.0121 & $-$0.0096 & 1.0000 \\
 & hi & $-$0.0105 & \textbf{$-$0.0103} & $-$0.0145 & $-$0.0103 & 1.0000 \\
 & ur & \textbf{$-$0.0107} & $-$0.0117 & $-$0.0158 & $-$0.0107 & 1.0000 \\
 & zh & \textbf{$-$0.0072} & $-$0.0075 & $-$0.0128 & $-$0.0072 & 1.0000 \\
\midrule
\multirow{4}{*}{8}
 & de & \textbf{0.0002} & $-$0.0020 & $-$0.0045 & 0.0002 & 0.9998 \\
 & hi & \textbf{$-$0.0053} & $-$0.0063 & $-$0.0093 & $-$0.0053 & 1.0000 \\
 & ur & \textbf{$-$0.0044} & $-$0.0054 & $-$0.0069 & $-$0.0044 & 1.0000 \\
 & zh & \textbf{0.0000} & $-$0.0023 & $-$0.0027 & 0.0000 & 1.0000 \\
\midrule
\multirow{4}{*}{12}
 & de & \textbf{0.0099} & 0.0065 & 0.0058 & 0.0099 & 0.9901 \\
 & hi & \textbf{$-$0.0025} & $-$0.0035 & $-$0.0057 & $-$0.0025 & 1.0000 \\
 & ur & \textbf{$-$0.0031} & $-$0.0042 & $-$0.0075 & $-$0.0031 & 1.0000 \\
 & zh & \textbf{0.0071} & 0.0044 & 0.0046 & 0.0071 & 0.9929 \\
\midrule
\multirow{4}{*}{16}
 & de & \textbf{0.0284} & 0.0239 & 0.0259 & 0.0284 & 0.9716 \\
 & hi & \textbf{0.0111} & 0.0075 & 0.0077 & 0.0111 & 0.9889 \\
 & ur & \textbf{0.0059} & 0.0005 & 0.0042 & 0.0059 & 0.9941 \\
 & zh & \textbf{0.0166} & 0.0148 & 0.0128 & 0.0166 & 0.9834 \\
\midrule
\multirow{4}{*}{20}
 & de & \textbf{0.0363} & 0.0322 & 0.0359 & 0.0363 & 0.9637 \\
 & hi & 0.0114 & 0.0085 & \textbf{0.0123} & 0.0123 & 0.9877 \\
 & ur & 0.0049 & \textbf{0.0079} & 0.0054 & 0.0079 & 0.9921 \\
 & zh & \textbf{0.0281} & 0.0244 & 0.0227 & 0.0281 & 0.9719 \\
\midrule
\multirow{4}{*}{24}
 & de & 0.0554 & 0.0506 & \textbf{0.0603} & 0.0603 & 0.9397 \\
 & hi & \textbf{0.0322} & 0.0278 & 0.0275 & 0.0322 & 0.9678 \\
 & ur & \textbf{0.0156} & 0.0134 & 0.0126 & 0.0156 & 0.9844 \\
 & zh & 0.0264 & 0.0274 & \textbf{0.0275} & 0.0275 & 0.9725 \\
\midrule
\multirow{4}{*}{28}
 & de & 0.0834 & 0.0764 & \textbf{0.0841} & 0.0841 & 0.9159 \\
 & hi & 0.0323 & 0.0353 & \textbf{0.0408} & 0.0408 & 0.9592 \\
 & ur & 0.0393 & 0.0322 & \textbf{0.0450} & 0.0450 & 0.9550 \\
 & zh & 0.0577 & 0.0522 & \textbf{0.0597} & 0.0597 & 0.9403 \\
\bottomrule
\end{tabular}
\vspace{6pt}
\caption{Layer-wise $R^2$ scores for each target language (source: English). Bold indicates the best estimator per row.}
\label{tab:r2_layerwise_llama}
\end{table}

%% file: appendix/contrastive.tex
\section{Contrastive Alignment Experiment Details}
\label{app:contrastive_full}

\subsection{Data Preparation}
\label{app:data_prep}

For the contrastive alignment fine-tuning phase, the primary training corpus was derived from the Zero-Shot Relation Extraction (zsRE) dataset. To construct precisely aligned cross-lingual training batches, the dataset was filtered by unique case IDs to ensure that every factual assertion possessed parallel structural components—specifically, the probing prompt and the target subject anchor—across all chosen target languages. Alongside this factual objective, we integrated a language modeling regularization corpus consisting of cleaned Wikipedia documents and their corresponding translations. This secondary corpus was partitioned into training and validation splits to continuously monitor perplexity and mitigate catastrophic forgetting of general linguistic capabilities during the alignment optimization. Following this fine-tuning procedure, the CounterFact dataset was reserved as the primary downstream benchmark to rigorously evaluate the cross-lingual generalization and specificity of the resulting edits.

\subsection{Training Details}
\label{app:training_details}

The complete list of hyperparameters utilized during the cross-lingual contrastive alignment fine-tuning phase is detailed in Table~\ref{tab:alignment_hyperparams}.

\begin{table}[h!]
\centering
\small
\begin{tabular}{lc}
\toprule
\textbf{Hyperparameter} & \textbf{Value} \\
\midrule
\multicolumn{2}{c}{\textit{Model \& Architecture Details}} \\
\midrule
Base Architecture & Llama-3.1-8B / Gemma-3-4B \\
Target Modules & \texttt{up\_proj}, \texttt{down\_proj} \\
Targeted Semantic Layers & 16 -- 25 \\
\midrule
\multicolumn{2}{c}{\textit{Contrastive Alignment Objective}} \\
\midrule
Training Epochs & 10 \\
Learning Rate & $1 \times 10^{-4}$ \\
Alignment Batch Size ($B$) & 128 \\
Maximum Fact Capacity & 2000 \\
InfoNCE Temperature ($\tau$) & 0.07 \\
Fact Anchor Token & \texttt{subject\_last} \\
Value-Space Optimization & False \\
\midrule
\multicolumn{2}{c}{\textit{Language Modeling Regularization}} \\
\midrule
LM Loss Weights ($\beta$) & $\{0.5, 1.0, 2.0\}$ \\
LM Batch Size & 8 \\
Max Sequence Length & 256 \\
Wikipedia Corpus Train Split & 80\% \\
PPL Degradation Tolerance & 5\% ($\Delta \text{PPL} \le 0.05$) \\
Eval Batches per Language & 100 \\
\bottomrule
\end{tabular}
\vspace{6pt}
\caption{Hyperparameters utilized during the cross-lingual contrastive alignment fine-tuning phase.}
\label{tab:alignment_hyperparams}
\end{table}

\subsection{Per-Epoch Perplexity Monitoring and Optimal Training Window}
\label{app:ppl_monitoring}

\begin{figure}[htbp]
    \centering
    \begin{subfigure}[b]{0.48\textwidth}
        \centering
        \includegraphics[width=\textwidth]{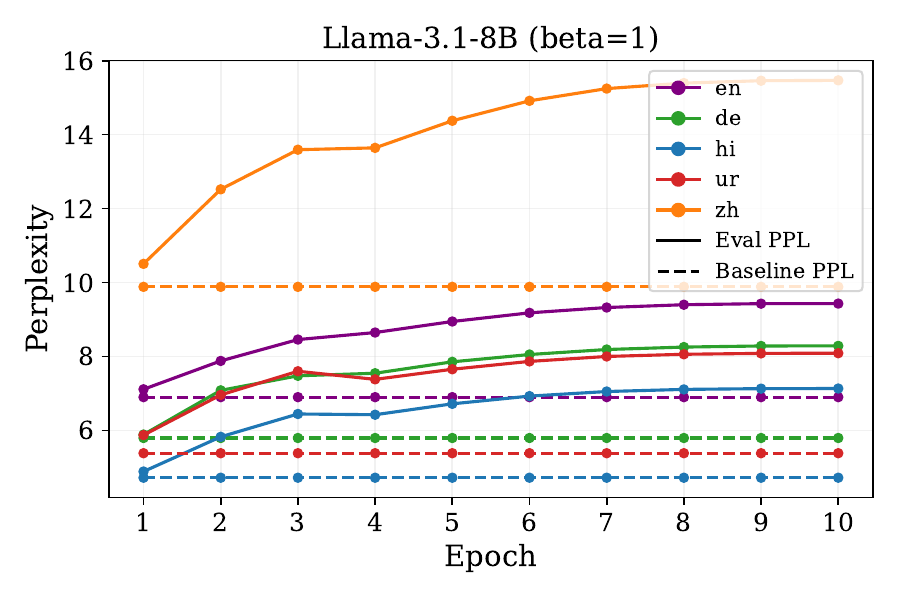}
        \caption{Llama-3.1-8B}
        \label{fig:llama_ppl}
    \end{subfigure}
    \hfill
    \begin{subfigure}[b]{0.48\textwidth}
        \centering
        \includegraphics[width=\textwidth]{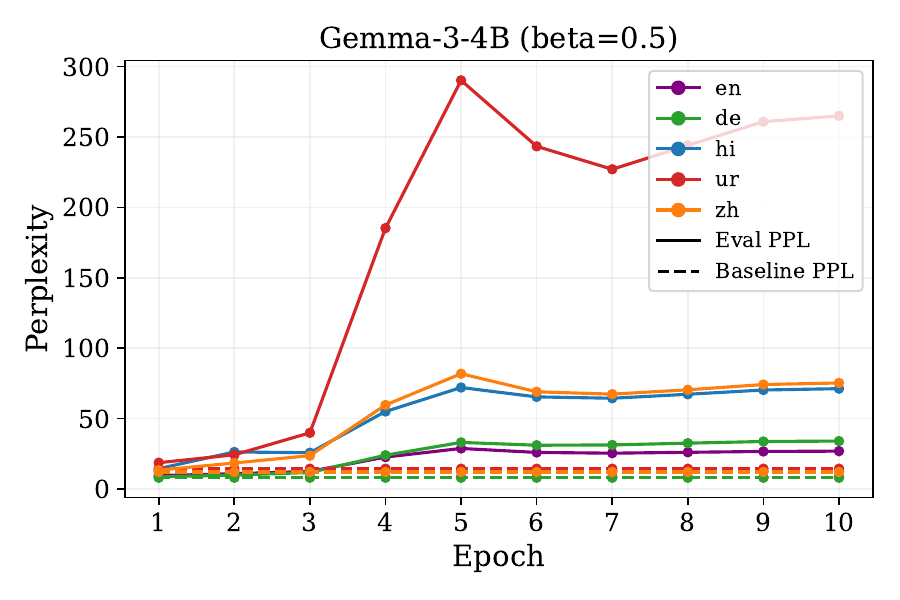}
        \caption{Gemma-3-4B}
        \label{fig:gemma_ppl}
    \end{subfigure}
    
    \caption{\textbf{Evaluation perplexity and the optimal training window.} Solid lines represent evaluation perplexity; dashed lines represent the pre-edit baseline. The optimal training window is highlighted by the stable region in Epochs 1--2 (Llama) and Epochs 1--3 (Gemma). Training beyond these points results in a rapid explosion of perplexity -- most notably the Urdu (\texttt{ur}) peak in Gemma -- signaling the onset of substantial degradation in language-modeling performance.}
    \label{fig:ppl_stability}
\end{figure}

To identify the optimal duration for contrastive alignment, we analyze the trajectory of evaluation perplexity relative to the pre-finetuning baseline. As illustrated in Figure~\ref{fig:ppl_stability}, the best training window is extremely narrow. For \textit{Llama-3.1-8B}, linguistic integrity is relatively preserved only during Epochs 1--2, after which the surprise direction $(\bm{e}_{o_2} - \bm{q}^{(0)})$ is severely corrupted by rising perplexity in distant languages like Chinese.

The degradation is significantly more volatile in \textit{Gemma-3-4B}. The model remains viable only through Epochs 1--3. By Epoch 4, the model reaches a catastrophic failure point where the language modeling regularizer ($\beta = 0.5$) fails completely; perplexity for Urdu (\texttt{ur}) explodes from a baseline of ${\sim}15$ to a peak of nearly $300$ by Epoch 5. This empirical "explosion" confirms that extended contrastive training beyond the first three epochs induces a total loss of multilingual fluency, necessitating early stopping to maintain model utility.

\subsection{Comparative Performance: Baseline vs. Contrastively Aligned Models}
\label{app:intervention_evaluation}

To assess the effect of contrastive alignment, we compare editing performance before and after alignment. Table~\ref{tab:sft_vs_base_efficacy} shows mixed effects on cross-lingual editing efficacy. Alignment improves performance in several settings, including MEMIT on Gemma-3-4B at layers 16--20 and 20--24, and ROME on Llama-3.1-8B for Urdu and Chinese at layer 20. However, it also reduces performance in other settings, notably ROME on Gemma-3-4B (14 of 15 language--layer settings). Table~\ref{tab:sft_vs_base_nll} reports the corresponding changes in log-likelihood margins, which also vary across models and editing methods. These results show that improving representation alignment does not consistently improve editing success, as the value-space mismatch identified in Section~\ref{sec:value_space_barrier} remains unresolved.

\begin{table*}[ht]
\centering
\setlength{\tabcolsep}{5pt}
\renewcommand{\arraystretch}{1.2}
\resizebox{\textwidth}{!}{%
\begin{tabular}{llccccccc}
\toprule
\textbf{Method} & \textbf{Model} & \textbf{Layers}
  & \textbf{en (Aligned / Base)}
  & \textbf{de (Aligned / Base)}
  & \textbf{hi (Aligned / Base)}
  & \textbf{ur (Aligned / Base)}
  & \textbf{zh (Aligned / Base)} \\
\midrule

\multirow{6}{*}{ROME}
  & \multirow{3}{*}{\textit{Gemma-3-4B}}
    & Layer 16 & \textbf{0.5466} / 0.4615 & \textbf{0.3682} / 0.2053 & \textbf{0.0129} / -0.0813 & \textbf{-0.0883} / -0.1102 & \textbf{-0.0948} / -0.1403 \\
  & & Layer 20 & 0.3437 / \textbf{1.0960} & 0.1776 / \textbf{0.4939} & 0.0016 / \textbf{0.1400} & -0.0334 / \textbf{0.0477} & 0.0055 / \textbf{0.0244} \\
  & & Layer 24 & 0.5908 / \textbf{1.1212} & 0.3594 / \textbf{0.6356} & 0.0232 / \textbf{0.1119} & -0.0190 / \textbf{0.0086} & 0.0032 / \textbf{0.0401} \\
\cmidrule(lr){2-8}
  & \multirow{3}{*}{\textit{Llama-3.1-8B}}
    & Layer 16 & \textbf{1.0365} / -0.8460 & \textbf{0.7196} / -0.1831 & \textbf{-0.0186} / -0.2192 & \textbf{0.0403} / -0.2403 & \textbf{-0.3892} / -0.5233 \\
  & & Layer 20 & \textbf{1.4522} / 0.2435 & \textbf{1.1788} / 0.3758 & \textbf{-0.0012} / -0.0200 & \textbf{0.0577} / -0.0344 & \textbf{-0.1535} / -0.1970 \\
  & & Layer 24 & \textbf{1.8763} / -0.3794 & \textbf{1.4756} / 0.2568 & \textbf{0.0365} / 0.0119 & \textbf{0.0559} / -0.0011 & -0.0682 / \textbf{-0.0497} \\

\midrule

\multirow{6}{*}{MEMIT}
  & \multirow{3}{*}{\textit{Gemma-3-4B}}
    & Layers 16--20 & \textbf{-2.1923} / -3.7245 & \textbf{-2.5449} / -3.4626 & \textbf{-0.3590} / -3.5444 & \textbf{1.3378} / -2.8797 & \textbf{-1.5266} / -3.1953 \\
  & & Layers 20--24 & \textbf{0.2459} / -2.8631 & \textbf{-1.2547} / -4.4298 & \textbf{-2.1805} / -4.8292 & \textbf{-2.7245} / -6.7825 & \textbf{-1.5381} / -5.5828 \\
  & & Layers 24--28 & -1.5916 / \textbf{-1.2627} & \textbf{-1.7864} / -2.2729 & \textbf{-4.0304} / -5.7285 & \textbf{-2.7668} / -7.0345 & \textbf{-1.5802} / -3.2861 \\
\cmidrule(lr){2-8}
  & \multirow{3}{*}{\textit{Llama-3.1-8B}}
    & Layers 16--20 & \textbf{-3.9945} / -6.8043 & \textbf{-0.7920} / -2.0293 & \textbf{-0.2815} / -0.5678 & \textbf{-0.3256} / -0.6332 & \textbf{-1.1679} / -1.6074 \\
  & & Layers 20--24 & \textbf{-4.5408} / -8.0563 & \textbf{-0.2910} / -1.4522 & -0.1778 / \textbf{-0.0219} & -0.2413 / \textbf{-0.2167} & -0.5649 / \textbf{-0.2793} \\
  & & Layers 24--28 & \textbf{-6.0167} / -8.9126 & \textbf{-1.1943} / -1.6937 & -0.2144 / \textbf{-0.0167} & -0.0873 / \textbf{-0.0490} & -0.1957 / \textbf{-0.0940} \\

\midrule

\multirow{6}{*}{AlphaEdit}
  & \multirow{3}{*}{\textit{Gemma-3-4B}}
    & Layers 16--20 & \textbf{-2.4923} / -15.0990 & \textbf{-2.0107} / -13.4295 & \textbf{-0.9019} / -10.9384 & \textbf{0.3683} / -9.6519 & \textbf{-1.6365} / -11.3857 \\
  & & Layers 20--24 & -5.8981 / \textbf{-1.8940} & -7.7703 / \textbf{-0.9953} & -10.2174 / \textbf{0.9035} & -12.5731 / \textbf{0.9206} & -9.8445 / \textbf{-0.9681} \\
  & & Layers 24--28 & \textbf{3.7558} / 1.0111 & \textbf{3.9620} / -0.1612 & \textbf{2.6001} / 1.5192 & \textbf{3.5245} / -2.9782 & \textbf{5.7793} / 3.6437 \\
\cmidrule(lr){2-8}
  & \multirow{3}{*}{\textit{Llama-3.1-8B}}
    & Layers 16--20 & \textbf{8.0434} / 7.8645 & \textbf{4.7837} / 4.6289 & 0.2265 / \textbf{0.2879} & 0.0954 / \textbf{0.1759} & 0.0953 / \textbf{0.1612} \\
  & & Layers 20--24 & \textbf{0.2121} / -0.0211 & 0.3214 / \textbf{0.3711} & \textbf{0.0303} / 0.0218 & -0.0021 / \textbf{0.0348} & 0.0039 / \textbf{0.0054} \\
  & & Layers 24--28 & 0.0652 / \textbf{0.3981} & 0.0529 / \textbf{0.1366} & \textbf{0.0040} / -0.0277 & \textbf{-0.0033} / -0.0068 & 0.0014 / \textbf{0.0238} \\

\bottomrule
\end{tabular}%
}
\caption{$\Delta\mathrm{NLL}_{\mathrm{margin}}$ score comparison between Base and Aligned models on CounterFact. Higher positive values indicate greater success in updating factual knowledge.}
\label{tab:sft_vs_base_nll}
\end{table*}

\begin{table*}[ht]
\centering

\setlength{\tabcolsep}{5pt}
\renewcommand{\arraystretch}{1.2}
\resizebox{\textwidth}{!}{%
\begin{tabular}{llccccccc}
\toprule
\textbf{Method} & \textbf{Model} & \textbf{Layers}
  & \textbf{en (Aligned / Base)}
  & \textbf{de (Aligned / Base)}
  & \textbf{hi (Aligned / Base)}
  & \textbf{ur (Aligned / Base)}
  & \textbf{zh (Aligned / Base)} \\
\midrule
\multirow{6}{*}{ROME}
  & \multirow{3}{*}{\textit{Gemma-3-4B}}
    & Layer 16 & 61.55 / \textbf{71.17} & 58.45 / \textbf{61.26} & 43.88 / \textbf{50.97} & 40.78 / \textbf{48.25} & 41.94 / \textbf{45.24} \\
  & & Layer 20 & 68.25 / \textbf{84.66} & 60.87 / \textbf{73.69} & 51.46 / \textbf{59.71} & 47.48 / \textbf{56.31} & 51.75 / \textbf{55.24} \\
  & & Layer 24 & 74.95 / \textbf{83.01} & 68.35 / \textbf{78.25} & 46.02 / \textbf{62.91} & \textbf{48.06} / 47.86 & 49.32 / \textbf{56.70} \\
\cmidrule(lr){2-8}
  & \multirow{3}{*}{\textit{Llama-3.1-8B}}
    & Layer 16 & \textbf{56.70} / 39.71 & \textbf{56.31} / 56.12 & \textbf{56.41} / 44.37 & \textbf{59.32} / 36.21 & \textbf{38.35} / 25.63 \\
  & & Layer 20 & \textbf{63.59} / 50.78 & 61.17 / \textbf{72.62} & \textbf{50.19} / 49.61 & \textbf{59.22} / 46.70 & \textbf{39.51} / 33.59 \\
  & & Layer 24 & \textbf{69.42} / 39.90 & 66.99 / \textbf{74.27} & \textbf{61.65} / 50.87 & \textbf{63.50} / 43.88 & \textbf{50.68} / 40.00 \\
\midrule
\multirow{6}{*}{MEMIT}
  & \multirow{3}{*}{\textit{Gemma-3-4B}}
    & Layers 16--20 & \textbf{31.78} / 21.96 & \textbf{28.86} / 20.41 & \textbf{45.97} / 12.44 & \textbf{73.37} / 14.77 & \textbf{26.53} / 10.30 \\
  & & Layers 20--24 & \textbf{56.37} / 21.96 & \textbf{37.90} / 15.35 & \textbf{21.96} / 8.94  & \textbf{15.35} / 2.92  & \textbf{24.59} / 4.96  \\
  & & Layers 24--28 & 27.99 / \textbf{29.35} & \textbf{23.91} / 23.13 & \textbf{3.89} / 3.40   & \textbf{8.84} / 0.97   & \textbf{17.88} / 4.28  \\
\cmidrule(lr){2-8}
  & \multirow{3}{*}{\textit{Llama-3.1-8B}}
    & Layers 16--20 & \textbf{15.35} / 5.15  & \textbf{27.89} / 13.02 & \textbf{18.08} / 7.09  & \textbf{11.86} / 8.65  & \textbf{1.36} / 0.87   \\
  & & Layers 20--24 & \textbf{13.70} / 3.89  & \textbf{42.66} / 19.14 & 25.46 / \textbf{34.69} & 14.87 / \textbf{20.51} & 7.77 / \textbf{17.30}  \\
  & & Layers 24--28 & \textbf{2.82} / 1.46   & \textbf{22.45} / 9.23  & 37.32 / \textbf{57.53} & \textbf{42.57} / 37.51 & 24.30 / \textbf{31.97} \\
\midrule
\multirow{6}{*}{AlphaEdit}
  & \multirow{3}{*}{\textit{Gemma-3-4B}}
    & Layers 16--20 & \textbf{32.82} / 7.86  & \textbf{33.79} / 8.16  & \textbf{46.70} / 7.67  & \textbf{64.85} / 4.76  & \textbf{30.58} / 5.44  \\
  & & Layers 20--24 & 32.23 / \textbf{45.53} & 25.73 / \textbf{49.13} & 11.26 / \textbf{56.70} & 6.50 / \textbf{59.32}  & 8.45 / \textbf{47.67}  \\
  & & Layers 24--28 & \textbf{68.16} / 47.38 & \textbf{65.53} / 48.45 & \textbf{64.66} / 57.67 & \textbf{71.17} / 30.87 & \textbf{85.63} / 66.12 \\
\cmidrule(lr){2-8}
  & \multirow{3}{*}{\textit{Llama-3.1-8B}}
    & Layers 16--20 & 96.70 / \textbf{96.99} & \textbf{90.87} / 89.61 & 52.14 / \textbf{58.74} & 47.77 / \textbf{54.08} & 51.46 / \textbf{57.86} \\
  & & Layers 20--24 & \textbf{57.57} / 52.72 & 52.52 / 52.52         & 47.86 / \textbf{48.54} & 40.29 / \textbf{45.83} & \textbf{52.04} / 51.17 \\
  & & Layers 24--28 & 56.99 / \textbf{58.45} & 52.23 / \textbf{53.69} & \textbf{47.77} / 45.92 & 35.44 / \textbf{44.95} & 48.93 / \textbf{64.76} \\
\bottomrule
\end{tabular}%
}
\caption{Efficacy score comparison on CounterFact. Values represent update success (Efficacy $\uparrow$), defined as the proportion of cases where the post-edit probability of the target answer exceeds that of the original answer.}
\label{tab:sft_vs_base_efficacy}
\end{table*}

\subsection{Downstream Performance on XNLI}
\label{app:xnli}

\begin{table}[H]
\centering
\setlength{\tabcolsep}{8pt}
\renewcommand{\arraystretch}{1.2}
\small
\begin{tabular}{llcc}
\toprule
\textbf{Model} & \textbf{Lang} & \textbf{Aligned} & \textbf{Base} \\
\midrule
\multirow{5}{*}{\textit{Llama-3.1-8B}}
  & en & 0.45 & \textbf{0.50} \\
  & de & 0.43 & \textbf{0.48} \\
  & hi & 0.41 & \textbf{0.47} \\
  & ur & 0.37 & \textbf{0.45} \\
  & zh & 0.38 & \textbf{0.42} \\
\bottomrule
\end{tabular}
\vspace{6pt}
\caption{XNLI evaluation scores for Llama-3.1-8B before (Base) and after (Aligned) finetuning across languages.}
\label{tab:xnli}
\end{table}

To monitor whether contrastive alignment affects general cross-lingual reasoning, we evaluated
downstream performance on the Cross-lingual Natural Language Inference (XNLI)~\cite{conneau-etal-2018-xnli} benchmark. As shown in Table~\ref{tab:xnli}, aligned models incur moderate accuracy drops relative to the
base model, but retain non-trivial zero-shot performance across the evaluated languages. This suggests
that the alignment intervention improves key-space reachability, with mixed effects on edit transfer, at the cost of some downstream NLU degradation,
motivating future work on better regularization.